\pdfoutput=1
\documentclass[10pt,twocolumn,letterpaper]{article}

\usepackage[pagenumbers]{cvpr}  %

\usepackage[T1]{fontenc}
\usepackage{amsthm}
\usepackage{mathtools}
\usepackage{multirow}
\usepackage{adjustbox}
\usepackage{colortbl}
\usepackage{microtype}
\usepackage{placeins}
\usepackage{float}

\newcommand{\E}{\mathbb{E}}

\DeclareMathOperator{\Var}{Var}
\DeclareMathOperator{\Cov}{Cov}
\DeclareMathOperator*{\argmin}{arg\,min}
\newcommand{\data}[1]{{\footnotesize\color{black!60}#1}}
\definecolor{tgood}{HTML}{D4EFD4}
\definecolor{tbad}{HTML}{F6D6D6}
\definecolor{tours}{HTML}{ECECEA}
\definecolor{vmsame}{HTML}{EBEBEB}
\definecolor{vmbetter}{HTML}{8FD18F}
\definecolor{vmworse}{HTML}{F29B9B}
\colorlet{vmbetterlt}{vmbetter!50!vmsame}
\colorlet{vmworselt}{vmworse!50!vmsame}
\colorlet{vmbetterxl}{vmbetter!25!vmsame}
\colorlet{vmworsexl}{vmworse!25!vmsame}
\definecolor{vmgood}{HTML}{1E8A1E}
\definecolor{vmbad}{HTML}{C62828}
\newcommand{\swatch}[2]{{\setlength{\fboxsep}{0.8pt}\colorbox{#1}{#2}}}
\newcommand{\dz}{\hphantom{0}}

\newcommand{\na}{\multicolumn{1}{c}{--}}
\newcommand{\hd}[1]{\multicolumn{1}{c}{#1}}
\newcommand{\best}[1]{{\setbox0\hbox{#1}\rlap{\rule[-1.1pt]{\wd0}{0.35pt}}\box0}}

\makeatletter
\newcommand{\oursrow}{\rlap{\color{tours}\rule[-\dp\@arstrutbox]{\linewidth}{\dimexpr\ht\@arstrutbox+\dp\@arstrutbox\relax}}}
\makeatother

\definecolor{cvprblue}{rgb}{0.21,0.49,0.74}
\usepackage[pagebackref,breaklinks,colorlinks,allcolors=cvprblue]{hyperref}

\theoremstyle{plain}

\theoremstyle{definition}

\theoremstyle{remark}

\usepackage[textsize=tiny]{todonotes}

\def\confName{CVPR}
\def\confYear{2027}

\title{Guidance as Variance Matching}

\AtBeginDocument{\everymath\expandafter{\the\everymath\thickmuskip=0mu\medmuskip=0mu}}

\makeatletter
\renewcommand\paragraph{\@startsection{paragraph}{4}{\z@}%
  {1.2ex \@plus .4ex \@minus .2ex}{-1em}{\normalfont\normalsize\bfseries}}
\newcommand\tightdisplays{%
  \abovedisplayskip=3pt plus 2pt minus 2pt
  \belowdisplayskip=3pt plus 2pt minus 2pt
  \abovedisplayshortskip=0pt plus 2pt
  \belowdisplayshortskip=2pt plus 2pt minus 1pt}
\newcommand\roomydisplays{%
  \abovedisplayskip=7pt plus 2pt minus 2pt
  \belowdisplayskip=7pt plus 2pt minus 2pt
  \abovedisplayshortskip=5pt plus 2pt
  \belowdisplayshortskip=6pt plus 2pt minus 1pt}
\g@addto@macro\normalsize{\tightdisplays}
\g@addto@macro\small{\tightdisplays}
\makeatother
\setlist[itemize]{topsep=2pt, itemsep=1pt, parsep=0pt, leftmargin=*}
\makeatletter
\def\cvprsection{\@startsection{section}{1}{\z@}{-8pt plus -2pt minus -2pt}{5pt}{\large\bf}}
\def\cvprsubsection{\@startsection{subsection}{2}{\z@}{-6pt plus -2pt minus -1pt}{3pt}{\elvbf}}

\makeatother
\usepackage{docmute}

\author{%
Ali Backour$^{1*}$\qquad Artem Lukoianov$^{1*}$\qquad Frank Permenter$^{2}$\\
Tommi Jaakkola$^{1}$\qquad Justin Solomon$^{1}$\qquad Vincent Sitzmann$^{1}$\\[4pt]
{\normalsize $^{1}$~Massachusetts Institute of Technology\qquad $^{2}$~Toyota Research Institute}\\[4pt]
{\large \href{https://ottogin.github.io/vmg/}{\textbf{ottogin.github.io/vmg}}}
}

\begin{document}
\twocolumn[{%
\renewcommand\twocolumn[1][]{#1}%
\maketitle
\begin{center}
  \captionsetup{type=figure}
  \setlength{\unitlength}{1in}%
\begin{picture}(6.8750,2.5800)%
\put(0.0000,0){\includegraphics{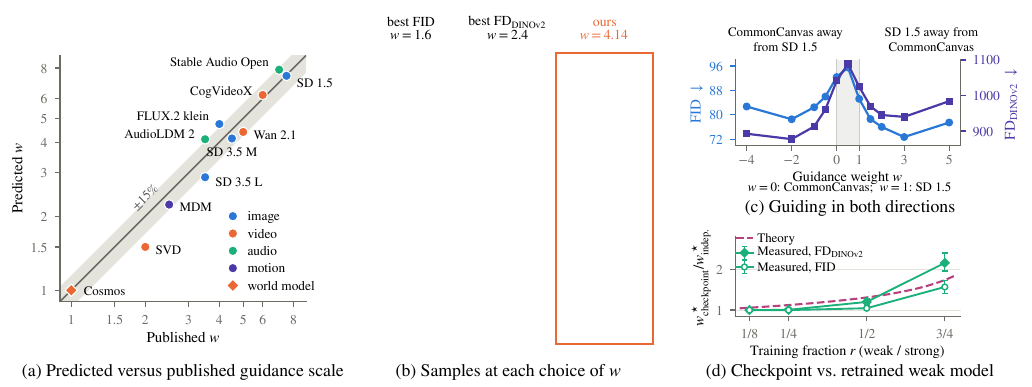}}%
\put(2.4311,1.5900){\includegraphics[width=0.6200in,height=0.6200in]{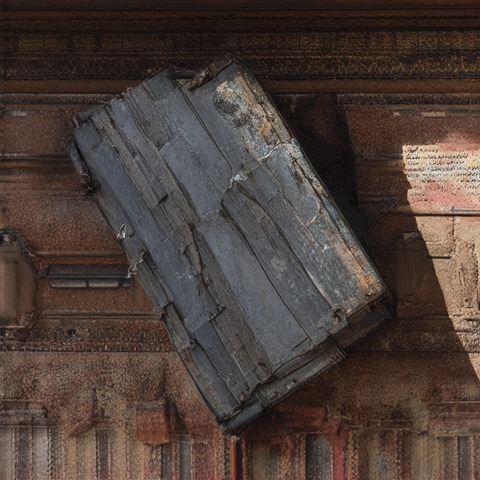}}%
\put(2.4311,0.9450){\includegraphics[width=0.6200in,height=0.6200in]{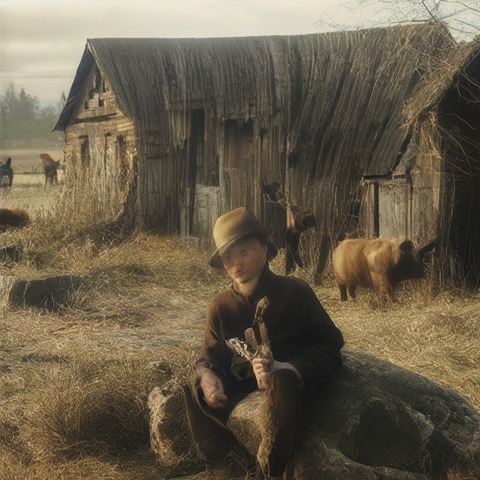}}%
\put(2.4311,0.3000){\includegraphics[width=0.6200in,height=0.6200in]{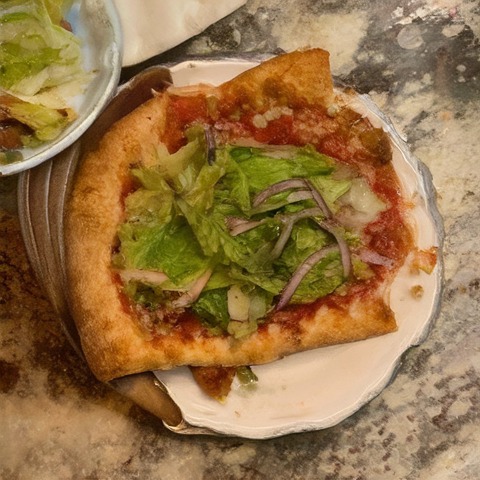}}%
\put(3.0761,1.5900){\includegraphics[width=0.6200in,height=0.6200in]{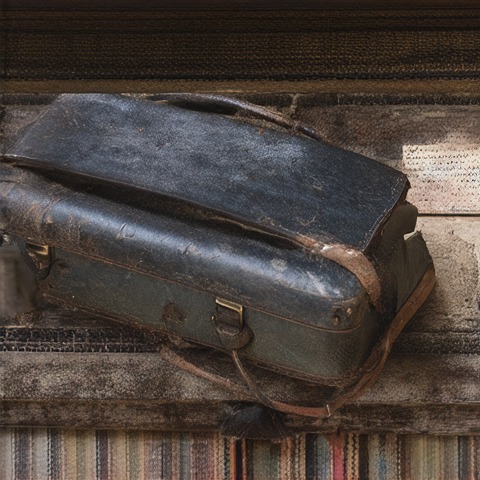}}%
\put(3.0761,0.9450){\includegraphics[width=0.6200in,height=0.6200in]{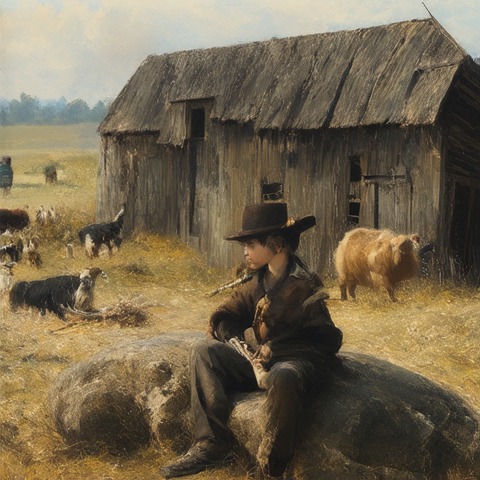}}%
\put(3.0761,0.3000){\includegraphics[width=0.6200in,height=0.6200in]{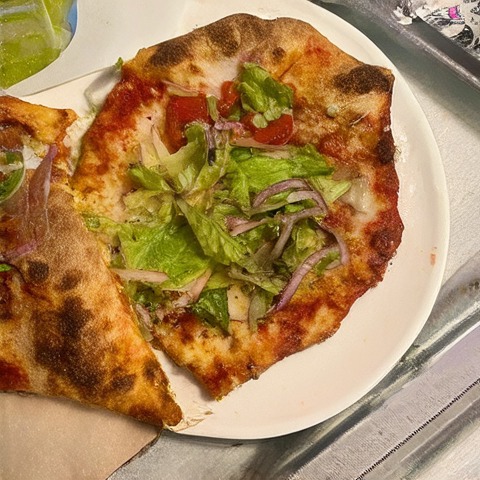}}%
\put(3.7211,1.5900){\includegraphics[width=0.6200in,height=0.6200in]{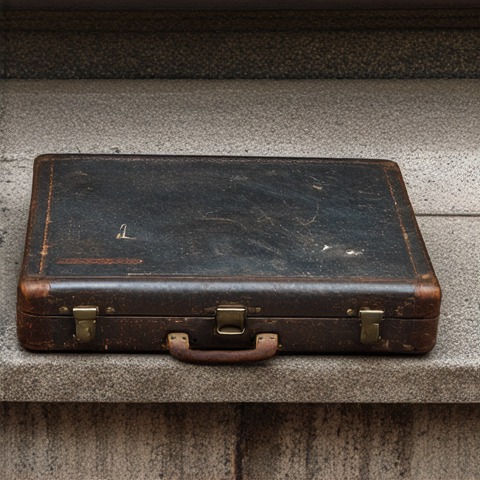}}%
\put(3.7211,0.9450){\includegraphics[width=0.6200in,height=0.6200in]{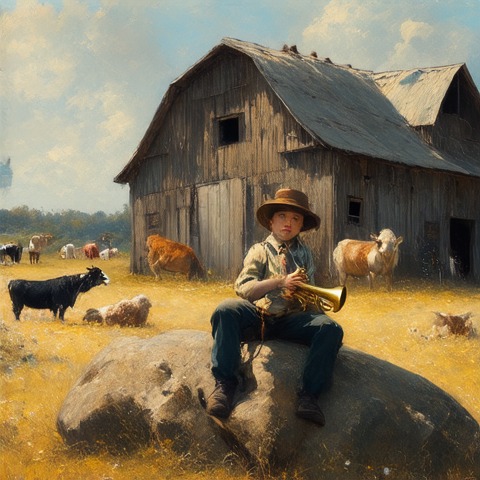}}%
\put(3.7211,0.3000){\includegraphics[width=0.6200in,height=0.6200in]{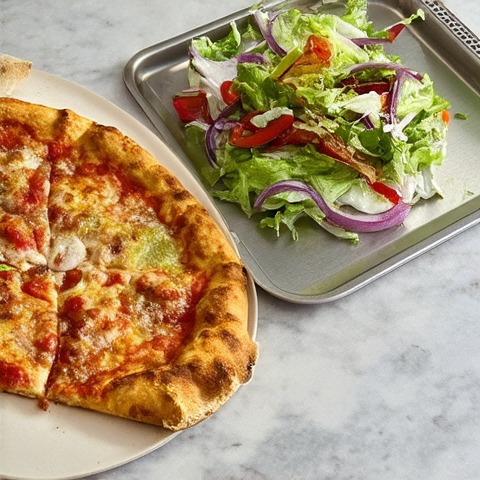}}%
\end{picture}%
{\phantomsubcaption\label{fig:wone}}{\phantomsubcaption\label{fig:teaser-samples}}{\phantomsubcaption\label{fig:random_guidance}}{\phantomsubcaption\label{fig:teaser-nested}}%

  \addtocounter{figure}{-1}%
  \captionof{figure}{We introduce a model of guidance that in closed form predicts the guidance scales used in practice. (a)~Our variance-matched scale recovers the published scales of widely used models across modalities (Spearman correlation $\rho = 0.96$). (b)~Stable Diffusion (SD)~3.5 Medium samples at the FID-optimal scale ($w=1.6$), the $\mathrm{FD}_{\mathrm{DINOv2}}$-optimal scale ($w=2.4$) and our variance-matched scale ($w=4.14$); each row uses the same prompt and initial noise. \textbf{Our interpretation suggests two hypotheses about guidance, which our experiments support.} (c)~Guidance can work in both directions: extrapolating CommonCanvas ($w<0$) or SD~1.5 ($w>1$) away from the other improves both FID and $\mathrm{FD}_{\mathrm{DINOv2}}$ over either base model, an effect not reported before, to our knowledge. (d)~Guiding a model with its earlier checkpoint needs a larger scale than with a retrained model: our model (dashed) and measured ratio of optimal scales (EDM2-XS, ImageNet-64).}
  \label{fig:teaser}
\end{center}
}]
{\renewcommand\thefootnote{\fnsymbol{footnote}}\footnotetext[1]{Equal contribution.}}

\begin{abstract}{
\looseness=-1 Guidance repels the prediction of a diffusion model away from a weaker reference prediction. It substantially improves sample quality, but its mechanism remains poorly understood.
We first find that, at the scales used in practice, guidance does not improve the prediction of the posterior mean (the model's training target); it makes the prediction worse, while samples improve.
Second, posterior-mean predictions have less variance than the data, and we find that guidance restores it. We introduce \textit{variance-matched guidance}: a closed-form rule that picks the guidance scale by matching the variance of the guided predictions to that of the data.
Without tuning, it recovers the hand-picked scales of eleven widely used models across image, video, audio, motion, and world generation; to our knowledge, no previous explanation of guidance predicts them.
Applied per noise level, it yields a schedule without tuned hyperparameters that users prefer to hand-crafted ones on SD~3.5 Medium. Third, we offer an interpretation of the mechanism behind guidance: each model makes a random error in predicting the posterior mean, and guidance rescales the difference of these errors into a sample with the posterior's covariance, which improves sampling. This view also suggests two new hypotheses that our experiments support: guidance can work with negative scales, and guiding a model with its own checkpoint needs a larger scale than guiding it with an independently trained model. \par}
\end{abstract}

\makeatletter\g@addto@macro\small{\spaceskip\z@}\g@addto@macro\footnotesize{\spaceskip\z@}\g@addto@macro\scriptsize{\spaceskip\z@}\g@addto@macro\normalsize{\spaceskip=.25em plus .15em minus .09em}\makeatother\normalsize
\section{Introduction}
\begin{figure*}[t]
    \centering
    \includegraphics[width=\textwidth]{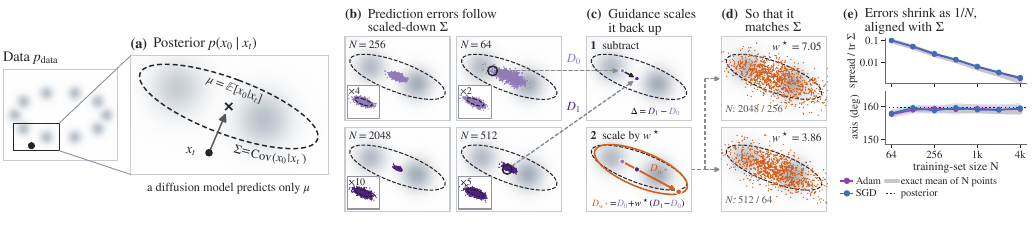}
    \caption{Guidance as variance matching, on a two-dimensional example with real trained models. Left: the data, a mixture of ten Gaussians, and one noisy input $x_t$ at $\sigma=0.4$. (a)~The posterior $p(x_0|x_t)$ has mean $\mu$ ($\times$) and covariance $\Sigma$ (dashed); a diffusion model is trained to predict only $\mu$. (b)~Predictions of $1{,}024$ models per training-set size $N$, each retrained with a new seed and training set for $40{,}000$ steps, distributed as scaled-down $\Sigma$. (c)~For a sampled pair of models, one weak ($N{=}64$) and one strong ($N{=}512$), guidance steps from $D_0$ past $D_1$. (d)~Over all pairs, the guided predictions take the total spread of $\Sigma$. (e)~The spread falls as $1/N$ and keeps the posterior's axis (dotted), for Adam, SGD and the exact posterior mean of $N$ points (gray). More details and the setup are in Appendix~\ref{app:trained-pairs}.}
    \label{fig:concept}\label{fig:story}
\end{figure*}

\label{sec:intro}

Diffusion~\citep{sohldickstein2015deep,ho2020ddpm,song2021sde} and flow-based~\citep{lipman2023flow,liu2023flow} models generate images~\citep{rombach2022ldm,esser2024sd3}, videos~\citep{ho2022video,blattmann2023svd}, audio~\citep{kong2021diffwave,liu2023audioldm} and robot actions~\citep{chi2023diffusionpolicy,black2024pi0}. In practice, their sample quality hinges on \emph{guidance}~\citep{dhariwal2021beatgans,ho2022cfg}, which combines two predictions of the clean sample from a noisy input $x_t$:
\roomydisplays
\begin{equation}
\label{eq:guided-prediction}
    D_w(x_t)=D_0(x_t)+w\bigl[D_1(x_t)-D_0(x_t)\bigr].
\end{equation}\tightdisplays
In classifier-free guidance (CFG)~\citep{ho2022cfg}, $D_1$ and $D_0$ are a network's conditional and unconditional predictions; in autoguidance~\citep{karras2024guiding}, $D_0$ is a smaller or less-trained version of $D_1$. While $w=1$ recovers $D_1$, practitioners hand-pick far larger scales, such as $w=7.5$ for Stable Diffusion (SD)~1.5~\citep{rombach2022ldm}.

Today, we lack a model that explains the effectiveness of guidance, let alone one that predicts the effective guidance scale $w$. CFG is commonly understood as sampling from a sharpened conditional distribution, but this interpretation does not seem to hold in practice~\citep{bradley2025predictor,moufad2025gibbslike}, and it does not explain why autoguidance~\cite{karras2024guiding} is effective. \citet{karras2024guiding} propose that guidance \emph{corrects errors}: the weak model makes the same errors as the strong one, only larger, so extrapolating away cancels them.
However, we find that guidance scales used in practice \emph{increase} the denoising error, and are well above those predicted by an error-reduction objective.

We instead start from a simple fact: posterior-mean predictions have less variance than the data (\cref{sec:more-moments}), and we find that guidance restores it. Matching the variance of the guided prediction to that of the data gives a closed-form scale, estimated from two models' predictions on a few noised data examples. We call this rule \emph{variance-matched guidance}. Without tuning, it recovers the hand-picked scales of eleven models across image, video, audio, motion, and world generation (\cref{fig:wone}), for example $w=7.44$ against the published $7.5$ for SD~1.5. To our knowledge, no previous explanation of guidance predicts these scales. Unlike a metric sweep, whose optimum strongly depends on the metric and feature space (\cref{fig:tuned-w-metric}), our rule is metric-agnostic, and in an informal user study, raters prefer its samples to those at the FID- and $\mathrm{FD}_{\mathrm{DINOv2}}$-optimal scales (\cref{fig:teaser-samples}, \cref{sec:exp-user}).

We then offer an interpretation of why this works. A sampler that at every DDIM step~\citep{song2021ddim} moves toward a \emph{sample of a Gaussian approximation} of the posterior $p(x_0|x_t)$, instead of its \emph{mean}, needs far fewer steps than DDIM when its moments are exact~\citep{schioppa2026covariance,akhtar2026lanczos}. Surprisingly, two trained models can provide such a sample. A diffusion model is trained to predict the posterior mean but, because training is stochastic, never recovers it exactly. If we retrain a model with different seeds and training samples, each run predicts the mean with a small, random error, and we observe (\cref{fig:story}) and explain with a simple model that these errors scatter in a scaled-down copy of the posterior covariance (\cref{fig:concept}b). A weaker model, such as the smaller network of autoguidance or the unconditional branch of CFG, scatters more widely in the same shape. Given two such models, a linear combination that keeps the mean can scale this scatter up to the full posterior covariance (\cref{fig:concept}c,d). That linear combination is the guidance formula of \cref{eq:guided-prediction}. In this view, \emph{guidance turns two estimates of the posterior mean into one prediction that carries the posterior's mean and covariance.}

Our interpretation also makes two new predictions, which our experiments support. First, the matching condition is quadratic in $w$, implying that guidance should be possible in both directions. Indeed, guiding SD~1.5 and  CommonCanvas~\citep{gokaslan2024commoncanvas}, in either direction, improves FID and $\mathrm{FD}_{\mathrm{DINOv2}}$, while every scale in between is worse than either model (\cref{fig:random_guidance}). Second, an earlier checkpoint of the strong model shares part of its training noise, so it needs a larger scale than an independently trained model (\cref{fig:teaser-nested}). Our interpretation also explains two known phenomena: CFG reduces the diversity of the distribution and hurts at high noise levels.

\noindent In summary, we:
\begin{itemize}
  \item \textbf{Derive variance-matched guidance,} a closed-form, metric-agnostic rule for the guidance scale.
  \item \textbf{Recover the published guidance scales} of eleven models across image, video, audio, motion, and world generation. Applied per noise level, it gives a schedule without tuned hyperparameters that raters prefer to hand-crafted ones.
  \item \textbf{Reinterpret guidance:} trained denoisers miss the posterior mean by errors shaped like a scaled-down posterior covariance, and guidance scales the difference between two of them up to one prediction with the posterior's mean and covariance.
  \item \textbf{Test two hypotheses that follow from our interpretation:} guidance can work with negative scales, and guiding a model with its earlier checkpoint needs a larger scale than with a retrained model.
\end{itemize}

\section{Related Work}
\label{sec:related-work}

\paragraph{Why does guidance work?}
Classifier-free guidance (CFG) improves generation by extrapolating conditional predictions away from unconditional ones \citep{ho2022cfg}. The common explanation in the literature is amplification of the conditioning signal where extrapolating away from the unconditional prediction strengthens the condition and is read as sampling from a sharpened distribution $p_w(x|c)\propto p(x|c)^w p(x)^{1-w}$ \citep{dhariwal2021beatgans,ho2022cfg}; related explanations include a predictor--corrector view \citep{bradley2025predictor}, increased conditional confidence \citep{wu2024theoretical}, and amplification of condition-specific features \citep{li2025mechanisms}.
This distributional interpretation does not seem to hold in practice: CFG does not sample from the sharpened distribution \citep{bradley2025predictor}, and the guided score is not the score of any noised distribution \citep{moufad2025gibbslike}. Conditioning is also not necessary: autoguidance \citep{karras2024guiding} and perturbation guidance \citep{ahn2024pag} extrapolate away from weaker or degraded predictions, and a learned corrector rivals autoguidance in unconditional generation, where there is no condition to amplify \citep{ahn2026correction}. These observations motivate the second explanation, due to \citet{karras2024guiding}: a weaker version of a model makes the same systematic errors as the stronger one, only larger, so extrapolation cancels them. This shared-error view \citep{kaiser2024unreasonable} extends to CFG by noting that the unconditional branch receives less training and never sees the condition (\cref{sec:shared-error}). We examine the mismatch between MSE-optimal and generation-improving scales and propose variance restoration instead.

\paragraph{Choosing the guidance scale.}
Guidance typically uses a constant scale selected through generation-metric sweeps or human evaluation, although different quality metrics can favor different weights. Prior work has explored time-dependent schedules \citep{wang2024schedulers} and interval guidance, which concentrates guidance in selected noise ranges \citep{kynkaanniemi2024interval}. Other approaches learn or optimize guidance scales through distributional objectives, control formulations, or online evaluator feedback \citep{galashov2026learn,azangulov2026adaptive,papalampidi2026dynamic,chen2026information}. CFG-Rescale \citep{lin2024noise} rescales each guided prediction to the standard deviation of the conditional one, and ReCFG \citep{xia2024recfg} computes coefficients from data to correct the mean, and APG \citep{sadat2025apg} and CFG++ \citep{chung2025cfgpp} reduce oversaturation and off-manifold drift; all still need a hand-picked $w$. In contrast, our training-free, metric-independent variance-matched criterion calibrates a constant scale and a noise-dependent schedule directly from noised data and existing model predictions, without generation sweeps, hand-designed intervals, or additional training.

\paragraph{Sampling with the posterior covariance.}
Several samplers go beyond the posterior mean. \citet{bao2022covariance} learn the reverse covariance when the mean is imperfect, and distributional diffusion models learn the full posterior instead of its mean \citep{debortoli2025distributional}. Two recent works obtain the posterior covariance from the denoiser itself through Tweedie's formula: \citet{schioppa2026covariance} add noise shaped by it to the DDIM step and improve few-step sampling, and \citet{akhtar2026lanczos} prove that matching it reduces the path-space KL divergence of DDPM from $\Omega(1/K)$ to $O(1/K^2)$ in the number of steps $K$ and sample from it with Lanczos iterations. Both need Jacobian-vector products of the denoiser at every step, one in \citet{schioppa2026covariance} and several in \citet{akhtar2026lanczos}. In our interpretation, guidance obtains a perturbation with the posterior's covariance from a second forward pass that guided models already compute.

\section{Background}
\label{sec:background}

\paragraph{Forward process and training data.}
Let $x_0\sim p_{\mathrm{data}}$ be a clean data sample, such as an image or its latent.
We write the forward diffusion process in the ``variance-exploding'' parameterization \citep{karras2022edm},
\begin{equation}
    x_t=x_0+\sigma_t\varepsilon,
    \qquad \varepsilon\sim\mathcal{N}(0,I),
    \label{eq:ve}
\end{equation}
with Gaussian noise $\varepsilon$ and its standard deviation $\sigma_t$, which grows with time $t\in[0,1]$ from $\sigma_0=0$.
We denote by $p_t$ the marginal density of the noisy sample $x_t$.
At high noise, the same $x_t$ could have been produced from many different clean samples.
Given $x_t$, these clean samples follow the posterior $p(x_0|x_t)$, with mean $\mathbb{E}[x_0|x_t]$ and covariance $\operatorname{Cov}(x_0|x_t)$.
Flow-matching models \citep{lipman2023flow,liu2023flow} interpolate instead, via $x_t=(1-t)x_0+t\varepsilon$, which agrees with~\eqref{eq:ve} after dividing $x_t$ by $1-t$, with $\sigma_t=t/(1-t)$. %
Everything in this paper applies to both families, and we use the diffusion convention.

\paragraph{Denoisers.}
A denoiser $D(x_t)$ predicts the clean sample from the noisy one. To simplify notation, we leave the noise level out of the arguments. It minimizes the squared error
\begin{equation}
\mathcal{L}(D)=\mathbb{E}_{t,x_0,\varepsilon}\bigl\|D(x_0+\sigma_t\varepsilon)-x_0\bigr\|^2,
\quad t\sim\mathcal{U}[0,1].
\label{eq:denoising-loss}
\end{equation}
Its unique minimizer is the posterior mean~\citep{karras2022edm,scarvelis2025closedform,kamb2025analytic,lukoianov2025locality}
\begin{equation}
    D^\star(x_t)=\mathbb{E}[x_0\mid x_t].
    \label{eq:tweedie}
\end{equation}
Networks often predict the noise $\varepsilon_\theta$ or, in flow matching, the velocity $v_\theta$ given by $v=\varepsilon-x_0$; each yields a denoiser
\begin{equation}
    D_\theta(x_t)=x_t-\sigma_t\varepsilon_\theta(x_t)=x_t-t\,v_\theta.
    \label{eq:denoiser-parameterizations}
\end{equation}
These parameterizations are %
equivalent with a change of variable; we work with $x_0$ predictions throughout the paper.

\paragraph{Sampling.}
DDIM \citep{song2021ddim} solves the probability-flow ODE
\begin{equation}
    \frac{\mathrm{d}x_t}{\mathrm{d}t}
    =\frac{\dot\sigma_t}{\sigma_t}\bigl(x_t-D_\theta(x_t)\bigr)
    \label{eq:probability-flow}
\end{equation}
from high noise at $t=1$ toward clean data at $t=0$. In practice, the ODE is integrated in a finite number of Euler steps:
\begin{equation}
    x_{t'}=\frac{\sigma_{t'}}{\sigma_t}\,x_t+\Bigl(1-\frac{\sigma_{t'}}{\sigma_t}\Bigr)D_\theta(x_t).
    \label{eq:ddim-step}
\end{equation}
where the step is from $t$ to $t'<t$.

\paragraph{Guidance.}
Guidance~\citep{dhariwal2021beatgans,ho2022cfg,karras2024guiding} combines a primary denoiser $D_1$ and a reference denoiser $D_0$ by \cref{eq:guided-prediction}. For CFG, $D_1$ and $D_0$ are the conditional and unconditional predictions of the same model; for autoguidance, $D_1$ is the stronger model and $D_0$ a weaker version of it. The scale $w=1$ recovers the primary predictor $D_1$, $w=0$ the reference predictor $D_0$; $w>1$ extrapolates. As the predictions above are affine in the network output, applying \cref{eq:guided-prediction} to clean-data, noise or velocity predictions gives the same guided sampler.

\section{Revisiting Prior Explanations of Guidance}
\label{revisit}
Before building our model of guidance in \cref{sec:Method}, we test two common explanations of guidance from prior work and find that neither fully matches the empirical evidence.

\begin{figure}[t]
    \centering
    \includegraphics[width=\linewidth]{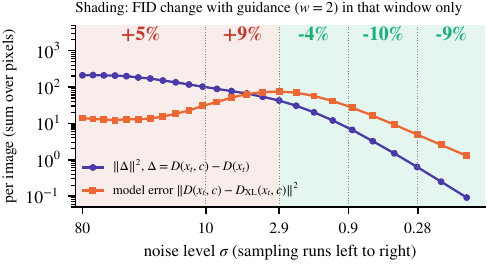}
    \caption{Where CFG helps (green), its difference $\Delta$ is dominated by model error rather than conditioning. EDM2-XS on ImageNet-64 trained with label dropout, two seeds. Per image, at the 32 sampler levels: $\|\Delta\|^2$ and the conditional prediction's error against EDM2-XL. Shading: change in FID with guidance ($w=2$) inside that window only \citep{kynkaanniemi2024interval}; green lowers it, red raises it (details in Appendix~\ref{app:cfg-regimes}).}
    \label{fig:cfg-regimes}
\end{figure}

\paragraph{CFG acts also as strong vs.\ weak, not only as conditioning.}
\label{cfg_autoguidance}
We find that much of the effectiveness of CFG may come from an autoguidance-like strong-versus-weak effect rather than from amplifying the conditioning, following the arguments of \citet{karras2024guiding} and \citet{kynkaanniemi2024interval}. The unconditional branch of CFG receives only a fraction of the training signal and is the worse fit to the data \citep{karras2024guiding}, and CFG helps only in a middle range of noise levels \citep{kynkaanniemi2024interval}, where the conditional and unconditional velocities of SD~3.5 nearly coincide (Appendix~\ref{app:vgap}). \Cref{fig:cfg-regimes} separates the two effects on ImageNet: we train EDM2-XS with label dropout, use EDM2-XL as a proxy for the posterior mean, and ask what part of the CFG difference $\Delta(x_t)=D(x_t,c)-D(x_t)$ is label information and what part is approximation error. Above $\sigma\approx5$, $\|\Delta(x_t)\|^2$ is fifteen times the model's own error, almost all of it is label information, and guidance there worsens FID (red). Below $\sigma\approx3$, $\Delta(x_t)$ falls under the model's own error, its seed-dependent share grows from $17\%$ to $42\%$, and guidance improves FID by $4$--$10\%$ (green).
Where guidance helps, the difference in prediction quality thus seems to dominate the difference in the prediction target caused by conditioning.
While both effects contribute to guidance, the experiments of \cref{sec:experiments} suggest that our model, built purely on the difference in prediction quality, also works for CFG.

\begin{figure}[t]
    \centering
    \includegraphics[width=\linewidth,trim=0 1.5pt 0 3pt,clip]{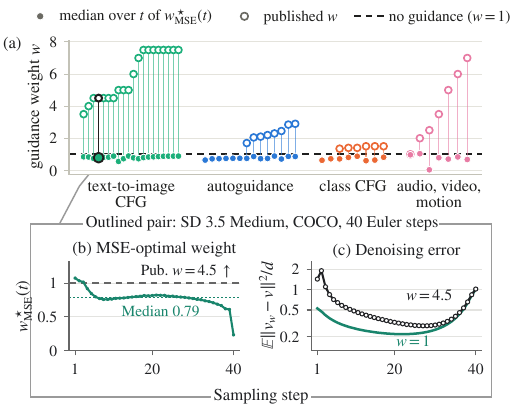}
    \caption{
    MSE-optimal scales (dots) generally lie below $w=1$, while published guidance scales (rings) favor extrapolation. (a) Each vertical stick is one model on one dataset, from its MSE-optimal scale to its published scale; sticks are grouped by guidance method, and the dashed line $w=1$ is no guidance. (b) The outlined SD~3.5 Medium pair expanded into $w^\star_{\mathrm{MSE}}(t)$ at every step: it stays below 1 at all but the first step, far below the published $w=4.5$. (c) The same pair's denoising error in velocity space, $\mathbb{E}_{x_t\sim p_t}\|v_w-v\|^2$ per latent coordinate, on a log axis: the published scale raises it above the conditional model alone ($w=1$) at every step.}
    \label{fig:wmse-published}
\vspace{-4pt}
\end{figure}

\paragraph{Guidance worsens posterior mean estimation.}
\label{sec:shared-error}
One might then expect guidance to improve the prediction itself, so that a guided prediction reaches a lower value of the training objective \eqref{eq:denoising-loss} than the primary model alone. We find the opposite in practice.
\Cref{fig:wmse-published}c evaluates this objective for SD~3.5 Medium: the published scale raises it above that of the conditional model alone at every sampling step, by up to $4\times$.
Conversely, the scale that minimizes the diffusion training objective \eqref{eq:denoising-loss} for each noise level $t$,
\begin{equation}
 w_{\mathrm{MSE}}^\star(t)=\argmin_w\mathop{\mathbb{E}}_{x_0,\varepsilon}\bigl\|D_w(x_0+\sigma_t\varepsilon)-x_0\bigr\|^2,
    \label{eq:wmse-main}
\end{equation}
which we call the \emph{MSE-optimal scale}, satisfies $w_{\mathrm{MSE}}^\star\le1$ for most of the tested models and guidance methods (\cref{fig:wmse-published}a,b).
This is at odds with the scales $w\gg1$ used in practice (details in Appendices~\ref{app:wmse-proof} and~\ref{app:wmse}).
Guidance thus makes the denoiser updates less accurate yet improves generation, so lowering the denoising error, at least on the training distribution $p_t$, does not explain it. A similar mismatch between the training objective and sample quality was observed earlier \citep{galashov2026learn,theis2016note,nichol2021improved,song2021maximum}.
We therefore interpret guidance through the whole sampling process instead of a single prediction step.

\section{Guidance as Variance Matching}
\label{sec:Method}

This section proposes that guidance edits the steps of DDIM %
to resemble those of a sampler that also captures the posterior covariance rather than the posterior mean only.

\subsection{Covariance-based generation}
\label{sec:more-moments}

At each step, a diffusion sampler starts from a noisy input $x_t$ and moves toward the denoiser's prediction of the clean sample. Ideally, the denoiser returns the posterior mean, $\E[x_0\mid x_t]$~\citep{karras2022edm}.
With this prediction, DDIM recovers the data distribution in the limit of infinitely many Euler steps, but practical samplers take a finite number of steps and use imperfect denoisers.
This procedure discards the uncertainty about which clean sample produced $x_t$.
Moreover, since
\[
    \Cov(\E[x_0|x_t]) = \Cov(x_0) - \E[\Cov(x_0|x_t)] \preceq \Cov(x_0),
\]
the marginal distribution of single-step posterior-mean predictions has less variance than the data.

\begin{figure}[t]
    \centering
    \includegraphics[width=\linewidth]{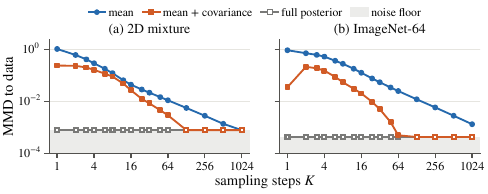}
    \vspace{-14pt}
    \caption{Samplers that use more than the posterior mean need fewer steps: mean (DDIM, \cref{eq:ddim-step}), mean + covariance (\cref{eq:M(2)}) and full posterior (\cref{eq:posterior-sampler}) on (a)~a 2D mixture of ten Gaussians and (b)~$10{,}000$ ImageNet-64 images (Appendix~\ref{sec:moments}).}
    \label{fig:moments}
\end{figure}

On the other extreme, we could make a modification to the DDIM step~\eqref{eq:ddim-step} by moving toward a random sample from the posterior $\hat x_0\sim p(x_0|x_t)$, instead of just its mean:
\begin{equation}
    x_{t'}
    = \frac{\sigma_{t'}}{\sigma_t}x_t
    + \left(1-\frac{\sigma_{t'}}{\sigma_t}\right)\hat{x}_0,
    \qquad \hat{x}_0\sim p(x_0\mid x_t).
    \label{eq:posterior-sampler}
\end{equation}
While DDIM requires solving an ODE to sample from $p_{\mathrm{data}}$, predictions of~\eqref{eq:posterior-sampler} over $x_t \sim p_t$ will match $p_{\mathrm{data}}$ exactly in just a single step.  This new sampler is unrealistic:  it requires sampling from the data distribution itself to advance to $t'$.  Hence, we consider a compromise that samples from a Gaussian approximation of $p(x_0|x_t)$ in each step:
\begin{equation}
    \widetilde{x}_0
    = \E[x_0\mid x_t] + \Cov(x_0\mid x_t)^{1/2}\varepsilon,\quad \varepsilon\sim\mathcal{N}(0,I),
\label{eq:M(2)}
\end{equation}
Unlike DDPM~\cite{ho2020ddpm}, which injects isotropic Gaussian noise, this sampler injects noise shaped by the posterior covariance $\Cov(x_0\mid x_t)$, like the covariance-aware DDIM sampler of \citet{schioppa2026covariance}, which improves few-step sampling. In \cref{fig:moments}, this \emph{mean + covariance} sampler needs far fewer steps $K$ than the \emph{mean} sampler, i.e., DDIM (\cref{eq:ddim-step}).
For DDPM, matching the full posterior covariance reduces the path-space KL divergence from $\Omega(1/K)$ to $O(1/K^2)$ \citep{akhtar2026lanczos}. In Appendix~\ref{app:m2-rate}, we prove the analogous bounds for the DDIM step: with exact posterior moments and smooth data, the error of \cref{eq:M(2)} is bounded by $O(K^{-2})$ and that of DDIM by $O(K^{-1})$.

While estimating the high-dimensional $\Cov(x_0|x_t)$ is difficult, previous work has demonstrated the benefits of modeling posterior covariance when the mean is imperfect \cite{bao2022covariance}, as well as learning the full posterior rather than its mean alone \cite{debortoli2025distributional}. Crucially, the added variability should follow the \emph{posterior} covariance; the right variance with the wrong shape does not help. On SD~3.5 Medium, adding Gaussian noise that restores the missing variance but ignores $x_t$, isotropic or proportional to the data covariance $\Cov(x_0)$, worsens every metric relative to adding none (Appendix~\ref{app:noise-control}).

Next, we relate guidance to the sampler of \cref{eq:M(2)}.

\subsection{Guidance as Sampling from the Covariance}
\label{sec:ideal_denoiser}\label{sec:second-moment}

\looseness=-1 We now show that guidance approximates such a sampler: it combines two estimates of the posterior mean into one prediction with the posterior's mean and covariance. We characterize a network's errors by their distribution over retrainings, which we first show empirically.

In large-scale training, a finite number of training steps exposes a network to only a finite number of samples near any $x_t$. To control this sample count directly in a low-dimensional setting, \cref{fig:story} varies the training-set size $N$ across $1{,}024$ networks trained on a mixture of Gaussians with different seeds. Across $N$, their predictions at a fixed $x_t$ spread along the posterior's axis, in a scaled-down copy of its covariance that shrinks as $1/N$, just like the exact posterior mean of $N$ points (\cref{fig:story}b,e). Such plots are not possible in high dimensions, but Appendix~\ref{app:vgap} shows that for SD~3.5 Medium the conditional and unconditional predictions vary along the same directions in the noise range where CFG helps \citep{kynkaanniemi2024interval}.

We propose that these observations can be understood through a simplified lens. Many clean samples could have produced the same $x_t$, so the network is shown several targets $x_0$ for nearly the same input, and the $L_2$ loss pulls its prediction toward the mean.
The set of targets the model sees for a given $x_t$ is random, since it depends on the drawn training set and on the order of training.
We therefore analyze the case where the network predicts exactly the mean of the targets it has observed for $x_t$, and we model these targets as $m(x_t)$ i.i.d.\ samples $x_0\sim p(\cdot|x_t)$ (Appendix~\ref{app:bootstrap} derives this for a model seeing $N$ fresh training pairs). Here $m(x_t)$ is the effective number of training samples seen for this $x_t$ and grows with $N$.
Let $p_N$ be the distribution of the resulting denoisers $D_N$ over training runs and training sets $S_N$ of size $N$ drawn i.i.d.\ from $p_{\mathrm{data}}$.
Averaging keeps the mean and divides the covariance by $m(x_t)$:
\begin{align}
    \E_{D_N\sim p_N}[D_N(x_t)]
    &= \E[x_0\mid x_t],\\
    \Cov_{D_N\sim p_N}(D_N(x_t))
    &= \frac{1}{m(x_t)}\Cov(x_0\mid x_t).
    \label{eq:denoiser-covariance}
\end{align}

Now consider two independent denoisers, a weaker model $D_0\sim p_{N_0}$ and a stronger model $D_1\sim p_{N_1}$, with $N_1>N_0$ and effective numbers $m_1(x_t)>m_0(x_t)$.
They model the strong and weak denoisers of autoguidance and, following \cref{cfg_autoguidance}, the two branches of CFG.
For their affine combination
\begin{equation}
    D_w(x_t) = w D_1(x_t)+(1-w) D_0(x_t),
    \label{eq:dw_derivation}
\end{equation}
the mean remains $\E[x_0|x_t]$, while independence gives\roomydisplays
\begin{equation}
    \medmuskip=3mu\thickmuskip=4mu
    \underset{\substack{D_0\sim p_{N_0}\\D_1\sim p_{N_1}}}{\Cov}\!(D_w(x_t))
    = \underbrace{\left(\tfrac{w^2}{m_1(x_t)}
       +\tfrac{(1-w)^2}{m_0(x_t)}\right)}_{c(w,m_1,m_0)}
       \Cov(x_0|x_t).
    \label{eq:cov-delta}
\end{equation}\tightdisplays
Thus, choosing $w^\star(x_t)$ so that $c(w^\star(x_t),m_1,m_0)=1$ makes the prediction $D_{w^\star(x_t)}(x_t)$ (written $D_{w^\star}(x_t)$) match both the mean and covariance of $p(x_0\mid x_t)$ over resampling denoisers from $p_{N_0},p_{N_1}$.
Guided pairs indeed take the posterior's size in \cref{fig:story}d, as the model predicts.

To summarize, the model above suggests an impractical way to implement the mean + covariance sampler \cref{eq:M(2)}.
At each step, we independently sample two training sets, train $D_0$ and $D_1$ on them, and compute $w^\star$ from \eqref{eq:cov-delta}.
Replacing the posterior sample in \eqref{eq:posterior-sampler} with the guided prediction gives
\begin{equation}
x_{t'} = \frac{\sigma_{t'}}{\sigma_t}x_t
+ \left(1-\frac{\sigma_{t'}}{\sigma_t}\right) D_{w^\star}(x_t),
\label{eq:guided-sampler}
\end{equation}
By the central limit theorem, $D_{w^\star}(x_t)$ is also approximately Gaussian when $m_0(x_t)$ and $m_1(x_t)$ are large.
Remarkably, this is the standard guidance formula~\eqref{eq:guided-prediction}.

\subsection{Method: Variance-Matched Guidance (VMG)}
\label{sec:vm-guidance}
In practice, we cannot resample training sets as in \cref{eq:guided-sampler}: we observe one pair $(D_0,D_1)$ at many $x_t$, and we know neither $m_0,m_1$ nor $\Cov(x_0|x_t)$.
We assume that $D_0$ and $D_1$ are independent of each other and make independent errors at distant inputs, since a network's error at $x_t$ depends on the targets it saw near $x_t$. Averaging over inputs then stands in for averaging over retrainings. We also assume that $m_0,m_1$ vary little across $x_t$ at a given $t$, and revisit these assumptions in \cref{sec:exp-hypotheses}.
Since $D_{w^\star}$ has mean $\E[x_0|x_t]$ and covariance $\Cov(x_0|x_t)$, letting $p=(p_{N_0},p_{N_1})$, we have
\predisplaypenalty=0
\begin{align}
    &\Cov_{x_t}\bigl(D_{w^\star}(x_t)\bigr) \approx \Cov_{p,x_t}(D_{w^\star}) \label{eq:total-cov} \\
    &\quad= \Cov_{x_t}\bigl(\E_{p}[D_{w^\star}|x_t]\bigr)+\E_{x_t}\bigl[\Cov_{p}(D_{w^\star}|x_t)\bigr] \nonumber \\
    &\quad= \Cov(\E[x_0|x_t])+\E[\Cov(x_0|x_t)] = \Cov(x_0). \nonumber
\end{align}\predisplaypenalty=10000
The first equality is our assumption, the second the law of total covariance, and the last the same law applied to $x_0$ (\cref{sec:more-moments}). We therefore estimate $w$ by matching the covariance of the guided prediction over inputs to that of the data, which only restores the variance missing from posterior-mean predictions. Since in our model the guided covariance already has the shape of the posterior's and $w$ only sets its size (\cref{eq:cov-delta}), it is enough to match its trace. We therefore, for each $t$, solve for $w(t)$:
\begin{equation}
    \Var_{x_t\sim p_t}[D_w(x_t)]
    = \Var_{x_0\sim p_{\mathrm{data}}}[x_0],
    \label{eq:varmatch_t}
\end{equation}
we call this method \textit{Variance-Matched Guidance}.
We draw $x_t=x_0+\sigma_t\varepsilon$ from data samples $x_0$ and, when $D_1$ is a conditional model, pass it the condition that comes with $x_0$. The left side is quadratic in $w$; with
\begin{equation}
\begin{aligned}
a(t) &= \Var_{x_t\sim p_t}[D_1-D_0],\\
b(t) &= \Var_{x_t\sim p_t}[D_0]-\operatorname{tr}\bigl[\Cov_{x_t\sim p_t}(D_0,D_1)\bigr],\\
c(t) &= \Var_{x_t\sim p_t}[D_0]-\Var_{x_0\sim p_{\mathrm{data}}}[x_0],
\end{aligned}
\label{eq:vm-coefficients}
\end{equation}
\cref{eq:varmatch_t} becomes $a(t)w(t)^2-2b(t)w(t)+c(t)=0$ with roots:
\begin{equation}
    w_{\mathrm{VM}}^\star(t)
    =\frac{ b(t)\pm\sqrt{ b(t)^2-a(t)c(t)}}{a(t)}.
    \label{eq:wstar}
\end{equation}
When $ c(t)<0$, the two roots have opposite signs. As a result, the rule predicts that guidance can work in both directions: the positive root extrapolates away from the weak model, as usual, and the negative root extrapolates the other way. This does not hold for every pair, but we were surprised to find that for some pairs guidance works in both directions, as we show in \cref{fig:random_guidance} and \cref{sec:exp-hypotheses}.
To our knowledge, this has not been reported before. Elsewhere we use the positive root.

\paragraph{Global scale.}
Most published models use a fixed scale across time. We can use our method to output such scales as well by solving:
\begin{equation}
    \E_{t\sim\mathcal{U}[0,1]}\Var_{x_t\sim p_t}[D_w(x_t)]
    = \Var_{x_0\sim p_{\mathrm{data}}}[x_0],
    \label{eq:varmatch}
\end{equation}
(Appendix~\ref{app:objective} ablates other ways to choose a constant scale). We define the coefficients'  averages over $t\sim\mathcal{U}[0,1]$, $\bar a$, $\bar b$ and $\bar c$, hence we get \begin{equation}
    w_{\mathrm{VM}}^\star
    =\frac{ \bar b\pm\sqrt{\bar b^2-\bar a \bar c}}{\bar a}.
    \label{eq:wconstant}
\end{equation}

Matching the full covariance instead of the variance trace  yields close $w$, as the theory predicts (Appendix~\ref{app:fullcov}).

\section{Experiments}
\label{sec:experiments}
\begin{figure}[t]
    \centering
    \includegraphics{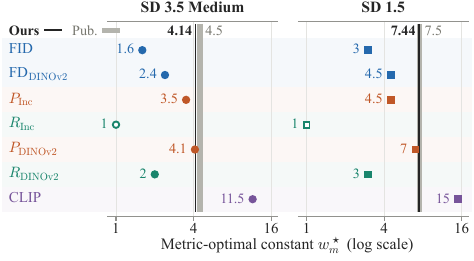}
    \vspace{-1pt}
    \caption{\textbf{The best guidance scale depends on the metric.} Each marker is the constant scale that is best for one metric; open markers indicate the lower end of the searched grid ($w=1$). Lines mark our constant (black) and the published scale (gray).}
    \label{fig:tuned-w-metric}
    \vspace{1pt}
\end{figure}
This section evaluates our rule for the guidance scale (\cref{eq:wconstant,eq:wstar}). We demonstrate that variance-matched guidance (VMG) (1) matches the published guidance scales across models, (2) is preferred by humans in an informal user study, %
 and (3) performs well \emph{across} metrics.

\paragraph{Models.}
We evaluate eleven pretrained CFG models across five modalities: text-to-image (SD~1.5 \citep{rombach2022ldm}, SD~3.5 Medium and Large \citep{esser2024sd3}, FLUX.2 [klein] 4B \citep{bfl2026flux2klein}), video (SVD \citep{blattmann2023svd}, Wan2.1 \citep{wan2025wan}, CogVideoX-2B \citep{yang2025cogvideox}), a video world model (Cosmos-Predict2.5 \citep{nvidia2025cosmos}), audio (AudioLDM~2 \citep{liu2024audioldm2}, Stable Audio Open \citep{evans2025sao}) and motion (MDM \citep{tevet2023mdm}). Each model keeps its released sampler and step count; compared settings share conditioning and initial noise. Scales are fitted from forward-noised calibration data, disjoint from the evaluation data, without sampling (Appendices~\ref{app:impl} and~\ref{app:ncalib}).

\paragraph{Baselines and metrics.}
The main challenge of evaluating guidance scales is that different metrics lead to different optima.
We use each modality's distribution distance, precision, recall and alignment score (Appendix~\ref{app:impl}).
In \cref{fig:tuned-w-metric}, the best constant for SD~3.5 Medium ranges from $w=1$ for recall~\citep{kynkaanniemi2019improved} in Inception-v3 features~\citep{szegedy2016rethinking} to $11.5$ for CLIP.

\subsection{Results}
\label{sec:exp-metrics}
\label{sec:exp-crossmetric}

\paragraph{VMG recovers the published scales.}
Most models come with a published scale.
How it was chosen is rarely documented, and \cref{fig:tuned-w-metric} and \cref{tab:tuned-w} show that it matches no metric's optimum (Appendix~\ref{app:frontier}).
We hypothesize that, in practice, published guidance scales are mostly picked by human judgment.
In \cref{fig:wone}, our constant predicts the published scales across image, video, audio, motion and world models, without any sampling: on SD~3.5 Medium the fit costs less than generating the evaluation set once, and far less than the sweep behind \cref{fig:tuned-w-metric} (Appendix~\ref{app:impl}). It also needs few examples: the scale's standard deviation is $0.06$ from $256$ photos on SD~3.5 Medium (Appendix~\ref{app:ncalib}).

\paragraph{Users prefer VMG in a blind study.}
\label{sec:exp-user}
We ran an informal blind pairwise study on SD~3.5 Medium: a rater sees two images from the same held-out COCO caption and initial noise, generated by two of seven settings (the two metric-tuned constants, our constant, our schedule and the three hand-designed schedules of \cref{tab:pub-sched}), and picks the better one or a tie. We omit the published constant ($w=4.5$), within $10\%$ of ours and $3\%$ on every metric (\cref{tab:pub-sched}). In \cref{tab:user-study}, Bradley--Terry ratings of $1{,}458$ votes from $41$ people prefer our constant to both tuned constants and our schedule to all three hand-designed ones. Captions were drawn at random; the protocol and the study are in Appendix~\ref{app:user-study}.

\paragraph{VMG is competitive with grid search on all but the tuned metric.}
In \cref{tab:tuned-w}, we score the constants tuned via grid search to each metric of \cref{fig:tuned-w-metric}, the published constant and ours on all metrics. Each tuned constant wins on its own metric but often loses on others, a metric-specific trade-off. Without any grid search, ours nearly matches the scale tuned to DINOv2 precision ($4.14$ vs.\ $4.1$) and outperforms the other tuned constants on most of their non-target metrics. EDM2~\citep{karras2024edm2,karras2024guiding} chose its published scales the same way, by a grid search for the lowest FID or FD$_\mathrm{DINOv2}$ under both CFG and autoguidance. Like the FID- and FD$_\mathrm{DINOv2}$-tuned constants here, which users rated below ours (\cref{tab:user-study}), these scales are smaller than our constant (Appendix~\ref{app:edm2-fail}; samples in Appendix~\ref{app:samples}).

\begin{table}[t]
\centering
\caption{Informal blind pairwise user study on SD 3.5 Medium ($1{,}458$ votes from $41$ people): Bradley--Terry ratings on the Elo scale, $\pm$ a $95\%$ bootstrap interval over raters. One fit covers all votes, so blocks are comparable. Hand-designed schedules use the published $w=4.5$; head-to-head results in Appendix~\ref{app:user-study}.}
\label{tab:user-study}
\renewcommand{\arraystretch}{0.9}
\footnotesize
\setlength{\tabcolsep}{3pt}
\begin{adjustbox}{max width=\linewidth}
\begin{tabular}[t]{@{}llr@{\,}c@{\,}l@{}}
\toprule
\multicolumn{5}{@{}c@{}}{\emph{Constant weight}} \\
\cmidrule{1-5}
Setting & $w$ & \multicolumn{3}{c@{}}{Elo} \\
\midrule
\textbf{Ours} & {\boldmath4.14} & \textbf{1048} & {\boldmath$\pm$} & \textbf{36} \\
\multicolumn{5}{@{}l}{\emph{Grid search over:}} \\
\quad FD$_\mathrm{DINOv2}$ & 2.4 & 985 & $\pm$ & 40 \\
\quad FID & 1.6 & 874 & $\pm$ & 44 \\
\bottomrule
\end{tabular}\hspace{14pt}%
\begin{tabular}[t]{@{}llr@{\,}c@{\,}l@{}}
\toprule
\multicolumn{5}{@{}c@{}}{\emph{Schedule}} \\
\cmidrule{1-5}
Setting & $w$ & \multicolumn{3}{c@{}}{Elo} \\
\midrule
\textbf{Ours} & {\boldmath$w(\sigma)$} & \textbf{1088} & {\boldmath$\pm$} & \textbf{43} \\
TV-CFG & 4.5 & 1054 & $\pm$ & 39 \\
Interval & 4.5 & 992 & $\pm$ & 35 \\
Linear & 4.5 & 959 & $\pm$ & 40 \\
\bottomrule
\end{tabular}

\end{adjustbox}
\vspace{-4pt}
\end{table}

\begin{table}[t]
\centering
\caption{Constant scales tuned to individual metrics on SD 3.5 Medium versus the published constant (Pub.) and ours (Ours). Bold marks each row's target metric. Each tuned cell is shaded against Ours: \swatch{vmbetter}{green} where ours is better, \swatch{vmworse}{red} where the tuned scale is (faint below $1\%$, light for $1$--$5\%$, full beyond $5\%$).} %
\label{tab:tuned-w}
\renewcommand{\arraystretch}{0.9}
\footnotesize
\setlength{\tabcolsep}{2pt}
\begin{adjustbox}{max width=\linewidth}
\begin{tabular}{llrrrrr}
\toprule
Tuned to & $w_m^\star$ & FID $\downarrow$ & FD$_\mathrm{DINOv2}$ $\downarrow$ & P$_\mathrm{DINOv2}$ $\uparrow$ & P$_\mathrm{Inc}$ $\uparrow$ & CLIP $\uparrow$ \\
\midrule
FID & 1.6 & \cellcolor{vmworse}\textbf{13.40} & \cellcolor{vmbetterlt}251.4 & \cellcolor{vmbetter}0.665 & \cellcolor{vmbetter}0.623 & \cellcolor{vmbetterlt}0.2613 \\
FD$_\mathrm{DINOv2}$ & 2.4 & \cellcolor{vmworse}15.60 & \cellcolor{vmworse}\textbf{227.1} & \cellcolor{vmbetterlt}0.722 & \cellcolor{vmbetterxl}0.678 & \cellcolor{vmbetterlt}0.2680 \\
P$_\mathrm{DINOv2}$ & 4.1 & \cellcolor{vmworsexl}20.10 & \cellcolor{vmworsexl}239.4 & \cellcolor{vmworsexl}\textbf{0.751} & \cellcolor{vmworsexl}0.686 & \cellcolor{vmworsexl}0.2714 \\
P$_\mathrm{Inc}$ & 3.5 & \cellcolor{vmworse}18.88 & \cellcolor{vmworselt}233.1 & \cellcolor{vmbetterlt}0.740 & \cellcolor{vmworselt}\textbf{0.694} & \cellcolor{vmbetterxl}0.2708 \\
CLIP & 11.5 & \cellcolor{vmbetter}23.26 & \cellcolor{vmbetter}280.5 & \cellcolor{vmbetterlt}0.722 & \cellcolor{vmbetter}0.614 & \cellcolor{vmworselt}\textbf{0.2740} \\
\midrule
Pub. & 4.5 & 20.58 & 242.4 & 0.744 & 0.685 & 0.2715 \\
\midrule
Ours & 4.14 & 20.15 & 240.0 & 0.748 & 0.684 & 0.2713 \\
\color{black!60}Ours ($w(\sigma)$) & \color{black!60} \ \ \ -- & \color{black!60}20.13 & \color{black!60}224.0 & \color{black!60}0.765 & \color{black!60}0.667 & \color{black!60}0.2749 \\
\bottomrule
\end{tabular}
\end{adjustbox}
\vspace{-4pt}
\end{table}

\paragraph{A noise-dependent VMG scale is competitive with the published constant.}
\label{sec:exp-schedules}
We now fit the scale at every noise level by the same rule (\cref{eq:wstar}) and compare it with each model's published constant. In \cref{tab:pub-all}, with no tuning, the schedule improves the distribution metrics on about half of the models, most on SVD and Wan2.1, stays close to the published setting on most others, and on CogVideoX-2B and AudioLDM~2 trades distribution distance for alignment.

\begin{table}[!b]
\centering
\caption{Published guidance (Pub.) versus our variance-matched schedule (Ours). Ours is shaded and colored by its relative change against Pub.: \textcolor{vmgood}{green} where better, \textcolor{vmbad}{red} where worse, \textbf{bold} beyond $5\%$, black within $1\%$. Protocols in Appendix~\ref{app:impl}.}
\label{tab:pub-all}
\scriptsize\setlength{\tabcolsep}{1.5pt}\renewcommand{\arraystretch}{0.85}\setlength{\lineskip}{0pt}
\def\namew{0pt}\def\trynm#1{\setbox0\hbox{#1}\ifdim\wd0>\namew\relax\edef\namew{\the\wd0}\fi}%
\trynm{SD 1.5}\trynm{SD 3.5 Medium}\trynm{SD 3.5 Large}\trynm{FLUX.2 [klein] 4B}\trynm{SVD}\trynm{Wan2.1 1.3B}\trynm{CogVideoX-2B}\trynm{AudioLDM 2}\trynm{Stable Audio Open}\def\nm#1{\makebox[\namew][l]{#1}}%
\begin{tabular*}{\linewidth}{@{}l@{\ \,}l@{\hspace{3pt}\extracolsep{\fill}}rrrrrrr@{}}
\toprule
\multicolumn{2}{@{}l}{\textit{Text-to-image}} & \multicolumn{3}{c}{Inception} & \multicolumn{3}{c}{DINOv2} & \\
\cmidrule(lr){3-5}\cmidrule(lr){6-8}
 & & \hd{FID$\downarrow$} & \hd{P$\uparrow$} & \hd{R$\uparrow$} & \hd{FD$\downarrow$} & \hd{P$\uparrow$} & \hd{R$\uparrow$} & \hd{CLIP$\uparrow$} \\
\nm{SD 1.5} & Pub. & 17.51 & 0.629 & 0.482 & 233.28 & 0.691 & 0.597 & 0.2690 \\
\oursrow & Ours & {\color{vmgood}17.27} & {\bfseries\color{vmbad}0.585} & {\bfseries\color{vmgood}0.513} & {\color{vmbad}239.41} & {\color{vmbad}0.663} & {\bfseries\color{vmbad}0.567} & 0.2699 \\
\nm{SD 3.5 Medium} & Pub. & 20.58 & 0.685 & 0.461 & 242.4\dz & 0.744 & 0.555 & 0.2715 \\
\oursrow & Ours & {\color{vmgood}20.13} & {\color{vmbad}0.667} & {\color{vmbad}0.440} & {\bfseries\color{vmgood}224.0}\dz & {\color{vmgood}0.765} & {\color{vmgood}0.578} & {\color{vmgood}0.2749} \\
\nm{SD 3.5 Large} & Pub. & 21.00 & 0.595 & 0.448 & 322.99 & 0.710 & 0.438 & 0.2690 \\
\oursrow & Ours & {\color{vmgood}20.33} & {\color{vmbad}0.570} & {\color{vmgood}0.470} & {\bfseries\color{vmgood}301.05} & {\color{vmgood}0.721} & {\color{vmgood}0.453} & 0.2705 \\
\nm{FLUX.2 [klein] 4B} & Pub. & 19.48 & 0.627 & 0.536 & 197.9\dz & 0.754 & 0.779 & 0.2746 \\
\oursrow & Ours & {\color{vmbad}19.69} & {\color{vmbad}0.610} & {\color{vmgood}0.545} & {\color{vmgood}194.0}\dz & 0.747 & 0.776 & 0.2745 \\
\end{tabular*}

\begin{tabular*}{\linewidth}{@{}l@{\ \,}l@{\hspace{3pt}\extracolsep{\fill}}rrrrrr@{}}
\midrule
\multicolumn{2}{@{}l}{\textit{Video}} & \hd{FVD$\downarrow$} & \hd{P$\uparrow$} & \hd{R$\uparrow$} & \hd{Dens.$\uparrow$} & \hd{Cov.$\uparrow$} & \hd{CLIP-T$\uparrow$} \\
\nm{SVD} & Pub. & 141.2 & 0.804 & 0.817 & 0.883 & 0.871 & \na \\
\oursrow & Ours & {\bfseries\color{vmgood}113.7} & {\color{vmgood}0.816} & {\color{vmgood}0.838} & {\bfseries\color{vmgood}0.972} & {\color{vmgood}0.894} & \na \\
\nm{Wan2.1 1.3B} & Pub. & 261.0 & 0.589 & 0.635 & 0.770 & 0.754 & 30.35 \\
\oursrow & Ours & {\bfseries\color{vmgood}220.0} & {\color{vmgood}0.606} & {\bfseries\color{vmgood}0.677} & {\color{vmbad}0.738} & {\color{vmgood}0.765} & 30.25 \\
\nm{CogVideoX-2B} & Pub. & 236.3 & 0.589 & 0.701 & 0.829 & 0.737 & 30.31 \\
\oursrow & Ours & {\bfseries\color{vmbad}262.1} & {\bfseries\color{vmbad}0.555} & {\bfseries\color{vmgood}0.752} & {\bfseries\color{vmbad}0.592} & {\bfseries\color{vmbad}0.685} & {\color{vmgood}30.66} \\
\end{tabular*}

\begin{tabular*}{\linewidth}{@{}l@{\ \,}l@{\hspace{3pt}\extracolsep{\fill}}rrrrr@{}}
\midrule
\multicolumn{2}{@{}l}{\textit{Audio}} & \hd{FAD$^\ddagger$$\downarrow$} & \hd{KL$\downarrow$} & \hd{CLAP$\uparrow$} & \hd{P$\uparrow$} & \hd{R$\uparrow$} \\
\nm{AudioLDM 2} & Pub. & 2.00 & 1.47\dz & 0.241\dz & 0.825 & 0.201 \\
\oursrow & Ours & {\bfseries\color{vmbad}2.46} & {\bfseries\color{vmgood}1.29}\dz & {\bfseries\color{vmgood}0.283}\dz & {\bfseries\color{vmbad}0.721} & {\bfseries\color{vmbad}0.167} \\
\nm{Stable Audio Open} & Pub. & 165.18 & 0.549 & 0.4213 & \na & \na \\
\oursrow\color{black!60}\,$^\ddagger$FD$_\mathrm{OpenL3}$ & Ours & 164.95 & {\bfseries\color{vmgood}0.513} & 0.4182 & \na & \na \\
\end{tabular*}

{\setlength{\tabcolsep}{1pt}\begin{tabular*}{\linewidth}{@{}l@{\ \,}l@{\hspace{3pt}\extracolsep{\fill}}rr@{\hspace{2.5pt}}!{\color{black!35}\vrule}@{\hspace{2.5pt}}lrrr@{}}
\midrule
\multicolumn{2}{@{}l}{\textit{World model}} & \hd{FVD$\downarrow$} & \hd{PSNR$\uparrow$} & \textit{Motion} & \hd{FID$\downarrow$} & \hd{MM-Dist$\downarrow$} & \hd{MMod.$\uparrow$} \\
Cosmos-Predict2.5 & Pub. & 62.6 & 25.50 & MDM & 0.536 & 3.115 & 1.54 \\
\oursrow & Ours & 63.0 & 25.49 & & {\color{vmgood}0.523} & {\color{vmbad}3.158} & {\bfseries\color{vmgood}1.72} \\
\bottomrule
\end{tabular*}}
\end{table}

A line of work replaces the constant scale with a hand-designed function of the noise level, and other methods learn the schedule with extra training or evaluator feedback \citep{galashov2026learn,azangulov2026adaptive,papalampidi2026dynamic}. Among the hand-designed schedules, interval guidance \citep{kynkaanniemi2024interval} guides only within a range of noise levels, the linear schedule of \citet{wang2024schedulers} raises the weight linearly over sampling, and TV-CFG \citep{jin2026stagewise} raises it toward the middle of sampling and lowers it again. Each needs a base weight $w$, and interval guidance also the interval, whereas our schedule has none to tune. On SD~3.5 Medium, our schedule is preferred to all three in the user study (\cref{tab:user-study}). In \cref{tab:pub-sched}, we compare them on the metrics: out of the box, our schedule gives the best DINOv2 precision and CLIP score of all settings, while interval and linear guidance reach lower FID, FD$_\mathrm{DINOv2}$ and recall. Its higher mean scale ($7.82$) may account for part of this gain. Setting their base weight to our constant instead of the published one lowers the FID of all three, so our weight is also search-free for them.
\begin{table}[t]
\centering
\caption{Hand-designed schedules on SD 3.5 Medium, Interval \citep{kynkaanniemi2024interval}, TV-CFG \citep{jin2026stagewise} and Linear \citep{wang2024schedulers}, at the published weight (Pub.) and at our constant (Ours). Mean $w$ is over sampling steps. Our rows are shaded and colored against Pub.: \textcolor{vmgood}{green} where better, \textcolor{vmbad}{red} where worse, \textbf{bold} beyond $1\%$. Underline marks the best per column.}
\label{tab:pub-sched}
\scriptsize\setlength{\tabcolsep}{2pt}\renewcommand{\arraystretch}{0.9}
\begin{adjustbox}{max width=\linewidth}
\begin{tabular}{llrrrrrrrr}
\toprule
 & & & \multicolumn{3}{c}{Inception} & \multicolumn{3}{c}{DINOv2} & \\
\cmidrule(lr){4-6}\cmidrule(lr){7-9}
\multicolumn{2}{l}{Schedule} & \hd{mean $w$} & \hd{FID$\downarrow$} & \hd{P$\uparrow$} & \hd{R$\uparrow$} & \hd{FD$\downarrow$} & \hd{P$\uparrow$} & \hd{R$\uparrow$} & \hd{CLIP$\uparrow$} \\
\midrule
Constant & Pub. & 4.50 & 20.58 & \best{0.685} & 0.461 & 242.4 & 0.744 & 0.555 & 0.2715 \\
\rowcolor{tours} & Ours & 4.14 & {\bfseries\color{vmgood}20.15} & {\color{vmbad}0.684} & {\bfseries\color{vmbad}0.454} & {\color{vmgood}240.0} & {\color{vmgood}0.748} & {\bfseries\color{vmgood}0.570} & {\color{vmbad}0.2713} \\
Interval & Pub. & 4.02 & 15.80 & 0.632 & 0.506 & 222.9 & 0.734 & 0.588 & 0.2687 \\
\rowcolor{tours} & Ours & 3.71 & {\bfseries\color{vmgood}15.44} & {\color{vmgood}0.633} & {\color{vmgood}0.510} & {\color{vmgood}221.3} & 0.734 & {\bfseries\color{vmgood}0.604} & {\color{vmbad}0.2684} \\
TV-CFG & Pub. & 4.82 & 19.88 & 0.665 & 0.453 & 237.1 & 0.740 & 0.571 & 0.2713 \\
\rowcolor{tours} & Ours & 4.43 & {\bfseries\color{vmgood}19.20} & {\color{vmgood}0.669} & {\bfseries\color{vmgood}0.463} & {\bfseries\color{vmgood}233.8} & {\color{vmgood}0.741} & {\color{vmbad}0.570} & {\color{vmbad}0.2710} \\
Linear & Pub. & 3.28 & 13.91 & 0.652 & 0.541 & \best{213.5} & 0.733 & \best{0.621} & 0.2673 \\
\rowcolor{tours} & Ours & 3.05 & \best{{\bfseries\color{vmgood}13.60}} & {\color{vmbad}0.646} & \best{{\color{vmgood}0.544}} & {\color{vmbad}214.3} & {\color{vmbad}0.728} & \best{0.621} & {\color{vmbad}0.2667} \\
\midrule
\rowcolor{tours} Our schedule & $w(\sigma)$ & 7.82 & 20.13 & 0.667 & 0.440 & 224.0 & \best{0.765} & 0.578 & \best{0.2749} \\
\bottomrule
\end{tabular}
\end{adjustbox}
\vspace{-8pt}
\end{table}

\section{What Variance Matching Predicts}
\label{sec:exp-hypotheses}
Our interpretation makes predictions that, to our knowledge, no other explanation of guidance makes. We test two of them and use the same lens on two known effects.

\paragraph{Guidance works in both directions.}
The matching condition is quadratic in $w$ (\cref{eq:wstar}), and its two roots often have opposite signs: the positive root extrapolates the strong model away from the weak one, the negative root extrapolates the weak model away from the strong one and restores the same covariance. Error correction would suggest a gain in one direction only, away from the model that makes the larger error. We test this with two text-to-image models that share the latent space of SD~1.5 but were trained on different data: SD~1.5 itself \citep{rombach2022ldm} and CommonCanvas \citep{gokaslan2024commoncanvas}, trained on Creative-Commons images. With $D_0$ the CommonCanvas prediction and $D_1$ the SD~1.5 prediction in \cref{eq:guided-prediction}, $w>1$ guides SD~1.5 away from CommonCanvas and $w<0$ guides CommonCanvas away from SD~1.5. In \cref{fig:random_guidance}, both directions improve both metrics, while every $0<w<1$ is worse than either model. The two models were trained on different data, so there is no condition to amplify, and a gain in both directions is hard to reconcile with error correction alone. Two more pairs, SD~2.1 with SD~1.5 and with CommonCanvas, behave the same way (Appendix~\ref{app:both-directions}). Not every pair does, and what separates those that do is left to future work.

\paragraph{Guiding with a checkpoint needs a larger scale than with an independent model.}
Our interpretation in \cref{sec:second-moment} assumes that the two models are trained on independent data, but in autoguidance the weak model is often an earlier checkpoint of the strong one. The strong model has then seen all $N_0$ samples of the weak one and $N_1-N_0$ new ones. In our simplified model of training, the $m_0(x_t)$ samples that the weak model averages are thus also averaged by the strong one, and their errors cancel in the difference $\Delta=D_1-D_0$ that guidance amplifies. The difference thus has a smaller variance than with an independent model (Appendix~\ref{app:guidance-nested}):
\begin{equation}
    \thickmuskip=0mu\medmuskip=0mu\thinmuskip=0mu\Cov(\Delta_{\mathrm{cp}})\!=\!\bigl(\tfrac{1}{m_0}\!-\!\tfrac{1}{m_1}\bigr)\Sigma\!\prec\!\bigl(\tfrac{1}{m_0}\!+\!\tfrac{1}{m_1}\bigr)\Sigma\!=\!\Cov(\Delta_{\mathrm{id}})
    \label{eq:cov-diff}
\end{equation}
for a checkpoint and an independent model, where $\Sigma=\Cov(x_0|x_t)$. Guidance must therefore stretch the difference further: $w^\star\approx(1/m_0\mp1/m_1)^{-1/2}$ for large $m_0,m_1$. We test this with two EDM2-XS runs on ImageNet-64 that differ only in their seed, taking the weak model from the strong model's own run or from the other run at the same step. The checkpoint indeed needs the larger FID- and FD$_\mathrm{DINOv2}$-optimal scale (\cref{fig:teaser-nested}), but only once it has at least half of the strong model's training, and by less than \cref{eq:cov-diff} implies. The gap shows how training forgets: our model averages all samples alike, but SGD moves the weights toward each new batch, so an early checkpoint shares little with the final model. An exponential moving average over batches with one fitted memory length reproduces the measured ratios (\cref{fig:teaser-nested}, Appendix~\ref{app:ema}).

\FloatBarrier

\begin{figure}[t]
  \centering
  \includegraphics[width=\linewidth]{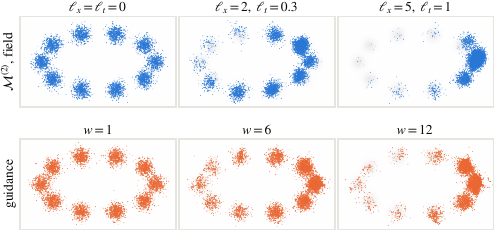}
  \caption{Correlated errors reproduce the diversity drop of guidance on the ten-Gaussian mixture of \cref{fig:story} (true density in gray; $5{,}000$ samples per panel). Top: the exact mean + covariance sampler (\cref{eq:M(2)}, labeled $\mathcal{M}^{(2)}$) with independent noise $\varepsilon$ replaced by one fixed Gaussian random field $g(x_t,\sigma)$ shared across all trajectories (Appendix~\ref{app:diversity}), with correlation lengths $\ell_x$ across inputs and $\ell_t$ across $\log\sigma$ ($\ell_x=\ell_t=0$ is independent noise). Bottom: guidance (\cref{eq:guided-prediction}) with one trained model pair for increasing scale $w$. Setup in Appendix~\ref{app:trained-pairs}.}
  \label{fig:diversity-ladder}
\end{figure}

\paragraph{Guidance reduces diversity.}
Reduced sample diversity is a well-known side effect of guidance~\citep{ho2022cfg,kynkaanniemi2024interval}. \Cref{sec:Method} treats the models' errors at different $x_t$ as independent, so that each trajectory draws its own sample of the posterior. For nearby $x_t$, however, the errors are correlated: a fixed pair of trained models defines a smooth error function of $(x_t,\sigma)$~\citep{rahaman2019spectral} (we measure these correlations for EDM2 in Appendix~\ref{app:diversity}). Without guidance the predictions stay close to the posterior mean, and a small perturbation of $x_t$ can tip a trajectory into another mode. In our view, guidance amplifies this shared error and pushes every trajectory in a neighborhood toward the same point of the posterior rather than independent draws, so all of them commit to one mode. To illustrate this effect, we modify the exact mean + covariance sampler (\cref{eq:M(2)}) by replacing its independent noise $\varepsilon$ with a single fixed random field $g(x_t,\sigma)$ shared by all trajectories. At each $(x_t,\sigma)$, the field is still standard Gaussian, so the local posterior covariance is unchanged, but nearby inputs (within distance $\ell_x$) and noise levels (within $\ell_t$ in $\log\sigma$) receive correlated perturbations, with $\ell_x=\ell_t=0$ recovering independent noise. On the ten-Gaussian mixture of \cref{fig:story}, increasing $\ell_x$ and $\ell_t$ concentrates samples onto fewer modes while keeping them on the true clusters (\cref{fig:diversity-ladder}, top), matching the effect of increasing the guidance scale $w$ with a trained pair (\cref{fig:diversity-ladder}, bottom). Appendix~\ref{app:diversity} defines $g(x_t,\sigma)$ formally and shows the same trade-off on ImageNet-$64$.

\paragraph{Why CFG hurts at high noise.}
CFG is known to harm generation at high noise levels~\citep{kynkaanniemi2024interval}, which \citet{karras2024guiding} attribute to the conditional and unconditional denoisers targeting different distributions. Our interpretation gives this a geometric form: guidance restores the posterior covariance only if the two prediction covariances are aligned with it. In CFG at high noise, the principal directions of the conditional and unconditional predictions are substantially misaligned, whereas in the interval where CFG helps they agree (Appendix~\ref{app:vgap}). Under autoguidance both models target the same distribution, and their prediction covariances stay aligned throughout (Appendix~\ref{app:vgap}). Our mechanism thus breaks down where CFG is observed to hurt.

\section{Conclusion}
\label{sec:conclusion}
{\spaceskip=.25em plus .15em minus .09em\looseness=-1 At the scales used in practice, guidance moves a model's prediction away from the posterior mean, yet the samples improve (\cref{sec:shared-error}). Matching the variance of the guided prediction to that of the data gives variance-matched guidance, a closed-form, metric-free rule: its constant recovers the published scales of image, video, audio, motion and world models, and its schedule is competitive with published and hand-designed guidance without tuning (\cref{sec:experiments}). To explain why, we proposed that the difference between a strong and a weak model follows the posterior covariance, so that guidance turns two posterior-mean predictions into one with the posterior's mean and covariance (\cref{sec:Method}). This view also predicts that guidance works in both directions for some pairs and that a checkpoint needs a larger scale than a retrained model, which our experiments support (\cref{sec:exp-hypotheses}).\par}

\paragraph{Limitations and future work.}
While our rule recovers the guidance scales used in practice, our model of CFG covers only the prediction errors of its two branches, not the conditioning information that separates them (\cref{sec:shared-error}). Conditioning may add effects that this view misses, such as a push toward prompt alignment, which would explain why only the positive root works for CFG.

\section*{LLM Usage Statement}
We used LLM-based assistants during this work. Claude (Anthropic), through Claude Code, helped draft and edit the text, check grammar, typeset \LaTeX, and write and debug analysis and plotting code; OpenAI Codex was used to refine the layout of some figures. Claude was also run as an additional rater in the user study, reported separately from the human votes (Appendix~\ref{app:ai-rater}). All ideas, derivations, experiments, and claims are the authors' own, and every LLM-assisted passage, proof, and script was checked by the authors, who take full responsibility for its content.

\section*{Acknowledgments}
We thank Chenyang Yuan, Kotaro Ikeda, and Hamid Kamkari for thoughtful conversations, insightful ideas, and helpful feedback on the manuscript.

This work was supported by the National Science Foundation under Grant No.~2211260 and Cooperative Agreement EEC-2330040 (NSF Engineering Research Center for Human AugmentatioN via Dexterity, HAND), and AMD via the MIT AI Hardware Program. The MIT Geometric Data Processing Group acknowledges the generous support of National Science Foundation grants IIS2335492 and OAC2403239, from the CSAIL Future of Data and FinTechAI programs, from the MIT--IBM Watson AI Laboratory, from the Wistron Corporation, from the MIT Generative AI Impact Consortium, from the Toyota--CSAIL Joint Research Center, and from Schmidt Sciences.

\clearpage
{
    \small
    \bibliographystyle{ieeenat_fullname}
    \bibliography{bib/references}
}

\clearpage
\def\maketitlesupplementary{\twocolumn[{\centering{\Large\bfseries Appendix}\par\vspace{1.2em}}]}
\def\bibliography#1{}
\def\bibliographystyle#1{}

\maketitlesupplementary
\appendix

\noindent\textbf{Contents.}
\begin{itemize}[leftmargin=1.2em,itemsep=1pt,topsep=2pt]
\item \hyperref[app:proofs]{\textbf{\ref*{app:proofs}\quad Proofs}}: \hyperref[app:m2-rate]{\ref*{app:m2-rate} convergence rates};
  \hyperref[app:bootstrap]{\ref*{app:bootstrap} finite training and local estimation};
  \hyperref[app:guidance-indep]{\ref*{app:guidance-indep} independent models};
  \hyperref[app:guidance-nested]{\ref*{app:guidance-nested} checkpoints};
  \hyperref[app:ema]{\ref*{app:ema} training as a weighted moving average};
  \hyperref[app:wmse-proof]{\ref*{app:wmse-proof} MSE-optimal scale and shared-error model}.
\item \hyperref[app:setup]{\textbf{\ref*{app:setup}\quad Experimental setup}}: \hyperref[app:impl]{\ref*{app:impl} implementation};
  \hyperref[app:trained-pairs]{\ref*{app:trained-pairs} trained pairs};
  \hyperref[app:cfg-regimes]{\ref*{app:cfg-regimes} label information versus model error};
  \hyperref[app:shared-error-setup]{\ref*{app:shared-error-setup} EDM2 errors};
  \hyperref[sec:moments]{\ref*{sec:moments} samplers with posterior moments};
  \hyperref[app:tuned-w]{\ref*{app:tuned-w} metric-tuned scales};
  \hyperref[app:nested-setup]{\ref*{app:nested-setup} checkpoint guidance}.
\item \hyperref[app:more]{\textbf{\ref*{app:more}\quad Additional experiments}}: \hyperref[app:mean-worse]{\ref*{app:mean-worse} distance to the posterior mean};
  \hyperref[app:vgap]{\ref*{app:vgap} CFG versus autoguidance at high noise};
  \hyperref[app:diversity]{\ref*{app:diversity} diversity};
  \hyperref[app:both-directions]{\ref*{app:both-directions} guidance in both directions};
  \hyperref[app:frontier]{\ref*{app:frontier} metric trade-offs};
  \hyperref[app:edm2-fail]{\ref*{app:edm2-fail} effectiveness of VMG on ImageNet}.
\item \hyperref[app:ablations]{\textbf{\ref*{app:ablations}\quad Ablations}}: \hyperref[app:fullcov]{\ref*{app:fullcov} full covariance};
  \hyperref[app:objective]{\ref*{app:objective} objectives for the constant scale};
  \hyperref[app:ncalib]{\ref*{app:ncalib} calibration size};
  \hyperref[app:noise-control]{\ref*{app:noise-control} noise with the wrong covariance}.
\item \hyperref[app:results]{\textbf{\ref*{app:results}\quad Full results and user study}}: \hyperref[app:extra-models]{\ref*{app:extra-models} every metric};
  \hyperref[app:user-study]{\ref*{app:user-study} user study}.
\item \hyperref[app:samples]{\textbf{\ref*{app:samples}\quad Additional samples}}.
\end{itemize}

\section{Proofs}
\label{app:proofs}

\subsection{Convergence rates}
\label{app:m2-rate}\label{app:posterior-draws}
\label{sec:posterior-draws}

In \cref{sec:more-moments}, we saw empirically (\cref{fig:moments}) that modifying the DDIM step \eqref{eq:ddim-step} to move toward a Gaussian sample with the posterior's mean and covariance \eqref{eq:M(2)} reaches a given error in far fewer steps than standard DDIM, which uses only the posterior mean. This gap motivates our view of guidance in \cref{sec:second-moment} as turning two posterior-mean estimators into a prediction that carries the posterior covariance. Below we derive the convergence rates stated in \cref{sec:more-moments}: with $K$ steps, the Gaussian-kernel MMD to $p_{\mathrm{data}}$ is $O(K^{-1})$ for standard DDIM using the exact posterior mean, denoted $M(1)$, and $O(K^{-2})$ for the sampler using the exact posterior mean and covariance, denoted $M(2)$.

\paragraph{Samplers and one-step operators.}
Following \cref{sec:background}, we use the variance-exploding forward process \eqref{eq:ve}, $x_t = x_0 + \sigma_t \varepsilon$ with $\varepsilon \sim \mathcal{N}(0, I)$, where $\sigma_t$ grows with $t \in [0, 1]$ from $\sigma_0 = 0$ to a finite maximum noise level $\sigma_1 > 0$, and write $p_t$ for the marginal density of $x_t$. Denote the exact posterior mean and covariance at $x_t$ by
\[
\begin{aligned}
D^\star_t(x_t) &= \underset{x_0 \sim p(x_0 \mid x_t)}{\E}[x_0], \\[4pt]
\Sigma_t(x_t) &= \underset{x_0 \sim p(x_0 \mid x_t)}{\Cov}(x_0).
\end{aligned}
\]
Take a time grid $1 = t_0 > \cdots > t_K = 0$ with noise levels $\sigma_1 = \sigma_{t_0} > \cdots > \sigma_{t_K} = 0$ and increments $\delta_k = \sigma_{t_k} - \sigma_{t_{k+1}} > 0$, so that $\sum_{k=0}^{K-1} \delta_k = \sigma_1$. Given a clean-sample prediction $\tilde x_0$ at $x_{t_k}$, the DDIM update \eqref{eq:ddim-step} from $t_k$ to $t_{k+1}$ is
\[
\begin{aligned}
x_{t_{k+1}}
&= \frac{\sigma_{t_{k+1}}}{\sigma_{t_k}} x_{t_k}
   + \Bigl(1 - \frac{\sigma_{t_{k+1}}}{\sigma_{t_k}}\Bigr)\tilde x_0 \\
&= x_{t_k} - \frac{\delta_k}{\sigma_{t_k}}(x_{t_k} - \tilde x_0).
\end{aligned}
\]
We compare three choices of $\tilde x_0$:
\begin{itemize}[leftmargin=1.2em,itemsep=1pt,topsep=2pt]
\item $M(1)$ (standard DDIM): $\tilde x_0 = D^\star_{t_k}(x_{t_k})$;
\item $M(2)$ (mean + covariance, \cref{eq:M(2)}): $\tilde x_0 = D^\star_{t_k}(x_{t_k}) + \Sigma_{t_k}(x_{t_k})^{1/2}\varepsilon$ with fresh $\varepsilon \sim \mathcal{N}(0, I)$;
\item full posterior (\cref{eq:posterior-sampler}): $\tilde x_0 = \hat x_0 \sim p(x_0 \mid x_{t_k})$.
\end{itemize}
To compare a sampler's output distribution $p_0^{(r)}$ against $p_{\mathrm{data}}$ in MMD, we compare expectations $\E[\varphi(x)]$ over a family of smooth scalar probe functions $\varphi \colon \mathbb{R}^d \to \mathbb{R}$, commonly called \emph{test functions} (specifically, functions in $C_b^3$, meaning bounded with bounded derivatives up to order three; as we note at the end, Gaussian-kernel MMD measures the worst-case difference over a subset of such smooth functions). Rather than tracking probability densities directly, it is simpler to track how one step changes the expectation of any test function $f$. Writing the residual $M_r(x_{t_k}, t_k) = x_{t_k} - \tilde x_0$ for $r \in \{1, 2\}$, define the one-step expectation operators
\[
\begin{aligned}
Q_k^{(r)} f(x_{t_k})
&= \underset{\varepsilon \sim \mathcal{N}(0, I)}{\E}\!\left[
  f\!\left(x_{t_k} - \frac{\delta_k}{\sigma_{t_k}} M_r(x_{t_k}, t_k)\right)
\right], \\[4pt]
P_k f(x_{t_k})
&= \underset{\hat x_0 \sim p(x_0 \mid x_{t_k})}{\E}\!\left[
  f\!\left(x_{t_k} - \frac{\delta_k}{\sigma_{t_k}}(x_{t_k} - \hat x_0)\right)
\right],
\end{aligned}
\]
which give the expected value of $f(x_{t_{k+1}})$ conditioned on starting step $k$ at $x_{t_k}$. For $r = 1$, the update does not depend on $\varepsilon$ and the expectation in $Q_k^{(1)}$ drops out.

The full-posterior operator $P_k$ maps the exact marginal $p_{t_k}$ to $p_{t_{k+1}}$ at every step. Indeed, when $x_{t_k} \sim p_{t_k}$ and $\hat x_0 \sim p(x_0 \mid x_{t_k})$, the pair $(\hat x_0, x_{t_k})$ has the same joint distribution as $(x_0, x_{t_k})$, so $\hat\varepsilon = (x_{t_k} - \hat x_0)/\sigma_{t_k} \sim \mathcal{N}(0, I)$ is independent of $\hat x_0 \sim p_{\mathrm{data}}$. The update therefore equals
\[
x_{t_k} - \frac{\delta_k}{\sigma_{t_k}}(x_{t_k} - \hat x_0)
= \hat x_0 + \sigma_{t_{k+1}}\hat\varepsilon \sim p_{t_{k+1}}.
\]
Starting from $p_1$, applying $P_0 \cdots P_{K-1}$ thus recovers $p_{\mathrm{data}}$ exactly, which lets us bound the sampling error of $M(1)$ and $M(2)$ by comparing $Q_k^{(r)}$ against $P_k$.

\paragraph{Regularity assumptions.}
To expand a test function $f \in C_b^3$ around each step and keep its derivatives controlled across $K$ steps, we assume that $(D^\star_t(x_t) - x_t)/\sigma_t$ and $\Sigma_t(x_t)^{1/2}/\sigma_t$ have uniformly bounded spatial derivatives up to order three, that $\Sigma_t(x_t)^{1/2}/\sigma_t$ is uniformly bounded, and that the posterior's third central moment scales as
\begin{equation}
\underset{x_0 \sim p(x_0 \mid x_t)}{\E}\bigl[\|x_0 - D^\star_t(x_t)\|^3\bigr] \le C \sigma_t^3.
\label{eq:m2-third-moment}
\end{equation}
The derivative bounds keep the test functions smooth under repeated steps, while \eqref{eq:m2-third-moment} controls the third-order Taylor remainder of the non-Gaussian posterior draw in $P_k$.

\paragraph{Error of a single step.}
Fix $x_{t_k}$ and let $a = x_{t_k} - \frac{\delta_k}{\sigma_{t_k}}(x_{t_k} - D^\star_{t_k}(x_{t_k}))$ be the deterministic DDIM point of $M(1)$. All three updates take the form $a + \delta_k u$, where $u = 0$ for $M(1)$ and
\[
u = \begin{cases}
\Sigma_{t_k}(x_{t_k})^{1/2}\varepsilon / \sigma_{t_k}, & M(2), \\
(\hat x_0 - D^\star_{t_k}(x_{t_k})) / \sigma_{t_k}, & \text{full posterior}.
\end{cases}
\]
At fixed $x_{t_k}$, write $\E_u$ for the expectation over $\varepsilon \sim \mathcal{N}(0, I)$ for $M(2)$ and over $\hat x_0 \sim p(x_0 \mid x_{t_k})$ for $P_k$. Both have zero mean and the same covariance,
\[
\E_u[u] = 0, \qquad
\E_u[u u^\top] = \frac{\Sigma_{t_k}(x_{t_k})}{\sigma_{t_k}^2}.
\]
Their third moments $\E_u[\|u\|^3]$ are also uniformly bounded by a constant $C'$: for $M(2)$ because $\varepsilon$ is standard Gaussian and $\Sigma_t^{1/2}/\sigma_t$ is bounded, and for $P_k$ by \eqref{eq:m2-third-moment}.

Now take any test function $f \in C_b^3$ and expand $f(a + \delta_k u)$ around $a$ to third order. Taking $\E_u$, the first-order term vanishes because $\E_u[u] = 0$, leaving
\[
\E_u[f(a + \delta_k u)]
= f(a) + \frac{\delta_k^2}{2}\,\nabla^2 f(a) : \E_u[u u^\top]
  + O(\delta_k^3 \|f\|_{C^3}),
\]
where the remainder is bounded by $\frac{1}{6}\delta_k^3 \|D^3 f\|_\infty \E_u[\|u\|^3]$. For $M(1)$, $u = 0$ so $Q_k^{(1)} f(x_{t_k}) = f(a)$; it misses the covariance term of $P_k$, which is $O(\delta_k^2 \|f\|_{C^3})$ since $\Sigma_t/\sigma_t^2$ is uniformly bounded. For $M(2)$, the covariance term matches that of $P_k$, so the first two orders cancel and only the $O(\delta_k^3 \|f\|_{C^3})$ remainder survives. Because both bounds hold uniformly in $x_{t_k}$, each step satisfies
\begin{equation}
\|(Q_k^{(r)} - P_k) f\|_\infty \le C_1 \delta_k^{r+1} \|f\|_{C^3},
\qquad r \in \{1, 2\}.
\label{eq:m2-one-step}
\end{equation}

\paragraph{Accumulating the error over $K$ steps.}
Fix $r \in \{1, 2\}$ and write $Q_k = Q_k^{(r)}$. Let $p_0^{(r)}$ be the law of the output $x_0^{(r)}$ of $M(r)$ initialized from $x_1^{(r)} \sim p_1$, and let $\varphi \in C_b^3$ be any test function. Because both $M(r)$ and the full-posterior process start from $p_1$, and $P_0 \cdots P_{K-1}$ maps $p_1$ to $p_{\mathrm{data}}$,
\[
\begin{aligned}
\underset{x \sim p_0^{(r)}}{\E}[\varphi(x)]
&= \underset{x \sim p_1}{\E}\bigl[(Q_0 \cdots Q_{K-1})\varphi(x)\bigr], \\[4pt]
\underset{x \sim p_{\mathrm{data}}}{\E}[\varphi(x)]
&= \underset{x \sim p_1}{\E}\bigl[(P_0 \cdots P_{K-1})\varphi(x)\bigr].
\end{aligned}
\]
We split the difference into one-step errors using the telescoping identity
\[
\begin{aligned}
&Q_0 \cdots Q_{K-1} - P_0 \cdots P_{K-1} \\
&\quad= \sum_{k=0}^{K-1} P_0 \cdots P_{k-1} (Q_k - P_k) Q_{k+1} \cdots Q_{K-1}.
\end{aligned}
\]
Writing $f_{k+1} = Q_{k+1} \cdots Q_{K-1}\varphi$ (with $f_K = \varphi$) and using that $P_0 \cdots P_{k-1}$ maps $p_1$ to $p_{t_k}$, the expectation error becomes
\[
\begin{aligned}
&\underset{x \sim p_0^{(r)}}{\E}[\varphi(x)]
 - \underset{x \sim p_{\mathrm{data}}}{\E}[\varphi(x)] \\[4pt]
&\quad= \sum_{k=0}^{K-1} \underset{x \sim p_1}{\E}\bigl[P_0 \cdots P_{k-1}(Q_k - P_k) f_{k+1}(x)\bigr] \\[4pt]
&\quad= \sum_{k=0}^{K-1} \underset{x \sim p_{t_k}}{\E}\bigl[(Q_k - P_k) f_{k+1}(x)\bigr].
\end{aligned}
\]

To apply \eqref{eq:m2-one-step} to every term, we need $\|f_{k+1}\|_{C^3}$ to stay bounded independently of $K$. In one step, $Q_k$ shifts $x_{t_k}$ by $\delta_k$ times $(D^\star_{t_k}(x_{t_k}) - x_{t_k})/\sigma_{t_k}$ plus (for $r = 2$) $(\Sigma_{t_k}(x_{t_k})^{1/2}/\sigma_{t_k})\varepsilon$. Differentiating $f$ composed with this map through order three by the chain rule (using our uniform spatial derivative bounds and that $\varepsilon \sim \mathcal{N}(0, I)$ has finite moments through order three) gives $\|Q_k f\|_{C^3} \le (1 + C_2 \delta_k)\|f\|_{C^3}$. Iterating from $f_K = \varphi$,
\[
\begin{aligned}
\|f_{k+1}\|_{C^3}
&\le \prod_{j=k+1}^{K-1}(1 + C_2 \delta_j)\,\|\varphi\|_{C^3} \\
&\le e^{C_2 \sum_{j=k+1}^{K-1} \delta_j}\,\|\varphi\|_{C^3}
 \le e^{C_2 \sigma_1}\|\varphi\|_{C^3},
\end{aligned}
\]
since $\sum_{j=0}^{K-1} \delta_j = \sigma_1$. Substituting this bound and the one-step error \eqref{eq:m2-one-step} into the telescoping sum gives
\[
\begin{aligned}
&\left|\underset{x \sim p_0^{(r)}}{\E}[\varphi(x)]
 - \underset{x \sim p_{\mathrm{data}}}{\E}[\varphi(x)]\right| \\[4pt]
&\quad\le C_1 \sum_{k=0}^{K-1} \delta_k^{r+1}\|f_{k+1}\|_{C^3} \\
&\quad\le C_1 e^{C_2 \sigma_1}\|\varphi\|_{C^3} \sum_{k=0}^{K-1} \delta_k^{r+1} \\
&\quad\le C_1 e^{C_2 \sigma_1}\sigma_1 \|\varphi\|_{C^3}(\max_k \delta_k)^r.
\end{aligned}
\]
With $K$ steps of size $\max_k \delta_k = O(K^{-1})$, every test function $\varphi \in C_b^3$ therefore satisfies
\begin{equation}
\left|\underset{x \sim p_0^{(r)}}{\E}[\varphi(x)]
 - \underset{x \sim p_{\mathrm{data}}}{\E}[\varphi(x)]\right|
\le C K^{-r}\|\varphi\|_{C^3},
\qquad r \in \{1, 2\}.
\label{eq:m2-global}
\end{equation}

\paragraph{Convergence in MMD.}
By definition \citep{gretton2012kernel}, the Gaussian-kernel MMD is the largest gap on the left side of \eqref{eq:m2-global} over the unit ball of the kernel's reproducing kernel Hilbert space (RKHS), the space of smooth functions formed from linear combinations of Gaussian kernels $k(x, \cdot)$ with norm at most $1$. Because the Gaussian kernel is infinitely differentiable with bounded derivatives, every function in this unit ball has uniformly bounded derivatives through order three (so $\|\varphi\|_{C^3}$ is bounded by a constant). Taking the supremum over this unit ball in \eqref{eq:m2-global} gives
\begin{equation}
\begin{aligned}
\operatorname{MMD}\bigl(M(1), p_{\mathrm{data}}\bigr) &= O(K^{-1}), \\
\operatorname{MMD}\bigl(M(2), p_{\mathrm{data}}\bigr) &= O(K^{-2}).
\end{aligned}
\label{eq:m2-rates}
\end{equation}

\paragraph{Related results.} The sampler $M(2)$ adds noise with the posterior covariance to the DDIM step. \citet{schioppa2026covariance} recently proposed a sampler of this form, which estimates the covariance with Tweedie's formula and one Jacobian-vector product of the denoiser per step, and showed that it improves few-step sampling of pixel-space models. For DDPM with Gaussian reverse kernels, \citet{akhtar2026lanczos} prove the analogous gap in path-space KL divergence: $\Omega(1/K)$ with an isotropic reverse covariance and $O(1/K^2)$ with the posterior covariance, which they sample with Lanczos iterations of Jacobian-vector products. The derivation above gives the same gap for the DDIM step, measured by the MMD to the data. Both samplers estimate the covariance at every step at the cost of extra network evaluations; in our interpretation (\cref{sec:second-moment}), guidance obtains a perturbation with the posterior's covariance from the second forward pass that guided models already compute.

\subsection{Finite training and local posterior estimation}
\label{app:bootstrap}

This section formalizes the training model of \cref{sec:second-moment} for a network trained for a finite number of steps, and shows how such a network estimates the posterior at each input.

\paragraph{Pointwise denoising objective and optimal denoiser.}
A diffusion denoiser $D\colon \mathbb{R}^d \to \mathbb{R}^d$ is trained by minimizing the expected squared error over clean data samples $x_0 \sim p_{\mathrm{data}}$ and noise vectors $\varepsilon \sim \mathcal{N}(0, I)$:
\[
\begin{aligned}
\mathcal{L}(D) &= \underset{\substack{x_0 \sim p_{\mathrm{data}} \\ \varepsilon \sim \mathcal{N}(0, I)}}{\E} \bigl[ \| D(x_t) - x_0 \|^2 \bigr], \\[4pt]
x_t &= x_0 + \sigma_t \varepsilon.
\end{aligned}
\]
Writing the joint distribution of $(x_0, x_t)$ as $p_t(x_t)\, p(x_0 \mid x_t)$, the loss decomposes pointwise:
\[
\mathcal{L}(D) = \underset{x_t \sim p_t}{\E} \Bigl[ \underset{x_0 \sim p(x_0 \mid x_t)}{\E} \bigl[ \| D(x_t) - x_0 \|^2 \bigr] \Bigr].
\]
For any fixed input $x_t$, the expected Euclidean distance to $x_0$ is minimized by the conditional expectation (posterior mean):
\begin{equation}
D^\star(x_t) = \underset{x_0 \sim p(x_0 \mid x_t)}{\E}[x_0].
\label{eq:bayes-denoiser}
\end{equation}
The uncertainty remaining in the clean target given $x_t$ is described by the posterior covariance:
\begin{equation}
\begin{aligned}
\Sigma(x_t) &\equiv \underset{x_0 \sim p(x_0 \mid x_t)}{\Cov}(x_0) \\[4pt]
&= \underset{x_0 \sim p(x_0 \mid x_t)}{\E}\bigl[ (x_0 - D^\star)(x_0 - D^\star)^\top \bigr],
\end{aligned}
\label{eq:bayes-cov}
\end{equation}
with $D^\star = D^\star(x_t)$.

\paragraph{Piecewise-constant denoisers.}
In this paragraph we will consider a model of a denoiser approximated with a piecewise constant function.

In practice, a model sees a finite number $N$ of training pairs $\{ (x_0^{(i)}, x_t^{(i)}) \}_{i=1}^N$ drawn from the forward diffusion process, with $x_0^{(i)} \sim p_{\mathrm{data}}$, $\varepsilon^{(i)} \sim \mathcal{N}(0, I)$ and $x_t^{(i)} = x_0^{(i)} + \sigma_t \varepsilon^{(i)}$.
Since $\varepsilon$ is continuous, no $x_t$ is sampled twice during training, so each $x_t$ has only a single target $x_0^{(i)}$ that the network was shown. A model that fits each pair exactly would return $x_0^{(i)}$ at $x_t^{(i)}$, effectively overfitting. In reality, however, neural networks are known to have a smoothness bias~\citep{rahaman2019spectral}. To model this effect in a simple and intuitive way, we consider the case where the neural network is piecewise constant over a fixed set of cells.

In more detail, suppose the input space is partitioned into a collection of cells $\{ R_k \}_k$. A piecewise-constant denoiser assigns a constant prediction vector $c_k \in \mathbb{R}^d$ to every point in cell $R_k$:
\[
D(x_t) = \sum_k c_k \, \mathbf{1}\{ x_t \in R_k \}.
\]
On the training dataset, the empirical squared error splits into independent sums, one for each cell:
\[
\mathcal{L}_{\mathrm{emp}}(D) = \frac{1}{N} \sum_k \sum_{i:\, x_t^{(i)} \in R_k} \| c_k - x_0^{(i)} \|^2.
\]
Write $I_k=\{i:\, x_t^{(i)} \in R_k\}$ for the pairs whose inputs fall in cell $R_k$ and $m_k=|I_k|$ for their number. Only the terms with $i \in I_k$ depend on $c_k$, and their gradient is
\[
\begin{aligned}
\nabla_{c_k} \mathcal{L}_{\mathrm{emp}}(D) &= \frac{2}{N} \sum_{i \in I_k} \bigl( c_k - x_0^{(i)} \bigr) \\
&= \frac{2}{N} \Bigl( m_k\, c_k - \sum_{i \in I_k} x_0^{(i)} \Bigr).
\end{aligned}
\]
Setting it to zero and solving for $c_k$ gives
\begin{equation}
c_k = \frac{1}{m_k} \sum_{i \in I_k} x_0^{(i)}.
\label{eq:cell-mean}
\end{equation}
The Hessian with respect to $c_k$ is $\frac{2 m_k}{N} I$, which is positive definite when $m_k \ge 1$, so \eqref{eq:cell-mean} is the unique minimizer: each cell predicts the average of the targets $x_0^{(i)}$ whose inputs fall in it. An empty cell does not enter the loss and leaves $c_k$ undetermined.
Let $R(x_t)$ denote the cell containing $x_t$, $I(x_t)$ the corresponding index set, and $m(x_t)=|I(x_t)|$.
Assuming every cell received at least one sample during training, the network's prediction at $x_t$ is then
\begin{equation}
D(x_t) = \frac{1}{m(x_t)} \sum_{i \in I(x_t)} x_0^{(i)}.
\label{eq:piecewise-constant-denoiser}
\end{equation}

\paragraph{Sample count.}
Write $S = \{ (x_0^{(i)}, x_t^{(i)}) \}_{i=1}^N$ for the training pairs. A retrained model sees a new draw of $S$, so $\E_S$ and $\Cov_S$ below are the expectation and covariance over retraining, written $\E_{D_N \sim p_N}$ and $\Cov_{D_N \sim p_N}$ in \cref{eq:denoiser-covariance}.
The count $m(x_t)$ is random. Each of the $N$ pairs has $x_t^{(i)} \in R(x_t)$ independently, with probability $p_t(R(x_t))=\int_{R(x_t)} p_t(x')\,dx'$, so
\begin{equation}
\begin{gathered}
m(x_t) \sim \mathrm{Binomial}\bigl(N,\, p_t(R(x_t))\bigr), \\
\E_S\bigl[m(x_t)\bigr] = N\, p_t\bigl(R(x_t)\bigr).
\end{gathered}
\label{eq:m-count}
\end{equation}
It grows in proportion to the number of samples the network has seen, and depends on $x_t$ and the noise level through the size of the region and the density of $p_t$ on it.

\paragraph{Distribution of the targets in a region.}
The prediction \eqref{eq:piecewise-constant-denoiser} is random over retraining, because the training pairs are. To find its mean and covariance, we first find the distribution of the targets it averages. The pairs are independent, so given which of them fall in $R=R(x_t)$, the targets $x_0^{(i)}$, $i \in I(x_t)$, are independent of each other, and each follows the distribution of $x_0$ given that its noisy version lies in $R$. Let
\[
p_R(x_t') = \frac{p_t(x_t')\, \mathbf{1}\{ x_t' \in R \}}{p_t(R)}
\]
be the density of a noisy input conditioned to lie in $R$. The targets then follow
\begin{equation}
q_R(x_0) = \int p(x_0 \mid x_t')\, p_R(x_t')\, dx_t',
\label{eq:region-posterior}
\end{equation}
the posterior averaged over the inputs in the region. We renumber these targets $x_0^{(1)},\dots,x_0^{(m(x_t))}$ in the order the network sees them. By the law of total expectation and the law of total covariance over $x_t' \sim p_R$, with $D^\star$ and $\Sigma$ from \eqref{eq:bayes-denoiser} and \eqref{eq:bayes-cov}, their mean and covariance are
\begin{equation}
\begin{aligned}
\mu_R &\equiv \underset{x_0 \sim q_R}{\E}[x_0] = \underset{x_t' \sim p_R}{\E}\bigl[ D^\star(x_t') \bigr], \\[4pt]
\Sigma_R &\equiv \underset{x_0 \sim q_R}{\Cov}(x_0) \\[4pt]
&= \underset{x_t' \sim p_R}{\E}\bigl[ \Sigma(x_t') \bigr] + \underset{x_t' \sim p_R}{\Cov}\bigl( D^\star(x_t') \bigr).
\end{aligned}
\label{eq:region-moments}
\end{equation}

\paragraph{Mean and covariance of the prediction.}
Fix the count $m=m(x_t) \ge 1$, and write $\E_q$ and $\Cov_q$ for the expectation and covariance over the i.i.d.\ targets $x_0^{(1)},\dots,x_0^{(m)} \sim q_R$. By linearity, the mean of \eqref{eq:piecewise-constant-denoiser} is
\begin{equation}
\E_q\bigl[ D(x_t) \bigr] = \frac{1}{m} \sum_{j=1}^{m} \E_q\bigl[x_0^{(j)}\bigr] = \mu_R.
\label{eq:pc-mean}
\end{equation}
Its covariance is a double sum over pairs of targets,
\[
\Cov_q\bigl( D(x_t) \bigr) = \frac{1}{m^2} \sum_{j=1}^{m} \sum_{l=1}^{m} \Cov_q\bigl( x_0^{(j)}, x_0^{(l)} \bigr).
\]
The targets are independent, so every term with $j \ne l$ is zero, and each of the $m$ terms with $j=l$ equals $\Sigma_R$. Therefore
\begin{equation}
\Cov_q\bigl( D(x_t) \bigr) = \frac{1}{m^2} \cdot m\, \Sigma_R = \frac{\Sigma_R}{m}.
\label{eq:pc-cov}
\end{equation}
Both results are exact for the piecewise-constant model. Over retraining the count $m$ is itself random \eqref{eq:m-count}. Since the mean \eqref{eq:pc-mean} does not depend on $m$, the law of total covariance over $S$ gives $\Cov_S(D(x_t)) = \E_S[1/m]\, \Sigma_R$. When $N p_t(R)$ is large, $m$ concentrates around its mean, with relative spread at most $1/\sqrt{N p_t(R)}$, and $\E_S[1/m] \approx 1/\E_S[m]$. Below we write $m(x_t)$ for this typical count.

\paragraph{Relation to the posterior at $x_t$.}
So far the prediction estimates the moments of $q_R$, the posterior averaged over the region. To relate them to the posterior at $x_t$ itself, we suppose that the region is small enough that the posterior does not change within it, $p(x_0 \mid x_t') = p(x_0 \mid x_t)$ for all $x_t' \in R(x_t)$. Then $q_R = p(x_0 \mid x_t)$ by \eqref{eq:region-posterior}. In \eqref{eq:region-moments}, $D^\star(x_t') = D^\star(x_t)$ and $\Sigma(x_t') = \Sigma(x_t)$ on $R$, so $\mu_R = D^\star(x_t)$, the first term of $\Sigma_R$ becomes $\Sigma(x_t)$, and the second term vanishes. Substituting into \eqref{eq:pc-mean} and \eqref{eq:pc-cov}, over retraining,
\begin{equation}
\begin{aligned}
\underset{D_N\sim p_N}{\E}\bigl[D_N(x_t)\bigr] &= \underset{x_0 \sim p(x_0 \mid x_t)}{\E}[x_0], \\[4pt]
\underset{D_N\sim p_N}{\Cov}\bigl(D_N(x_t)\bigr) &= \frac{1}{m(x_t)} \underset{x_0 \sim p(x_0 \mid x_t)}{\Cov}(x_0),
\end{aligned}
\label{eq:denoiser-moments}
\end{equation}
which are the mean and covariance in \eqref{eq:denoiser-covariance}. The covariance of the prediction is thus proportional to the posterior covariance, with factor $1/m(x_t)$.
For a real network the posterior changes within the region, and the region size trades bias for variance. A larger region contains more pairs, so $m(x_t)$ grows and the covariance shrinks, but $\mu_R$ moves away from $D^\star(x_t)$ and the second term of $\Sigma_R$ adds spread that does not follow $\Sigma(x_t)$.
As \eqref{eq:denoiser-moments} predicts, the scatter of retrained models in \cref{fig:story}b,e keeps the shape of $\Sigma(x_t)$ and shrinks as the models see more samples.

\subsection{Guidance with independent models}
\label{app:guidance-indep}

Following \cref{sec:second-moment}, we derive the distribution of the guided prediction for two independently trained denoisers and show that two scales match the posterior covariance, which gives guidance in both directions (\cref{sec:exp-hypotheses}).

\paragraph{Linear combination and moments.}
Let $D_0$ and $D_1$ denote a weaker and a stronger denoiser, with effective numbers $m_0$ and $m_1$ satisfying $m_0 < m_1$.
Let $D_0$ and $D_1$ be trained independently (with separate seeds and separate samples).
By independence, over $p=(p_{N_0},p_{N_1})$ as in \cref{sec:vm-guidance},
\begin{align*}
\E_p[D_0(x_t)] = \E_p[D_1(x_t)] &= \underset{x_0 \sim p(x_0 \mid x_t)}{\E}[x_0], \\[4pt]
\Cov_p(D_0(x_t)) &= \frac{\Sigma(x_t)}{m_0}, \\
\Cov_p(D_1(x_t)) &= \frac{\Sigma(x_t)}{m_1}, \\
\Cov_p\bigl( D_0(x_t), D_1(x_t) \bigr) &= 0.
\end{align*}
The guided prediction with scale $w$ is
\[
\begin{aligned}
D_w(x_t) &= (1 - w) D_0(x_t) + w D_1(x_t) \\
&= D_0(x_t) + w \Delta(x_t),
\end{aligned}
\]
where $\Delta(x_t) = D_1(x_t) - D_0(x_t)$ is the difference of the two predictions.
Taking the expectation over retraining runs:
\[
\begin{aligned}
\E_p[D_w(x_t)] &= (1 - w) \E_p[D_0(x_t)] + w \E_p[D_1(x_t)] \\
&= \underset{x_0 \sim p(x_0 \mid x_t)}{\E}[x_0],
\end{aligned}
\]
so for every $w \in \mathbb{R}$ the guided prediction is still an unbiased estimator of the posterior mean, and guidance changes only its covariance.

\paragraph{Covariance over retrainings.}
Using the independence condition $\Cov_p(D_0, D_1) = 0$, the covariance of the guided prediction decomposes as:
\begin{align*}
\Cov_p\bigl( D_w(x_t) \bigr) &= \Cov_p\bigl( (1 - w) D_0(x_t) + w D_1(x_t) \bigr) \\
&= (1 - w)^2 \Cov_p(D_0) + w^2 \Cov_p(D_1) \\
&= \left[ \frac{(1 - w)^2}{m_0} + \frac{w^2}{m_1} \right] \Sigma(x_t) \\
&= c(w,m_1,m_0) \, \Sigma(x_t),
\end{align*}
with the factor $c(w,m_1,m_0)$ of \cref{eq:cov-delta}, which we write $c(w)$ below. It is strictly convex in $w$.
Differentiating with respect to $w$ and setting the derivative to zero:
\[
c'(w) = \frac{-2(1 - w)}{m_0} + \frac{2w}{m_1} = 0.
\]
Therefore,
\[
w^\star_{\mathrm{MSE}} = \frac{m_1}{m_0 + m_1}.
\]
Because $m_0 < m_1$, the variance minimizer lies in the interpolation interval $w^\star_{\mathrm{MSE}} \in (\frac{1}{2}, 1)$. This is the minimum-variance combination, which minimizes the estimation error and shrinks the spread below that of either model:
\[
c(w^\star_{\mathrm{MSE}}) = \frac{1}{m_0 + m_1} < \frac{1}{m_1} < \frac{1}{m_0}.
\]

\paragraph{Matching the posterior covariance.}
According to our interpretation in \cref{sec:second-moment}, guidance should instead restore the missing spread, so that the prediction has the full posterior covariance.
Setting $c(w) = 1$ requires:
\[
\frac{(1 - w)^2}{m_0} + \frac{w^2}{m_1} = 1.
\]
Multiplying both sides by $m_0 m_1$:
\[
m_1 (1 - 2w + w^2) + m_0 w^2 = m_0 m_1,
\]
which rearranges to the quadratic equation:
\begin{equation}
(m_0 + m_1) w^2 - 2 m_1 w + (m_1 - m_0 m_1) = 0.
\label{eq:quadratic-indep}
\end{equation}
The discriminant of \eqref{eq:quadratic-indep} is:
\begin{align*}
\delta &= (-2m_1)^2 - 4(m_0 + m_1)(m_1 - m_0 m_1) \\
&= 4 m_1^2 - 4\bigl( m_0 m_1 - m_0^2 m_1 + m_1^2 - m_0 m_1^2 \bigr) \\
&= 4 m_0 m_1 (m_0 + m_1 - 1).
\end{align*}
Because $m_0 \ge 1$ and $m_1 > 1$, we have $m_0 + m_1 - 1 > 0$, guaranteeing that $\delta > 0$ and that two real roots always exist.
The two roots are
\begin{align}
w^\star_\pm &= \frac{2 m_1 \pm \sqrt{4 m_0 m_1 (m_0 + m_1 - 1)}}{2 (m_0 + m_1)} \nonumber \\
&= \frac{m_1 \pm \sqrt{m_0 m_1 (m_0 + m_1 - 1)}}{m_0 + m_1}.
\label{eq:exact-roots-indep}
\end{align}

\paragraph{Asymptotic behavior and guidance in both directions.}
For $m_0, m_1 \gg 1$,
\begin{align*}
\frac{\sqrt{m_0 m_1 (m_0 + m_1 - 1)}}{m_0 + m_1} &= \sqrt{\frac{m_0 m_1}{m_0 + m_1}} \sqrt{1 - \frac{1}{m_0 + m_1}} \\
&= \frac{1}{\sqrt{1/m_0 + 1/m_1}} \\
&\quad \times \left( 1 - O\left( \frac{1}{m_0 + m_1} \right) \right),
\end{align*}
and the first term $m_1/(m_0 + m_1)$ lies in $(0, 1)$, so
\begin{equation}
w^\star_\pm = \pm \frac{1}{\sqrt{1/m_0 + 1/m_1}} + O(1).
\label{eq:asymp-roots-indep}
\end{equation}
The positive root, $w^\star_+ > 1$, extrapolates past $D_1$ away from $D_0$, as usual guidance does. The negative root, $w^\star_- < 0$, extrapolates the other way, past $D_0$ away from $D_1$. Since $c(w^\star_-) = c(w^\star_+) = 1$, both give the posterior covariance $\Sigma(x_t)$.
This is consistent with \cref{fig:random_guidance}, where guidance between independently trained models improves samples at both positive and negative scales.

\subsection{Guidance with checkpoints}
\label{app:guidance-nested}

In autoguidance, the weak denoiser $D_0$ is often an earlier checkpoint of the strong model $D_1$. We derive \cref{eq:cov-diff} of \cref{sec:exp-hypotheses} and the matched scales for this case, with $p$ now the distribution of the training run that produces both models.

In our idealized model of training, the strong model $D_1$ averages all $m_0$ targets $x_0^{(1)},\dots,x_0^{(m_0)}$ that the checkpoint $D_0$ averages, plus $m_1 - m_0$ new independent targets $x_0^{(m_0+1)}, \dots, x_0^{(m_1)}$:
\begin{align*}
D_0(x_t) &= \frac{1}{m_0} \sum_{j=1}^{m_0} x_0^{(j)}, \\
D_1(x_t) &= \frac{m_0}{m_1} D_0(x_t) + \frac{1}{m_1} \sum_{j=m_0+1}^{m_1} x_0^{(j)}.
\end{align*}
Because the targets $x_0^{(m_0+1)}, \dots, x_0^{(m_1)}$ are drawn independently of the first $m_0$ targets, the sum $\sum_{j=m_0+1}^{m_1} x_0^{(j)}$ is uncorrelated with $D_0(x_t)$.
The cross-covariance between the two checkpoints across retrainings is therefore:
\begin{align*}
\Cov_p\bigl( D_1(x_t), D_0(x_t) \bigr) &= \frac{m_0}{m_1} \Cov_p\bigl( D_0(x_t) \bigr) \\
&= \frac{m_0}{m_1} \left( \frac{\Sigma(x_t)}{m_0} \right) = \frac{\Sigma(x_t)}{m_1}.
\end{align*}

\paragraph{Variance of the difference for a checkpoint.}
The difference of the two predictions is $\Delta = D_1 - D_0$.
Both models have mean $\E_{x_0 \sim p(x_0 \mid x_t)}[x_0]$, so its expectation across retrainings is $\E_p[\Delta] = \E_p[D_1] - \E_p[D_0] = 0$.
Expanding its covariance:
\begin{align}
\Cov_p(\Delta) &= \Cov_p(D_1 - D_0) \nonumber \\
&= \Cov_p(D_0) + \Cov_p(D_1) \nonumber \\
&\quad - 2 \Cov_p(D_1, D_0) \nonumber \\
&= \frac{\Sigma(x_t)}{m_0} + \frac{\Sigma(x_t)}{m_1} - 2\left( \frac{\Sigma(x_t)}{m_1} \right) \nonumber \\
&= \left( \frac{1}{m_0} - \frac{1}{m_1} \right) \Sigma(x_t).
\label{eq:cov-delta-nested-step}
\end{align}
Comparing \eqref{eq:cov-delta-nested-step} to the independent difference covariance $\Cov_p(\Delta_{\mathrm{id}}) = (1/m_0 + 1/m_1)\Sigma(x_t)$:
\begin{align*}
\Cov_p(\Delta_{\mathrm{cp}}) &= \left( \frac{1}{m_0} - \frac{1}{m_1} \right) \Sigma(x_t) \\
&\prec \left( \frac{1}{m_0} + \frac{1}{m_1} \right) \Sigma(x_t) = \Cov_p(\Delta_{\mathrm{id}}).
\end{align*}
Because the $m_0$ targets shared between the two checkpoints cancel in the difference $D_1 - D_0$, the variance of $\Delta_{\mathrm{cp}}$ is strictly smaller than that of an independent model pair $\Delta_{\mathrm{id}}$.

\paragraph{Orthogonality and guided covariance.}
For a checkpoint, the strong model $D_1$ and the difference $\Delta$ are uncorrelated:
\begin{align*}
\Cov_p(D_1, \Delta) &= \Cov_p(D_1, D_1 - D_0) \\
&= \Cov_p(D_1) - \Cov_p(D_1, D_0) \\
&= \frac{\Sigma(x_t)}{m_1} - \frac{\Sigma(x_t)}{m_1} = 0.
\end{align*}
Writing the guided prediction as $D_w = D_1 + (w - 1)\Delta$, the cross terms vanish:
\begin{align}
\Cov_p\bigl( D_w(x_t) \bigr) &= \Cov_p\bigl( D_1 + (w - 1)\Delta \bigr) \nonumber \\
&= \Cov_p(D_1) + (w - 1)^2 \Cov_p(\Delta) \nonumber \\
&= \frac{\Sigma(x_t)}{m_1} + (w - 1)^2 \left( \frac{1}{m_0} - \frac{1}{m_1} \right) \Sigma(x_t) \nonumber \\
&= \Bigl( \frac{1}{m_1} + (w - 1)^2 \, \frac{m_1 - m_0}{m_0 m_1} \Bigr) \Sigma(x_t).
\label{eq:nested-guided-cov-full}
\end{align}

\paragraph{Matched scale for a checkpoint.}
Setting the covariance factor in \eqref{eq:nested-guided-cov-full} equal to $1$:
\[
\frac{1}{m_1} + (w - 1)^2 \left( \frac{1}{m_0} - \frac{1}{m_1} \right) = 1.
\]
Subtracting $1/m_1$ from both sides:
\[
(w - 1)^2 \left( \frac{m_1 - m_0}{m_0 m_1} \right) = 1 - \frac{1}{m_1} = \frac{m_1 - 1}{m_1}.
\]
Multiplying by $m_0 m_1 / (m_1 - m_0)$:
\[
(w - 1)^2 = \frac{m_0 (m_1 - 1)}{m_1 - m_0},
\]
so
\begin{equation}
w^\star_\pm = 1 \pm \sqrt{\frac{m_0 (m_1 - 1)}{m_1 - m_0}}.
\label{eq:wstar-nested-exact-clean}
\end{equation}
For large sample counts $m_0, m_1 \gg 1$:
\begin{equation}
w^\star_\pm = 1 \pm \frac{1}{\sqrt{1/m_0 - 1/m_1}} \left( 1 + O\left( \frac{1}{m_1} \right) \right).
\label{eq:wstar-nested-asymp}
\end{equation}
Since $1/m_0 - 1/m_1 < 1/m_0 + 1/m_1$, the denominator in \eqref{eq:wstar-nested-asymp} is smaller than in \eqref{eq:asymp-roots-indep}, and
\[
w^\star_{\mathrm{cp}} > w^\star_{\mathrm{id}}.
\]
The shared targets cancel in the difference, so a checkpoint must be extrapolated further to restore the posterior covariance, as in \cref{eq:cov-diff}.

\subsection{Training as a weighted moving average}
\label{app:ema}

\Cref{sec:exp-hypotheses} finds that checkpoints need smaller scales than \cref{eq:cov-diff} implies. The derivation in \cref{app:guidance-nested} supposes that the network keeps all training targets with equal weight. It is more accurate to model training with stochastic gradient descent as a weighted moving average. Each step moves the weights toward the current batch, so the network reflects its recent batches more than its early ones. We model this and derive its effect on the scale for checkpoints.

\paragraph{Exponential moving average over minibatches.}
We model the network at step $k$ as an exponential moving average (EMA) over the targets $x_0^{(1)}, \dots, x_0^{(k)}$ of its training batches:
\[
D_k(x_t) = \sum_{i=1}^k c_{k,i} x_0^{(i)}, \qquad c_{k,i} = \frac{1 - \lambda}{1 - \lambda^k} \lambda^{k-i},
\]
where $\lambda \in (0, 1)$ is the per-step retention factor and $L \equiv \frac{1}{1 - \lambda}$ is the averaging length in steps.
For steps $k \gg L$, the normalization factor $1 - \lambda^k \approx 1$, yielding $c_{k,i} \approx (1 - \lambda) \lambda^{k-i}$.
The plain average corresponds to $\lambda \to 1$, where L'H\^opital's rule gives $c_{k,i} = 1/k$.

\paragraph{Variance of a checkpoint.}
The variance of checkpoint $D_k$ across retrained models is:
\[
\Cov_p(D_k) = V_k \, \Sigma(x_t), \qquad V_k \equiv \sum_{i=1}^k c_{k,i}^2.
\]
Evaluating $V_k$ for $k \gg L$:
\begin{align*}
V_k &\approx (1 - \lambda)^2 \sum_{j=0}^{k-1} \lambda^{2j} = (1 - \lambda)^2 \frac{1 - \lambda^{2k}}{1 - \lambda^2} \\
&\approx \frac{(1 - \lambda)^2}{1 - \lambda^2} = \frac{(1 - \lambda)^2}{(1 - \lambda)(1 + \lambda)} = \frac{1 - \lambda}{1 + \lambda}.
\end{align*}
For $k \gg L$, $V_k$ saturates at $\frac{1-\lambda}{1+\lambda}$ instead of decaying as $1/k$, in line with the seed-to-seed spread of our runs, which stops shrinking after about $2^{25}$ images.

\paragraph{Cross-covariance between checkpoints.}
Consider two checkpoints from the same training run at steps $k_0$ and $k_1$, with $k_1 > k_0$. With one target per step, $k_0$ and $k_1$ take the place of $m_0$ and $m_1$ in \cref{app:guidance-nested}.
For any early batch $i \le k_0$, the weight assigned by the final model $D_{k_1}$ is related to the weight assigned by the checkpoint $D_{k_0}$ by:
\[
c_{k_1, i} = \lambda^{k_1 - k_0}\,\frac{1 - \lambda^{k_0}}{1 - \lambda^{k_1}}\, c_{k_0, i}.
\]
The cross-covariance between the two checkpoints across retrainings is therefore:
\begin{align}
V_{01} &\equiv \sum_{i=1}^{k_0} c_{k_1, i} c_{k_0, i} \nonumber \\
&= \lambda^{k_1 - k_0}\,\frac{1 - \lambda^{k_0}}{1 - \lambda^{k_1}} \sum_{i=1}^{k_0} c_{k_0, i}^2 = \lambda^{k_1 - k_0}\,\frac{1 - \lambda^{k_0}}{1 - \lambda^{k_1}}\, V_{k_0}.
\label{eq:v01-decay}
\end{align}
For $k_0 \gg L$ the normalization ratio is close to $1$; in the plain-average limit $\lambda \to 1$ it tends to $k_0/k_1$, so $V_{01} = (k_0/k_1)\,V_{k_0} = 1/k_1$.
For two independently trained models, the batch sequences are completely independent, so $V_{01} = 0$.

\paragraph{Scale ratio for a weighted average.}
The variance of the difference $\Delta = D_{k_1} - D_{k_0}$ is:
\[
\Cov_p(\Delta) = (V_{k_1} + V_{k_0} - 2 V_{01}) \Sigma(x_t).
\]
Because the matched guidance scale is inversely proportional to the standard deviation of $\Delta$, the ratio of the autoguidance scale to the independent scale is:
\begin{equation}
\frac{w^\star_{\mathrm{cp}}}{w^\star_{\mathrm{id}}} = \sqrt{\frac{V_{k_1} + V_{k_0}}{V_{k_1} + V_{k_0} - 2 V_{01}}}.
\label{eq:ratio-formula-full}
\end{equation}

\paragraph{Comparison with measurements.}
For the plain average ($\lambda = 1$), $V_k = 1/k$ and $V_{01} = 1/k_1$, and with $r = k_0/k_1$, \eqref{eq:ratio-formula-full} becomes
\begin{equation}
\frac{w^\star_{\mathrm{cp}}}{w^\star_{\mathrm{id}}} = \sqrt{\frac{1/k_0 + 1/k_1}{1/k_0 - 1/k_1}} = \sqrt{\frac{1 + r}{1 - r}}.
\label{eq:no-forgetting-ratio}
\end{equation}
At $r = 1/8$, $1/4$, $1/2$ and $3/4$ this is $1.13$, $1.29$, $1.73$ and $2.65$, well above the ratios measured on EDM2-XS (\cref{fig:teaser-nested}), which stay at $1.0$ up to $r = 1/4$ and reach $1.6$ (FID) to $2.2$ (FD$_{\mathrm{DINOv2}}$) at $r = 3/4$.
A finite averaging length $L$ explains the gap. For an early checkpoint, $k_1 - k_0 \gg L$, the factor $\lambda^{k_1 - k_0} \approx e^{-(k_1 - k_0)/L}$ in \eqref{eq:v01-decay} vanishes, so $V_{01} \approx 0$. The strong model then gives almost no weight to the batches it shared with the checkpoint, the two behave as independent models, and the ratio is $1$. Only a late checkpoint, $k_1 - k_0 \lesssim L$, shares enough with the strong model to raise the ratio. \Cref{app:nested-setup} fits $L$ to the measured ratios.

\subsection{The MSE-optimal scale and the shared-error model}
\label{app:wmse-proof}\label{app:error-model}

We derive the closed form of the MSE-optimal scale \eqref{eq:wmse-main} of \cref{sec:shared-error} and show, with a model of the errors the two predictors share, that it lies below $1$ when they share little error.

\paragraph{The MSE-optimal scale.}
Write $\mathbb{E}_t$ for the expectation over $x_0\sim p_{\mathrm{data}}$, $\varepsilon\sim\mathcal{N}(0,I)$ and $x_t=x_0+\sigma_t\varepsilon$, let $\mathcal{L}_t(w)=\mathbb{E}_t\|D_w(x_t)-x_0\|^2$ be the error minimized in \eqref{eq:wmse-main}, and write $\Delta=D_1-D_0$, so that $D_w=D_0+w\Delta$. The error is quadratic in $w$; expanding the square,
\[
\mathcal{L}_t(w)
=\mathbb{E}_t\|D_0-x_0\|^2
-2w\,\mathbb{E}_t\langle x_0-D_0,\Delta\rangle
+w^2\,\mathbb{E}_t\|\Delta\|^2 .
\]
If $\mathbb{E}_t\|\Delta\|^2>0$, this is a convex parabola in $w$, and setting its derivative to zero gives
\begin{equation}
    w_{\mathrm{MSE}}^\star(t)
    =\frac{\mathbb{E}_t\langle x_0-D_0(x_t),\,D_1(x_t)-D_0(x_t)\rangle}
           {\mathbb{E}_t\|D_1(x_t)-D_0(x_t)\|^2}.
    \label{eq:wmse}
\end{equation}
Both expectations involve only the two predictions and the training pairs $(x_0,x_t)$, so $w_{\mathrm{MSE}}^\star$ can be estimated from samples without knowing $D^\star$.

\paragraph{The shared-error model.}
Following the shared-error intuition of \citet{karras2024guiding}, we write each prediction as the posterior mean plus an error shared by both models and an error of its own,
\begin{equation}
    D_i=D^\star+\alpha_i e+\beta_i e_i,\qquad i\in\{0,1\},
    \label{eq:error-model}
\end{equation}
where, at every $x_t$, the vectors $e$, $e_0$ and $e_1$ are orthonormal and the scalars $\alpha_i\in\mathbb{R}$ and $\beta_i\geq0$ set the error sizes. Since $\mathbb{E}_{x_0\sim p(x_0\mid x_t)}[x_0-D^\star(x_t)]=0$ at every $x_t$ and $\Delta$ is a function of $x_t$, the numerator of \eqref{eq:wmse} equals $\mathbb{E}_t\langle D^\star-D_0,\Delta\rangle$. Under \eqref{eq:error-model}, $D^\star-D_0=-\alpha_0e-\beta_0e_0$ and $\Delta=(\alpha_1-\alpha_0)e+\beta_1e_1-\beta_0e_0$, so by orthonormality
\begin{equation}
    w_{\mathrm{MSE}}^\star
    =\frac{\mathbb{E}_t\bigl[\beta_0^2+\alpha_0(\alpha_0-\alpha_1)\bigr]}
    {\mathbb{E}_t\bigl[\beta_0^2+\beta_1^2+(\alpha_0-\alpha_1)^2\bigr]}.
    \label{eq:wmse-model}
\end{equation}
If the models share almost no error, $\alpha_i\approx0$, this becomes
\[
w_{\mathrm{MSE}}^\star\approx\frac{\mathbb{E}_t[\beta_0^2]}{\mathbb{E}_t[\beta_0^2]+\mathbb{E}_t[\beta_1^2]}<1 .
\]
The best guided prediction then interpolates between the two models instead of extrapolating away from the weaker one. Extrapolation, $w_{\mathrm{MSE}}^\star>1$, needs the shared term $\mathbb{E}_t[\alpha_1(\alpha_0-\alpha_1)]$ to exceed the stronger model's own error $\mathbb{E}_t[\beta_1^2]$.

\section{Experimental setup}
\label{app:setup}
Below we document the model configurations, datasets, sampling protocols, and evaluation metrics needed to reproduce all experiments in the paper.

\subsection{Implementation details}
\label{app:impl}

\floatstyle{ruled}\newfloat{algorithm}{tbp}{loa}\floatname{algorithm}{Algorithm}
\ifdefined\crefname\crefname{algorithm}{Alg.}{Algs.}\Crefname{algorithm}{Algorithm}{Algorithms}\fi
\begin{algorithm}[t]
\caption{Variance-matched guidance (\cref{sec:vm-guidance}).}
\label{alg:vmg}
\small
\textbf{Input:} denoisers $D_0, D_1$; calibration set $\{(x_0^{(i)}, c^{(i)})\}_{i=1}^n$; sampler noise levels $\sigma_1,\dots,\sigma_K$.
\begin{enumerate}[leftmargin=*,itemsep=1.5pt,topsep=2pt]
\item \textbf{for} $k=1,\dots,K$ \textbf{do}
\begin{enumerate}[leftmargin=1.2em,itemsep=1pt,topsep=1pt,label=]
\item Noise $x^{(i)}=x_0^{(i)}+\sigma_k\varepsilon^{(i)}$, $\varepsilon^{(i)}\sim\mathcal{N}(0,I)$.
\item Evaluate $d_0^{(i)}=D_0(x^{(i)})$, $d_1^{(i)}=D_1(x^{(i)})$ (with $c^{(i)}$ where conditioned).
\item Compute $a_k, b_k, c_k$ over $i$ via \cref{eq:vm-coefficients}.
\item Set $w_k=\bigl(b_k+\sqrt{b_k^2-a_kc_k}\bigr)/a_k$ (\cref{eq:wstar}).
\end{enumerate}
\item \emph{Constant scale:} average $(a_k,b_k,c_k)$ over $k$ to $(\bar a,\bar b,\bar c)$; set $w=\bigl(\bar b+\sqrt{\bar b^2-\bar a\bar c}\bigr)/\bar a$ (\cref{eq:wconstant}).
\item \textbf{Return} schedule $\{w_k\}_{k=1}^K$ or constant $w$ for \cref{eq:guided-prediction}.
\end{enumerate}
\end{algorithm}

\Cref{alg:vmg} summarizes the method. For a model trained with another forward process, such as the flow-matching interpolation, the inputs $x^{(i)}$ are noised with that process at the noise level of step $k$.

\begin{table*}[t]
\centering
\caption{Models, samplers and data. Every model runs its released sampler at its default step count. Pub.\ is the published guidance scale in the convention of \cref{eq:guided-prediction}; SVD's is a per-frame linear ramp. $n$ is the number of calibration and evaluation examples, which never overlap.}
\label{tab:impl-models}\label{tab:impl-data}
\scriptsize
\setlength{\tabcolsep}{4pt}
\begin{adjustbox}{max width=\textwidth}
\begin{tabular}{@{}l c >{\raggedright\arraybackslash}l >{\raggedright\arraybackslash}p{2.2cm} >{\raggedright\arraybackslash}p{1.8cm} c l r r@{}}
\toprule
Model & Ref. & Checkpoint & Sampler, steps & Output & Pub.\ $w$ & Data & Calib.\ $n$ & Eval.\ $n$ \\
\midrule
SD 1.5 & \cite{rombach2022ldm} & \nolinkurl{stable-diffusion-v1-5/stable-diffusion-v1-5} & PNDM \cite{liu2022pndm}, 50 & $480^2$ image & 7.5 & COCO & 5,705 & 10,000 \\
SD 3.5 Medium & \cite{esser2024sd3} & \nolinkurl{stabilityai/stable-diffusion-3.5-medium} & Euler, 40 & $480^2$ image & 4.5 & COCO & 5,705 & 10,000 \\
SD 3.5 Large & \cite{esser2024sd3} & \nolinkurl{stabilityai/stable-diffusion-3.5-large} & Euler, 28 & $480^2$ image & 3.5 & COCO & 5,705 & 10,000 \\
FLUX.2 [klein] 4B & \cite{bfl2026flux2klein} & \nolinkurl{black-forest-labs/FLUX.2-klein-base-4B} & Euler, 50 & $1024^2$ image & 4.0 & MJHQ-30K & 1,024 & 10,000 \\
\addlinespace
SVD & \cite{blattmann2023svd} & \nolinkurl{stabilityai/stable-video-diffusion-img2vid} & Euler, Karras $\sigma$ \cite{karras2022edm}, 25 & 14 frames, $1024{\times}576$ & $1\to3$ & UCF-101 & 1,024 & 2,048 \\
Wan2.1 1.3B & \cite{wan2025wan} & \nolinkurl{Wan-AI/Wan2.1-T2V-1.3B-Diffusers} & UniPC \cite{zhao2023unipc}, 50 & 81 frames, $832{\times}480$ & 5 & OpenVid-1M & 256 & 1,024 \\
CogVideoX-2B & \cite{yang2025cogvideox} & \nolinkurl{zai-org/CogVideoX-2b} & DDIM \cite{song2021ddim}, 50 & 49 frames, $720{\times}480$ & 6 & OpenVid-1M & 256 & 1,024 \\
Cosmos-Predict2.5 2B & \cite{nvidia2025cosmos} & \nolinkurl{nvidia/Cosmos-Predict2.5-2B/robot/action-cond} & released, 35 per chunk & 13-frame chunks, $320{\times}256$ & 1 & Bridge & 1,024 chunks & 100 episodes \\
\addlinespace
AudioLDM 2 & \cite{liu2024audioldm2} & \nolinkurl{cvssp/audioldm2} & DDIM, 200 & 10.24\,s clip & 3.5 & AudioCaps & 2,223 & 4,411 \\
Stable Audio Open & \cite{evans2025sao} & \nolinkurl{stabilityai/stable-audio-open-1.0} & DPM-Solver++ 3M SDE \cite{lu2022dpmsolverpp}, 100 & 47\,s stereo clip & 7 & Song Describer & 356 & 586 \\
\addlinespace
MDM & \cite{tevet2023mdm} & \nolinkurl{kit_trans_enc_512} & DDPM \cite{ho2020ddpm}, 1,000 & KIT-ML pose sequence & 2.5 & KIT-ML & 864 & $20\times1{,}000$ \\
\addlinespace
\bottomrule
\end{tabular}
\end{adjustbox}
\end{table*}

\paragraph{Data.}
Every comparison evaluates on the same held-out prompts or examples, using those same ground-truth examples as the real reference set unless noted otherwise. We calibrate at every sampler step, except for MDM where we use $50$ evenly spaced steps out of $1{,}000$.
\begin{itemize}
    \item \emph{COCO} \citep{lin2014coco}: \texttt{val2014} photos with short side at least $480$\,px, shuffled in a fixed random order, center-cropped to $480^2$, and prompted with the longest of their five captions. Photos 0--5,704 are used for calibration and 5,705--15,704 for evaluation. Although $480^2$ is below the models' native resolution, upsampling COCO would blur $x_0$ and artificially reduce the target data variance.
    \item \emph{MJHQ-30K} \citep{li2024playground}: native $1024^2$ resolution; images 0--1,023 are used for calibration and 5,705--15,704 for evaluation.
    \item \emph{UCF-101} \citep{soomro2012ucf101}: for SVD, the first frame of each clip (center-cropped to 16:9 and resized to $1024{\times}576$) conditions generation, and the real clips receive the same preprocessing at $7$\,fps.
    \item \emph{OpenVid-1M} \citep{nan2025openvid}: calibration clips are taken from parts 30--70 and evaluation captions from parts 80--140, resampled to each model's native frame rate, frame count, and spatial size.
    \item \emph{Bridge} \citep{walke2023bridgedata}: the split of \citet{zhu2024irasim} used to evaluate Cosmos-Predict2.5. Validation calibration chunks are teacher-forced, and the variance moments are computed only over the generated latent frames; evaluation rolls out on the test episodes.
    \item \emph{AudioCaps} \citep{kim2019audiocaps}: $2{,}223$ validation captions (from $445$ clips) for calibration and all $4{,}411$ test captions (from $884$ clips) for evaluation.
    \item \emph{Song Describer} \citep{manco2023songdescriber}: $356$ captions (from $260$ tracks) for calibration and $586$ separate non-vocal captions for evaluation.
    \item \emph{KIT-ML} \citep{plappert2016kit}: $288$ validation motions (with three noise draws each, $864$ total) for calibration; evaluation uses $20$ replications of $1{,}000$ test motions against the test set as reference.
\end{itemize}

\paragraph{Metrics.}
\begin{itemize}
    \item \emph{Images}: FID and FD$_{\mathrm{DINOv2}}$ using the Inception-v3 and DINOv2 ViT-L/14 feature extractors from the EDM2 codebase \citep{karras2024edm2} (both real and generated images resized to $256^2$); precision, recall, density, and coverage \citep{naeem2020reliable} with $k=3$ on the same features; and CLIP score using CLIP ViT-L/14.
    \item \emph{Video}: FVD and $k$-NN metrics on StyleGAN-V's I3D features \citep{skorokhodov2022styleganv}, computed over $16$ evenly spaced frames (OpenVid-1M models) or non-overlapping $16$-frame windows (Cosmos-Predict2.5); CLIP-T averages CLIP ViT-L/14 similarity across $8$ frames against the prompt.
    \item \emph{Audio}: each model's standard evaluation suite. For AudioLDM~2: FAD on VGGish \citep{hershey2017vggish}, KL and Inception Score from PaSST, and CLAP; because that CLAP encoder also conditions the model, we add an independent LAION CLAP score and PaSST $k$-NN metrics. For Stable Audio Open: FD on OpenL3, KL on PaSST, and CLAP.
    \item \emph{Motion}: MDM's corrected KIT-ML evaluator.
    \item \emph{Uncertainty intervals}: for $k$-NN and video metrics, we report the spread of the paired difference across $100$ random half-samples (taking the same rows from both arms and the real reference), which matches the full-sample standard error. Cosmos-Predict2.5 uses a bootstrap over its $100$ episodes, MDM its $20$ replications, and Stable Audio Open the paired standard error over captions. We do not report error bars for FID, FD$_{\mathrm{DINOv2}}$, or CLIP score.
\end{itemize}

\paragraph{Cost.}
Fitting VMG on SD~3.5 Medium evaluates $D_0$ and $D_1$ on $5{,}705$ photos at $40$ noise levels, or about $4.6\times10^{5}$ network evaluations in total. By comparison, generating the $10{,}000$-image evaluation set at a single scale requires $8\times10^{5}$ evaluations, and the $42$-scale grid search behind \cref{tab:tuned-w} requires about $3.4\times10^{7}$.

\subsection{Strong and weak trained models}
\label{app:trained-pairs}
For the two-dimensional experiment in \cref{fig:story}, we use a mixture of ten Gaussians so that the exact posterior $p(x_0 \mid x_t)$ can be computed in closed form and compared directly against trained networks. The clean data distribution is an equal mixture of ten isotropic Gaussians, $p_{\mathrm{data}} = \frac{1}{10}\sum_{k=1}^{10} \mathcal{N}(\mu_k, s^2 I)$ with standard deviation $s = 0.15$, whose means $\mu_k$ are evenly spaced in angle around an ellipse with semi-axes $2$ and $1$. Adding Gaussian noise $x_t = x_0 + \sigma\varepsilon$ gives the observation likelihood $p(x_t \mid x_0) = \mathcal{N}(x_t; x_0, \sigma^2 I)$. By Bayes' rule, $p(x_0 \mid x_t) \propto p_{\mathrm{data}}(x_0)\, p(x_t \mid x_0)$. Because each prior component and the noise likelihood are both Gaussian in $x_0$, multiplying them produces another Gaussian for each cluster, so the posterior is again a ten-component Gaussian mixture:
\[
\begin{aligned}
p(x_0\mid x_t)&=\textstyle\sum_k\pi_k(x_t)\,\mathcal{N}\bigl(x_0;\,m_k(x_t),\,vI\bigr),\\
m_k(x_t)&=\mu_k+\tfrac{s^2}{s^2+\sigma^2}(x_t-\mu_k),\qquad v=\tfrac{s^2\sigma^2}{s^2+\sigma^2}.
\end{aligned}
\]
Here each posterior component mean $m_k(x_t)$ shifts the cluster center $\mu_k$ toward the noisy input $x_t$, each component has a smaller variance $v I$, and the mixing weight $\pi_k(x_t) \propto \mathcal{N}(x_t; \mu_k, (s^2+\sigma^2)I)$ reflects how likely $x_t$ is under the noised $k$-th cluster. From this closed-form mixture we obtain the exact posterior mean, covariance (via \cref{eq:mixture-moments}), and exact posterior samples.

\paragraph{Setup of \cref{fig:concept}.}
Each denoiser uses the preconditioning of \citet{karras2022edm} around a four-layer MLP ($128$ hidden units per layer, SiLU activations) that takes the scaled noisy point together with sine and cosine features of $\log\sigma$ at eight frequencies. We train each model with the denoising loss of \citet{karras2022edm}, drawing log-normal noise levels $\ln\sigma \sim \mathcal{N}(-1.0, 1.4^2)$, for $40{,}000$ steps with batch size $512$. We test two optimizers, Adam (learning rate $2\times10^{-3}$, $\beta=(0.9, 0.99)$) and SGD (momentum $0.9$, learning rate $0.1$), both with a $500$-step linear warm-up followed by cosine decay. To directly control the number of training samples each network sees (standing in for the finite sample budget of finite-step training in \cref{sec:second-moment} and \cref{app:bootstrap}), we vary the training-set size $N \in \{64, 128, \dots, 4{,}096\}$ and train $1{,}024$ independent models per $N$, each on its own fresh sample of $N$ training points and with its own random seed.

To evaluate the trained models, we fix $256$ noisy test inputs $x_t \sim p_\sigma$ at each noise level $\sigma \in \{0.1, 0.2, 0.4, 0.8, 1.6\}$ and use the same inputs across all $N$ and all models. At each input, we record every network's prediction alongside the exact posterior mean of its $N$-point empirical training set (the gray reference curve in \cref{fig:story}e). \Cref{fig:concept} shows one of these inputs at $\sigma=0.4$, and the shaded bands on the axis angle in \cref{fig:story}e are 5--95\% bootstrap intervals over the $1{,}024$ models. At this input, the spread of Adam-trained networks stays within $8\%$ of the exact empirical posterior mean up to $N=512$, and then grows slightly larger ($14\%$ more at $N=1{,}024$ and $38\%$ more at $N=4{,}096$). The cloud of network predictions also becomes rounder at large $N$, because the networks' own function-approximation error (which does not follow the shape of $\Sigma$) becomes comparable to the shrinking finite-sample spread.

\paragraph{Setup of \cref{fig:diversity-ladder}.}
The top row runs \cref{eq:M(2)} with the exact posterior moments for $K=512$ steps, with $\varepsilon$ replaced by one draw of the error field $g$ of \cref{eq:error-field}, shown mirrored left to right, which is an equally likely draw of $g$. For the bottom row, $D_1$ is the denoiser above with $256$ hidden units per layer, trained on a fresh batch from $p_{\mathrm{data}}$ at every step, and $D_0$ has $64$ hidden units per layer and is trained on its own $256$ points. Both are trained with Adam as above for $40{,}000$ steps, with features at six frequencies and noise levels drawn half from $\ln\sigma\sim\mathcal{N}(-0.6, 1.6^2)$ and half log-uniformly on $[0.002, 40]$, so that the whole sampling range is covered. We sample with $256$ deterministic Euler steps on the noise levels of \citet{karras2022edm} from $\sigma=20$ to $0.002$ and guide only at $\sigma\ge1$, using $D_1$ alone below. Both rows draw $5{,}000$ samples.

\subsection{CFG on ImageNet: label information versus model error}
\label{app:cfg-regimes}
This section describes the setup behind \cref{fig:cfg-regimes}, which separates conditioning signal from model error in CFG on ImageNet-64.

\paragraph{Models.}
Because EDM2's released ImageNet-64 models are strictly class-conditional, we train two EDM2-XS models with classifier-free label dropout ($p=0.1$) \citep{ho2022cfg}. Both follow the \texttt{edm2-img64-xs} preset of \citet{karras2024edm2} for $2^{27}$ training images at batch size $2{,}048$ (one-sixteenth of the published XS schedule), using two random seeds ($0$ and $1$, denoted runs $A$ and $B$). Changing the seed varies the weight initialization, data order, and noise draws. Throughout, we evaluate the training EMA weights at length $0.100$ and obtain the unconditional prediction by passing the zero label vector. Adding label dropout has little effect on seed-to-seed variability: the conditional-only runs trained for checkpoint guidance (\cref{app:nested-setup}) with the same preset and seeds differ by $\|D_A(x_t,c)-D_B(x_t,c)\|^2 = 13.3$ per image at $\sigma=2.9$, compared with $12.6$ for the label-dropout pair.

To approximate the true posterior mean, we use the released EDM2-XL model (EMA length $0.040$), which has nine times as many parameters and was trained by \citet{karras2024edm2} on five times as many images. This proxy is still not exact: at $\sigma=2.9$, EDM2-L achieves a lower denoising error than XL, implying that XL itself has a squared error of at least $13$ per image there, while the conditional XS models lie at squared distance $75$ per image from XL. The conditional-error curve in \cref{fig:cfg-regimes} therefore includes a small contribution from XL's own error. Importantly, the decomposition of $\Delta$ into seed-dependent and systematic parts below does not depend on XL at all.

\paragraph{Decomposition of $\Delta$.}
We take $4{,}096$ ImageNet training images with their class labels and add one noise draw to each image at each of the $32$ noise levels of the EDM2 sampler. At every noise level, we record the Gram matrix (all pairwise inner products) of the branch predictions (conditional and unconditional for both seeds $A$ and $B$, plus XL) centered around the mean image, from which every curve in \cref{fig:cfg-regimes} follows directly. Writing $\E_{x_0,c,x_t}$ for the average over these images, labels, and noisy inputs, we measure the squared CFG difference $\|\Delta\|^2$ (where $\Delta = D(x_t,c) - D(x_t)$) and the conditional prediction's error $\|D(x_t,c) - D_{\mathrm{XL}}(x_t,c)\|^2$. Using the two independent seeds, we split the average $\|\Delta\|^2$ into two parts that sum to $\frac{1}{2}\E[\|\Delta_A\|^2 + \|\Delta_B\|^2]$:
\begin{itemize}[leftmargin=1.2em,itemsep=1pt,topsep=2pt]
\item the \emph{seed-dependent} part, $\frac{1}{2}\E_{x_0,c,x_t}\|\Delta_A - \Delta_B\|^2$, which measures the random variance of $\Delta$ across independent training runs; and
\item the \emph{systematic} part, $\E_{x_0,c,x_t}\langle\Delta_A, \Delta_B\rangle$, which captures the part of $\Delta$ shared by both seeds (label information and any shared model bias).
\end{itemize}
The curves in \cref{fig:cfg-regimes} average both seeds, which agree to within line width. At the highest noise level ($\sigma=80$), the systematic part accounts for $97\%$ of $\|\Delta\|^2$ and matches within $2\%$ the reduction in denoising MSE from knowing the label, $\E_{x_0,c,x_t}\|D(x_t)-x_0\|^2 - \E_{x_0,c,x_t}\|D(x_t,c)-x_0\|^2$ (which are algebraically equal for exact conditional and unconditional posterior means).

\paragraph{Windowed guidance.}
Following interval guidance \citep{kynkaanniemi2024interval}, we generate samples using EDM2's $32$-step Heun sampler and turn guidance on ($w=2$) only at denoiser evaluations whose noise level $\sigma$ falls inside one of five windows: $(10,80]$, $(2.9,10]$, $(0.9,2.9]$, $(0.28,0.9]$, or $(0,0.28]$. These correspond to steps 1--11, 12--16, 17--19, 20--23, and 24--32 of the noise grid (because each Heun step evaluates the denoiser twice, its second evaluation can land in the next window). For every window, we generate $10{,}000$ images using the same initial noise and class labels and compute FID against the ImageNet-64 training set as in EDM2. Both training seeds agree on whether guidance helps or hurts in every window.

\subsection{Errors of EDM2 networks on ImageNet-64}
\label{app:shared-error-setup}
This section describes the setup for \cref{fig:mean-worse,fig:shared-error}, which test whether autoguidance reduces denoising error on ImageNet-64. Following the FID setup of \citet{karras2024guiding} (who use $w=1.7$), we guide EDM2-S (EMA length $0.045$, trained on $2^{30}$ images) with an EDM2-XS model trained on $2^{27}$ images, an eighth of S's budget (EMA length $0.110$). We use EDM2-XL (EMA length $0.040$, $1.1$B parameters versus $280$M for S) as a proxy for the posterior mean $D^\star$.

We evaluate on $10{,}000$ ImageNet validation images ($10$ per class, center-cropped to $64\times64$ as in EDM2), adding one shared noise draw per image at each of the $32$ noise levels of the EDM2 sampler. Expectations are averaged over these images. All errors are single-step denoising errors on forward-noised real images (not along guided sampling trajectories), measured as squared Euclidean distance in pixel space per coordinate ($d = 12{,}288$). The additional pairs mentioned in \cref{app:shared-error} (EDM2-L guided by S or M, and S guided by itself at EMA length $0.075$) use the exact same validation images and noise draws.

As in \cref{app:cfg-regimes}, EDM2-XL is an imperfect proxy for $D^\star$: between $\sigma \approx 2$ and $\sigma \approx 8$, EDM2-S actually has a $0.1$--$0.3\%$ lower denoising error than XL, and above $\sigma \approx 10$ the two agree within bootstrap noise. Any error of XL that is independent of both models adds equally to both estimated errors and biases the shared-error ratio $\kappa = \mathbb{E}_t[\alpha_0\alpha_1]/\mathbb{E}_t[\alpha_1^2]$ (the fraction of the strong model's error that the weak model repeats, with $\alpha_0,\alpha_1$ as in \cref{app:error-model}) upward toward $1$. To check this, for the two pairs with S as the strong model we repeat the measurement using the average of EDM2-M, L, and XL as $D^\star$, which lowers $\kappa$ from $0.55$--$0.67$ to $0.29$--$0.54$, reinforcing the conclusion that $\kappa < 1$.

For the sample-quality curves in \cref{fig:mean-worse}, we generate $10{,}000$ images per scale with EDM2's deterministic $32$-step Heun sampler (seeds $0$ through $9{,}999$ and random class labels) across $w \in \{-0.5, -0.25, \dots, 1.5, 1.7, 2.0, 2.25, 2.5, 3.0\}$, and compute FID and FD$_{\mathrm{DINOv2}}$ against EDM2's ImageNet-64 training reference statistics. Because we use $10{,}000$ samples rather than $50{,}000$, the absolute FID numbers are not directly comparable to the $50{,}000$-sample values reported by \citet{karras2024guiding}.

\subsection{Sampling with the first two posterior moments}
\label{sec:moments}\label{sec:moments-highd}
This section gives the setup for \cref{fig:moments}, which compares the mean ($M(1)$), mean + covariance ($M(2)$), and full-posterior ($M(\infty)$) samplers analyzed in \cref{app:m2-rate}. Both datasets in the figure have a closed-form mixture posterior $p(x_0 \mid x_t) = \sum_k \gamma_k(x_t)\, p_k(x_0 \mid x_t)$. By the laws of total expectation and total covariance, the posterior mean and covariance are
\begin{equation}
\begin{aligned}
    \underset{x_0\sim p(x_0\mid x_t)}{\E}[x_0]&=\textstyle\sum_k\gamma_k\,m_k=\bar m,\\[4pt]
    \underset{x_0\sim p(x_0\mid x_t)}{\Cov}(x_0)&=\textstyle\sum_k\gamma_k\,\Sigma_k\\
    &\quad+\textstyle\sum_k\gamma_k(m_k-\bar m)(m_k-\bar m)^{\!\top},
\end{aligned}
\label{eq:mixture-moments}
\end{equation}
where $m_k$ and $\Sigma_k$ are the mean and covariance of the $k$-th posterior component.

\paragraph{Two-dimensional mixture} (\cref{fig:moments}a).
We use the same ten-Gaussian ellipse mixture as in \cref{app:trained-pairs} ($s=0.15$, semi-axes $2$ and $1$). As derived there, conditioning on $x_t$ gives weights $\gamma_k \propto \mathcal{N}(x_t; \mu_k, (s^2+\sigma_t^2)I)$, means $m_k = \mu_k + \frac{s^2}{s^2+\sigma_t^2}(x_t - \mu_k)$, and covariances $\Sigma_k = \frac{s^2\sigma_t^2}{s^2+\sigma_t^2}I$.

\paragraph{ImageNet-64 images} ($d=12{,}288$, \cref{fig:moments}b).
We take $10{,}000$ random $64\times64$ ImageNet training images ($10$ per class), scaled so that the average per-coordinate variance is $1$, and treat the dataset as a discrete uniform distribution of $10{,}000$ point masses (weight $10^{-4}$ each). Here each component is a single image, so $m_k = x^{(k)}$ and $\Sigma_k = 0$, and the posterior weights $\gamma_k$ are simply a softmax of $-\|x_t - x^{(k)}\|^2 / (2\sigma_t^2)$ over the $10{,}000$ images. Thus the posterior mean $\bar m$ is the $\gamma$-weighted average of the dataset images, and the posterior covariance is their $\gamma$-weighted scatter around $\bar m$. Because forming a $12{,}288 \times 12{,}288$ covariance matrix at every step is impractical, $M(2)$ draws a Gaussian sample with the exact same mean and covariance directly via $\bar m + \sum_k \sqrt{\gamma_k}\,\xi_k(x^{(k)} - \bar m)$ using independent scalar draws $\xi_k \sim \mathcal{N}(0, 1)$.

\paragraph{Sampler and metrics.}
For both datasets we sample along the SD~3.5 Euler noise grid, drawing $10^6$ samples per step count and sharing the initial noise across $M(1)$, $M(2)$, and $M(\infty)$. We measure sample quality using Gaussian-kernel MMD to the data \citep{gretton2012kernel}, with kernel bandwidth $0.3$ on the 2D mixture and half the median nearest-neighbor distance between images on ImageNet-64 (so that only samples landing close to real images score well). The data expectation in MMD is computed directly from the data distribution. To establish the finite-sample noise floor, we evaluate eight independent runs of $10^6$ exact data samples; points within two standard deviations of zero are plotted on the floor.

\subsection{Metric-tuned scales}
\label{app:tuned-w}
To find the metric-tuned constants in \cref{tab:tuned-w,fig:tuned-w-metric}, we sweep constant guidance scales using each model's released sampler, score $10{,}000$ COCO images per scale, and pick the best scale for each metric individually. On SD~3.5 Medium we test $42$ scales from $w=1$ to $14$, a coarse grid refined to steps of $0.1$ around each metric's optimum ($40$ Euler steps), and on SD~1.5 we sweep $w=1$ to $16$ in steps of $0.5$ ($50$ PNDM steps).

\subsection{Guiding with an earlier checkpoint}
\label{app:nested-setup}
For \cref{fig:teaser-nested} (\cref{sec:exp-hypotheses}), we train two conditional-only EDM2-XS models on ImageNet-64 (runs $A$ and $B$) using the same preset, seeds, and EMA length ($0.100$) as in \cref{app:cfg-regimes} without label dropout, for $T = 2^{27}$ images, saving checkpoints every $2^{20}$ images. We take the strong model from either run at budget $T$ or $T/2$, and pair it with a weak guide at fraction $r \in \{1/8, 1/4, 1/2, 3/4\}$ of that budget, taken either from the same run (checkpoint) or from the other run (independent model). For each pair we generate $10{,}000$ images with EDM2's $32$-step Heun sampler (shared initial noise and class labels) across $w \in \{1, 1.25, 1.5, 2, 2.5, 3, 4, 5, 6.5, 8\}$, find the optimal scale $w^\star$ from a quadratic fit to FID or FD$_{\mathrm{DINOv2}}$ in $\log w$, and compute $w^\star_{\mathrm{cp}} / w^\star_{\mathrm{id}}$. Where guidance does not improve FID ($r = 1/8$, and $r = 1/4$ at budget $T$), both optima are $w = 1$ and the ratio is $1$.

The dashed curve in \cref{fig:teaser-nested} fits the single memory-length parameter $L$ of the EMA model (\cref{app:ema}) by least squares to the eight averaged ratios (pooling both seeds and strong budgets per metric and $r$), using one step per $2^{20}$ images at budget $T$, which gives $L = 0.69\,T$. Because direct estimates from unguided predictions are shorter, $L$ is best viewed as an effective summary of the scale ratios rather than a literal SGD memory length.

\renewcommand{\topfraction}{.9}\renewcommand{\dbltopfraction}{.9}\renewcommand{\bottomfraction}{.6}
\renewcommand{\textfraction}{.07}\renewcommand{\floatpagefraction}{.8}\renewcommand{\dblfloatpagefraction}{.8}
\setcounter{topnumber}{3}\setcounter{dbltopnumber}{2}\setcounter{bottomnumber}{2}\setcounter{totalnumber}{5}

\section{Additional experiments}
\label{app:more}

\subsection{Guidance moves the prediction away from the posterior mean}
\label{app:wmse}\label{app:shared-error}\label{app:mean-worse}
In \cref{sec:shared-error}, we found that the guidance scales used in practice improve sample quality despite increasing denoising error. Here, we examine this discrepancy more closely. We first test whether averaging over noise levels hides any benefit of guidance for predicting the posterior mean. We then compare mean-estimation accuracy and sample quality on a single model pair, and finally examine why the shared-error explanation of guidance does not account for these results.

\begin{figure}[!htb]
    \centering
    \includegraphics[width=\linewidth]{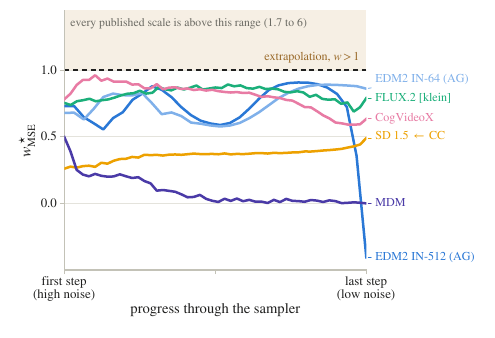}
    \caption{$w^\star_{\mathrm{MSE}}$ along the sampler for five of the pairs in \cref{fig:wmse-published} and one cross-model pair, from each sampler's first step (highest noise) to its last. AG marks autoguidance; SD~1.5 is guided away from CommonCanvas; the other pairs use CFG. The dashed line $w=1$ is no guidance, and every published scale ($1.7$ to $6$) lies above the plotted range.}
    \label{fig:wmse-sigma}
\end{figure}

\begin{figure}[!htb]
    \centering
    \includegraphics[width=\linewidth]{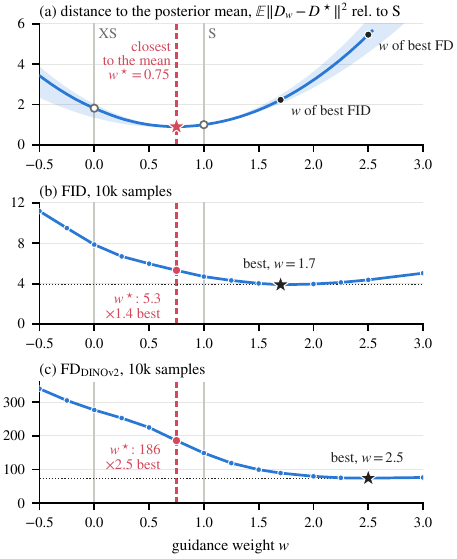}
    \caption{Distance to the posterior mean and sample quality (measured by FID and $\mathrm{FD}_{\mathrm{DINOv2}}$) against the guidance scale, for S guided by XS on ImageNet-64. (a)~Squared distance of the guided prediction $D_w$ to the posterior mean, relative to S, with XL standing in for $D^\star$: median over the 32 noise levels (line) and their range (band); the red line marks the closest scale $w^\star$. (b,c)~FID and $\mathrm{FD}_{\mathrm{DINOv2}}$ of 10{,}000 samples guided with a constant $w$; stars mark the best scale. Gray lines: XS ($w=0$) and S ($w=1$) alone.}
    \label{fig:mean-worse}
\end{figure}

\begin{figure*}[!t]
    \centering
    \includegraphics[width=\linewidth]{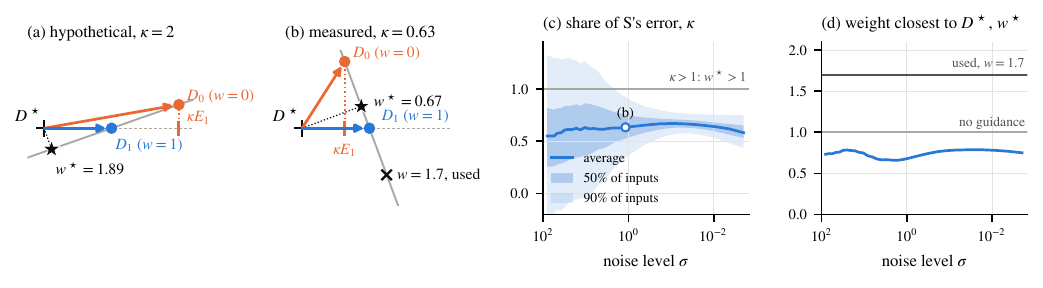}
    \caption{Error alignment of EDM2-S ($D_1$) and its guide EDM2-XS ($D_0$) on ImageNet-64, using EDM2-XL as a proxy for $D^\star$. (a,b)~2D geometry from averaged error inner products, normalized to $\|E_1\|=1$. The gray line traces $D_w$ through $D_0$ ($w=0$) and $D_1$ ($w=1$), the star marks the point closest to $D^\star$, and the dotted line projects XS's error onto S's ($\kappa E_1$). (a)~Idealized shared-error case ($\kappa=2$), where $D_w$ nears $D^\star$ around $w=2$. (b)~Measured at $\sigma=1.2$ ($\kappa=0.63$): the closest point falls between $D_0$ and $D_1$, while $w=1.7$ (cross) moves away from $D^\star$. (c)~$\kappa$ across noise levels (high to low) with 50\%/90\% per-input bands of $\alpha_0/\alpha_1$; the circle marks $\sigma=1.2$. (d)~Closest scale $w^\star$ (90\% bootstrap band) compared with $w=1.7$.}
    \label{fig:shared-error}
\end{figure*}

\paragraph{No noise level favors extrapolation for predicting the mean.} \Cref{sec:shared-error} reports the MSE-optimal scale \eqref{eq:wmse} pooled over the sampler; \cref{fig:wmse-sigma} resolves it step by step for five of the pairs of \cref{fig:wmse-published} and one cross-model pair. It stays below $1$ along the sampler, far below the published scales of $1.7$ to $6$.  This rules out the possibility that the pooled scale hides a few noise levels where guidance improves prediction.

\paragraph{The scale that best predicts the mean gives poor samples.} We test this on the pair for which autoguidance was introduced, EDM2-S guided by the smaller EDM2-XS on ImageNet-64. The denoising error splits as $\mathbb{E}_t\|D_w-x_0\|^2=\mathbb{E}_t\|D_w-D^\star\|^2+\mathbb{E}_t\|x_0-D^\star\|^2$, and the second term does not depend on $w$, so the denoising error and the distance to the posterior mean $D^\star$ change together. With the largest network, EDM2-XL, standing in for $D^\star$, the guided prediction is closest to it at $w^\star=0.75$ (\cref{fig:mean-worse}a). The sample metrics prefer very different scales: FID is best at $w=1.7$, the scale \citet{karras2024guiding} report, and $\mathrm{FD}_{\mathrm{DINOv2}}$ at $w=2.5$ (\cref{fig:mean-worse}b,c). At $w=1.7$ the prediction is $2.1$--$2.5$ times further from $D^\star$ than S alone, further even than the weaker XS, and at $w^\star$ FID is $1.4$ and $\mathrm{FD}_{\mathrm{DINOv2}}$ $2.5$ times its best value. The two objectives are therefore not just mildly misaligned but pull in opposite directions from $w=1$. This is the trade that motivates the paper, seen in full on one pair, as for SD~3.5 Medium in \cref{fig:wmse-published}c: guidance gives up accuracy in the posterior mean for better samples, so a rule that picks the scale must look at something other than the mean, which for us is the variance (\cref{sec:Method}).

\paragraph{Intuition for why the mean-optimal scale stays below $1$.} Extrapolating past $D_1$ ($w>1$) moves closer to $D^\star$ only if the weak model makes a larger version of the strong model's error. In \cref{app:error-model}, taking the shared axis along S's error $E_1 = D_1 - D^\star$ sets $\alpha_1 = \|E_1\|$, $\beta_1 = 0$, and $\alpha_0$ to the projection of XS's error onto $E_1$. By \eqref{eq:wmse-model}, $w_{\mathrm{MSE}}^\star > 1$ holds exactly when $\kappa = \mathbb{E}_t[\alpha_0\alpha_1]/\mathbb{E}_t[\alpha_1^2] > 1$. Unlike the idealized $\kappa=2$ case (\cref{fig:shared-error}a), XS repeats only about $0.6$ of S's error across all noise levels and adds a comparable orthogonal error (\cref{fig:shared-error}b,c), so the scale closest to $D^\star$ stays below $1$ (\cref{fig:shared-error}d, matching \cref{fig:wmse-sigma}). The same holds per input (bands in \cref{fig:shared-error}c) and for EDM2-L guided by S or M. Using XL as a proxy is conservative, since its own error enters both differences identically and pulls $\kappa$ toward $1$. The only exception we found is guiding S with a longer EMA from the same run, where shared training aligns the errors and lifts $w^\star$ above $1$. For separate models, each denoiser's error is largely independent, which explains $w_{\mathrm{MSE}}^\star < 1$ and supports \cref{sec:vm-guidance}: $D_1 - D_0$ acts as a random vector that guidance rescales to match the posterior covariance (\cref{sec:second-moment}).

\subsection{Why CFG and autoguidance differ at high noise}
\label{app:vgap}

Our interpretation of guidance relies on the two predictions having compatible means and covariance directions, so that rescaling their difference can restore the posterior covariance (\cref{sec:second-moment}). Here, we examine how these properties change throughout sampling. We first follow the two CFG predictions of SD~3.5 Medium for one prompt, which align in the noise range where guidance helps, and then compare CFG with autoguidance for one ImageNet class.

\begin{figure*}[!t]
    \centering
    \includegraphics[width=\textwidth]{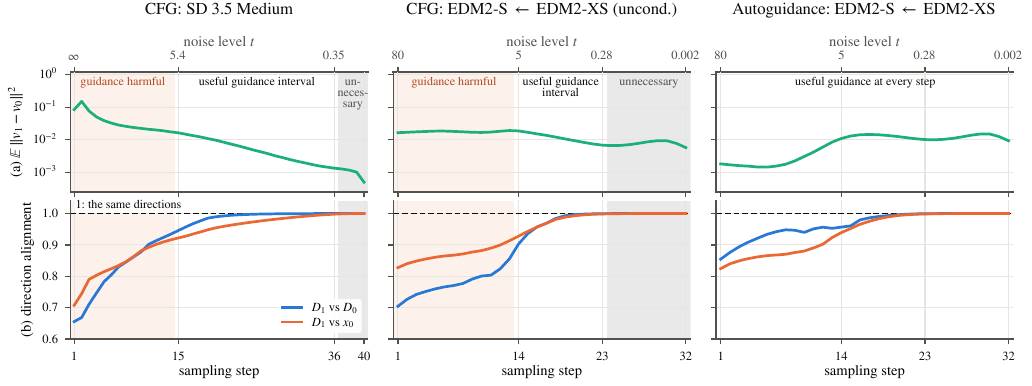}
    \caption{The two predictions of a guided pair along the sampler, without guidance ($w=1$), over $1{,}000$ seeds; $x_0$ are the final samples. Left: SD~3.5 Medium, one prompt. Middle and right: EDM2-S on ImageNet-64, one class (school bus), $32$ Heun steps, guided by CFG or by autoguidance with EDM2-XS. $D_1$ is the conditional prediction for CFG and EDM2-S for autoguidance; $D_0$ is the unconditional prediction or EDM2-XS. (a)~$\E_{x_t}\|v_1-v_0\|^2$ per coordinate, over the $1{,}000$ states $x_t$ at each step; shading marks where CFG harms or is unnecessary according to \citet{kynkaanniemi2024interval}. (b)~Direction alignment: the overlap of the top-$50$ principal subspaces ($1$: the same directions).}
    \label{fig:cov-trajectory}\label{fig:cov-trajectory-in64}
\end{figure*}

\paragraph{CFG predictions align as noise decreases.}
CFG is known to harm sample quality at high noise and help primarily within a middle range of noise levels \citep{kynkaanniemi2024interval}. Our view suggests an explanation: at high noise, the conditional and unconditional predictions may differ substantially in both their values and covariance directions, making their difference unsuitable for restoring the posterior covariance. To examine this, the left column of \cref{fig:cov-trajectory} follows the two predictions of SD~3.5 Medium over $1{,}000$ unguided sampling trajectories with the same prompt. As noise decreases, their mean squared velocity difference $\E_{x_t}\|v_c-v_u\|^2$, averaged over the $1{,}000$ states $x_t$ at each step, falls tenfold during the high-noise steps where guidance harms, and another tenfold through the interval where it helps (\cref{fig:cov-trajectory}a). At the same time, the overlap of their top-$50$ principal covariance subspaces approaches $1$, indicating that their predictions vary along increasingly similar directions (\cref{fig:cov-trajectory}b). Thus, the noise range where CFG helps is also where its two predictions become close and their covariance directions align, consistent with our variance-matching interpretation.

\paragraph{Autoguidance remains aligned at high noise.}
If this mismatch explains why CFG harms at high noise, it should be less pronounced in autoguidance. Unlike CFG, whose conditional and unconditional branches target different distributions, autoguidance combines two models trained toward the same target, differing primarily in their approximation errors. We test this prediction on ImageNet-64 using EDM2-S with CFG and with EDM2-XS autoguidance (middle and right columns of \cref{fig:cov-trajectory}). CFG shows the same pattern as SD~3.5: its predictions are farther apart and their covariance directions differ at high noise, becoming aligned only later in sampling. In contrast, autoguidance's velocity predictions are about ten times closer during the high-noise steps, and their covariance directions are already aligned. This agrees with the finding that CFG improves when the high-noise steps are excluded \citep{kynkaanniemi2024interval}, whereas autoguidance works throughout sampling \citep{karras2024guiding}.

\subsection{Guidance reduces diversity}
\label{app:diversity}

Our interpretation in \cref{sec:second-moment} treats the errors of a model pair as random samples from the posterior covariance. However, \cref{eq:M(2)} draws fresh noise at every step, whereas a fixed pair of trained models produces a deterministic error field: nearby inputs receive similar errors. The spatial range of these correlations depends on the noise level. At large $\sigma$, the model cannot distinguish fine details, and its errors are shared across a broad region of input space. Guidance amplifies these shared errors, potentially directing nearby sampling trajectories toward the same modes rather than letting them explore different ones.

\paragraph{Correlated errors can reduce diversity without reducing fidelity.}
We hypothesize that shared errors push nearby trajectories toward the same modes, reducing coverage while preserving fidelity: the samples still land in high-likelihood regions, but fewer modes receive substantial probability. To isolate this effect, we replace the independent noise $\varepsilon$ in \cref{eq:M(2)} with a Gaussian error field $g(x,\sigma)$ whose covariance is
\begin{equation}
\begin{aligned}
    &\Cov_g\!\big(g(x,\sigma),g(x',\sigma')\big)\\
    &\quad=\exp\!\Big(
    -\frac{\|x-x'\|^2}{2\ell_x^2}
    -\frac{(\log\sigma-\log\sigma')^2}{2\ell_t^2}
    \Big)I.
\end{aligned}
\label{eq:error-field}
\end{equation}

Its length scales $\ell_x$ and $\ell_t$ set how far the field stays correlated across inputs and across $\log\sigma$, and \cref{eq:M(2)} is the limit $\ell_x=\ell_t=0$. On the two-dimensional mixture, a single fixed field concentrates the samples on fewer modes as $\ell_x$ and $\ell_t$ grow, as guidance with a trained pair does as $w$ grows (\cref{fig:diversity-ladder}). We test whether this effect persists in high dimensions, on $1{,}000$ ImageNet-$64$ images, one per class, with isotropic Gaussians as a multimodal distribution with exact posterior moments and likelihood (Appendix~\ref{sec:moments}). We run \cref{eq:M(2)} with correlated error fields and measure fidelity by likelihood and diversity by coverage \citep{naeem2020reliable}.

\begin{figure}[t]
  \centering
  \includegraphics[width=\linewidth]{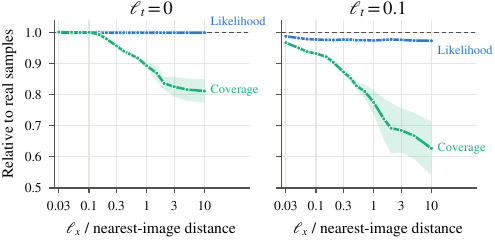}
  \caption{Likelihood and coverage of \cref{eq:M(2)} on $1{,}000$ ImageNet-$64$ images, each replaced by a Gaussian, with $\varepsilon$ replaced by the error field $g$ of \cref{eq:error-field}, against $\ell_x$ in units of the median distance between nearest images. Left: a new field at every step ($\ell_t=0$); right: $\ell_t=0.1$. Likelihood: exact likelihood of the samples under the data distribution, per dimension. Coverage: $k=3$, in a fixed random $64$-dimensional projection. Both relative to real samples (dashed). $K=128$ steps, $100{,}000$ samples; lines: mean over three fields; bands: $\pm1$ standard deviation.}
  \label{fig:diversity-lx}
\end{figure}

In \cref{fig:diversity-lx}, even without temporal correlations ($\ell_t=0$), increasing $\ell_x$ substantially reduces coverage while likelihood remains at the level of real samples. Introducing temporal correlations ($\ell_t=0.1$) further reduces coverage, although sufficiently strong correlations also lower likelihood. In this synthetic setting, spatially correlated noise therefore concentrates the samples on fewer modes without lowering their likelihood.

\paragraph{Trained models have correlated errors.}
We measure $\ell_x$ and $\ell_t$ for CFG on ImageNet-$64$, with EDM2-S as $D_1$ and an unconditional EDM2-XS as $D_0$, from the difference $\Delta=D_1-D_0$ that guidance amplifies. For $\ell_x$, we take two inputs of the same class at the same noise level and compute the cosine similarity of their values of $\Delta$; it is close to $1$ for nearby inputs and decays with their distance, and $\ell_x$ is the distance at which it reaches $e^{-1/2}\approx0.61$, the value of the kernel in \cref{eq:error-field} at distance $\ell_x$. We obtain $\ell_t$ the same way from one input at two noise levels, with the distance measured in $\log\sigma$.

\begin{figure}[t]
  \centering
  \includegraphics[width=\linewidth]{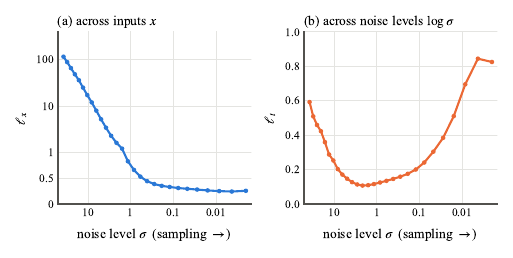}
  \caption{Correlation lengths of $\Delta=D_1-D_0$ (CFG: the class-conditional EDM2-S, minus an unconditional EDM2-XS) on ImageNet-$64$, at every step of the $32$-step EDM2 sampler from step $5$ ($\sigma=38$) on; before it, the correlation stays above $e^{-1/2}$ even between independent inputs. (a) $\ell_x$, in units of the median distance between nearest images, from $1{,}000$ noisy inputs of one class with $15$ partners each, ranging from nearly identical to independent inputs; linear axis below $1$, logarithmic above. (b) $\ell_t$, in $\log\sigma$, at the same inputs with the networks given a second noise level above or below $\sigma$ (both averaged).}
  \label{fig:errfield-lx}
\end{figure}

\Cref{fig:errfield-lx} shows both lengths along the sampler: $\ell_x$ decays as the noise decreases, and $\ell_t$ stays above $0.1$ at every step. Running \cref{eq:M(2)} in the experiment of \cref{fig:diversity-lx} with exactly these $\ell_x(\sigma)$ and $\ell_t(\sigma)$ gives a likelihood of $0.91$ and a coverage of $0.47$ relative to real samples.

\subsection{Guidance in both directions}
\label{app:both-directions}

\begin{figure}[t]
  \centering
  \begin{subfigure}{\linewidth}
    \centering
    \includegraphics[width=\linewidth]{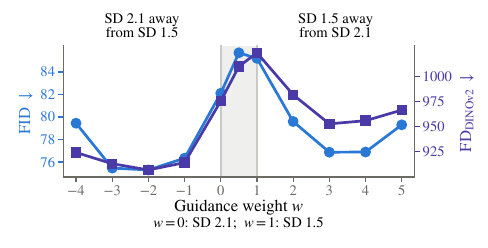}
    \caption{SD~1.5 and SD~2.1}
    \label{fig:cross-sd15-sd21}
  \end{subfigure}\\[4pt]
  \begin{subfigure}{\linewidth}
    \centering
    \includegraphics[width=\linewidth]{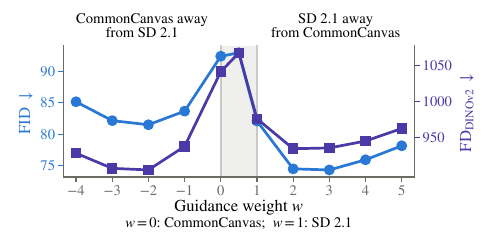}
    \caption{SD~2.1 and CommonCanvas}
    \label{fig:cross-sd21-cc}
  \end{subfigure}
  \caption{Bidirectional guidance for the two other pairs of models that share SD~1.5's latent space. The models sit at $w=0$ and $w=1$ (shaded); scales outside $[0,1]$ extrapolate away from either model. FID (circles) and FD$_\mathrm{DINOv2}$ (squares) on $1{,}000$ held-out COCO captions.}
  \label{fig:cross-pairs}
\end{figure}

\begin{figure}[t]
  \centering
  \includegraphics[width=\linewidth]{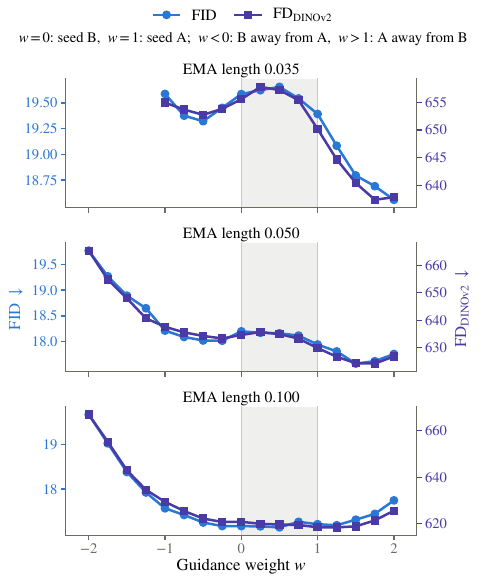}
  \caption{Bidirectional guidance between two EDM2-XS models trained with different seeds, $A$ and $B$, at three EMA lengths. $D=D_B+w(D_A-D_B)$, with $B$ at $w=0$ and $A$ at $w=1$. FID (circles) and FD$_\mathrm{DINOv2}$ (squares) on ImageNet-64, using $10{,}000$ samples per scale. Each panel has its own axis range; scales below $-1$ were not tested at EMA $0.035$.}
  \label{fig:cross-edm2xs}
\end{figure}

\paragraph{Bidirectional gains across models.}
\Cref{sec:exp-hypotheses} shows that guiding SD~1.5 and CommonCanvas away from each other improves both models, contrary to a simple error-correction explanation. We find the same with the two other pairs that share SD~1.5's latent space, SD~2.1 with SD~1.5 and with CommonCanvas (\cref{fig:cross-pairs}): every tested scale outside $[0,1]$, from $-4$ to $5$, improves both FID and FD$_\mathrm{DINOv2}$ in all four directions, with $95\%$ confidence intervals excluding zero. Although SD~2.1 is the stronger model in both pairs, guiding the weaker model away from it matches or surpasses its quality. In contrast, averaging the models ($w=0.5$) never outperforms the better one.

\paragraph{Even different seeds benefit.}
The SD models differ in their training data, so their gains might still reflect systematic differences. We therefore train two EDM2-XS models on ImageNet-64 with the same recipe ($2^{25}$ training images) but different seeds, and evaluate them with the 32-step Heun sampler using shared noise and labels. Neither model has a systematic advantage, yet extrapolation improves both FID and FD$_\mathrm{DINOv2}$ in both directions at EMA lengths $0.035$ and $0.050$ (\cref{fig:cross-edm2xs}). At $0.050$, the best FID gains are about $0.4$ for $A$ and $0.2$ for $B$, while averaging again offers no gain. This is consistent with our interpretation: even independent estimates of the same posterior mean can provide useful differences for guidance (\cref{sec:second-moment}).

\paragraph{Longer EMA removes the gain.}
As the EMA length increases, the benefit of guidance disappears (\cref{fig:cross-edm2xs}). The best FID gain on $A$'s side drops from $0.8$ at EMA $0.035$ to $0.4$ at $0.050$, and vanishes at $0.100$. Moreover, guiding the shorter-EMA models never outperforms the longer-EMA models alone. This suggests that between seeds, guidance amplifies the variation left by stochastic training, which longer EMA already averages away (\cref{app:ema}).

\subsection{Metric trade-offs}
\label{app:frontier}

\begin{figure}[!ht]
    \centering
    \includegraphics[width=\linewidth]{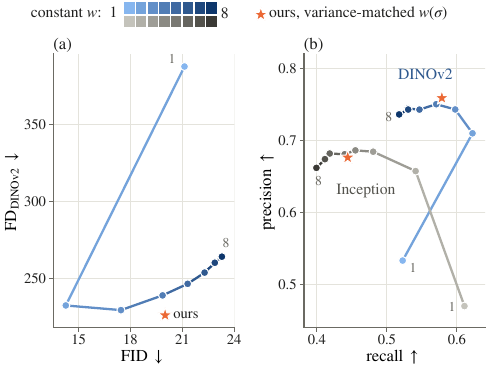}
    \caption{Quality trade-offs for SD~3.5 Medium on COCO. Shade darkens with the constant guidance scale $w=1,\ldots,8$; the star is our variance-matched schedule $w(\sigma)$. Panel (b) shows precision against recall in DINOv2 (blue) and Inception (gray) features. The schedule has slightly higher DINOv2 precision and lower $\mathrm{FD}_{\mathrm{DINOv2}}$ than every tested constant, with DINOv2 recall close to that of $w=3$--$4$.}
    \label{fig:frontier-sd35m}
\end{figure}

As shown in \cref{sec:exp-metrics}, different metrics favor different guidance scales, reflecting trade-offs across evaluation criteria. \Cref{fig:frontier-sd35m} plots these trade-offs for SD~3.5 Medium on COCO as the constant guidance scale $w$ varies from $1$ to $8$, alongside our variance-matched schedule $w(\sigma)$ (star). Increasing $w$ beyond $2$ raises precision while worsening FID and recall, illustrating the tension between different sample metrics.

\subsection{On the effectiveness of VMG on ImageNet}
\label{app:edm2-fail}

\begin{table}[t]
\centering
\caption{EDM2-XXL on ImageNet-512 at the scales NVIDIA tuned for FID (Pub.) and at our variance-matched constant (Ours) and schedule (Ours, $w(\sigma)$), fitted to each pair on $1{,}024$ validation images. $10{,}000$ samples per row with the same seeds, EDM2's 32-step Heun sampler; FID and FD$_\mathrm{DINOv2}$ against the training set, precision (P) and recall (R) with $k=3$ against $10{,}000$ validation images. Bold: best of each block's three rows.}
\label{tab:edm2-fail}
\scriptsize\setlength{\tabcolsep}{1.5pt}\renewcommand{\arraystretch}{0.9}
\begin{tabular*}{\linewidth}{@{}l@{\hspace{3pt}\extracolsep{\fill}}rrrrrrr@{}}
\toprule
 & & \multicolumn{3}{c}{Inception} & \multicolumn{3}{c}{DINOv2} \\
\cmidrule(lr){3-5}\cmidrule(lr){6-8}
 & \hd{$w$} & \hd{FID$\downarrow$} & \hd{P$\uparrow$} & \hd{R$\uparrow$} & \hd{FD$\downarrow$} & \hd{P$\uparrow$} & \hd{R$\uparrow$} \\
\midrule
\multicolumn{8}{@{}l}{\textit{Classifier-free guidance: EDM2-XXL $\leftarrow$ EDM2-XS (uncond.)}} \\
Pub. & 1.20 & {\bfseries 4.83} & 0.793 & {\bfseries 0.685} & {\bfseries 90.2} & 0.799 & {\bfseries 0.784} \\
Ours & 5.73 & 11.98 & {\bfseries 0.825} & 0.435 & 120.7 & {\bfseries 0.875} & 0.485 \\
Ours ($w(\sigma)$) & \na & 11.84 & 0.795 & 0.443 & 133.1 & 0.858 & 0.468 \\
\midrule
\multicolumn{8}{@{}l}{\textit{Autoguidance: EDM2-XXL $\leftarrow$ EDM2-M}} \\
Pub. & 2.05 & {\bfseries 4.41} & {\bfseries 0.751} & 0.737 & {\bfseries 64.8} & {\bfseries 0.807} & {\bfseries 0.836} \\
Ours & 9.45 & 30.20 & 0.432 & 0.766 & 178.3 & 0.684 & 0.771 \\
Ours ($w(\sigma)$) & \na & 14.47 & 0.539 & {\bfseries 0.794} & 124.2 & 0.716 & 0.818 \\
\bottomrule
\end{tabular*}
\end{table}

Our rule gives larger scales than those published for EDM2 on ImageNet-512: for EDM2-XXL guided by an unconditional EDM2-XS (classifier-free guidance) and by EDM2-M (autoguidance), our constants are about five times the published ones (\cref{tab:edm2-fail}). Unlike most published scales, whose choice is undocumented (\cref{sec:exp-metrics}), these come from a grid search over the scale and EMA lengths, run separately for FID and for FD$_\mathrm{DINOv2}$ \citep{karras2024edm2,karras2024guiding}; \cref{tab:edm2-fail} uses the FID-optimal ones. We think this is why the two differ, as these metrics need not track what people prefer: on SD~3.5 Medium, users rated the scales tuned to FID and FD$_\mathrm{DINOv2}$ below ours (\cref{tab:tuned-w,tab:user-study}). We ran no user study on ImageNet; \cref{tab:edm2-fail} reports every metric at the published scales and with our constant and schedule, and \cref{fig:samples-edm2-cfg,fig:samples-edm2-ag} show $100$ pairs at the published scales and with our schedule.

\section{Ablations}
\label{app:ablations}

This section ablates the choices behind variance-matched guidance (\cref{sec:vm-guidance}): matching only the trace of the covariance rather than the whole matrix (\cref{app:fullcov}), pooling the per-step mismatch into a constant scale by its average rather than by other fits (\cref{app:objective}), the number of calibration examples (\cref{app:ncalib}), and adding variance along $D_1-D_0$ rather than with noise that ignores the input (\cref{app:noise-control}).

\subsection{Matching the full covariance}
\label{app:fullcov}

In \cref{sec:vm-guidance}, we choose $w$ by matching only the trace of the covariance in \cref{eq:total-cov}: \cref{eq:varmatch_t} sets the total variance of the guided prediction over inputs to that of the data. Here, we ablate this choice by fitting the same scale $w$ to the whole matrix instead, and compare the two schedules. In our model they agree. By \cref{eq:cov-delta}, guidance multiplies the posterior covariance by the scalar $c(w,m_1,m_0)$, so with $m_0,m_1$ constant across $x_t$ the guided covariance in \cref{eq:total-cov} differs from $\Cov(x_0)$ by $(c-1)\,\E_{x_t}[\Cov(x_0|x_t)]$: $w$ sets its size but not its shape, and the scale that matches the trace matches every entry. The scale fitted to the whole matrix is
\[
w^\star_{\mathrm{cov}}(t)=\argmin_w\,\bigl\|M(w)-M_{\mathrm{data}}\bigr\|_F,
\]
with $M(w)=\mathbb{E}_{x_t\sim p_t}(D_w-\bar x_0)(D_w-\bar x_0)^\top$ and $M_{\mathrm{data}}=\mathbb{E}_{x_0\sim p_{\mathrm{data}}}(x_0-\bar x_0)(x_0-\bar x_0)^\top$.

\begin{figure}[t]
  \centering
  \includegraphics[width=\linewidth]{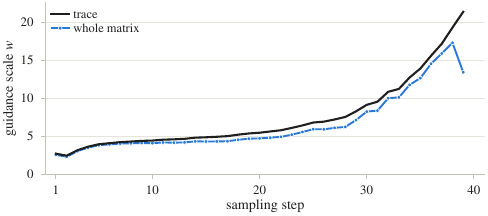}
  \caption{Guidance scale at each of the 40 Euler steps of SD~3.5 Medium on COCO, from matching the trace \eqref{eq:varmatch_t} (black) or fitting $w$ to the whole $16\times16$ covariance between latent channels (blue). The last step, where neither rule has a solution, is omitted.}
  \label{fig:covmatch}
\end{figure}

We fit it on SD~3.5 Medium with the $5{,}705$ COCO photos and $40$ noise levels of \cref{app:impl}, using the $16\times16$ covariance between latent channels, summed over spatial positions. The scale fitted to the whole matrix is very similar to the one that matches the trace (\cref{fig:covmatch}). This suggests that our model of the guided covariance (\cref{eq:cov-delta}) holds up to small deviations, so matching the trace is enough.

\subsection{Alternative objectives for the constant scale}
\label{app:objective}

In \cref{sec:vm-guidance}, we turn the per-step rule into a constant scale by matching the variance averaged over the sampler's steps (\cref{eq:varmatch}). A constant scale cannot match the variance of the data at every noise level, so this is one of several ways to pool the per-step mismatch. Here, we ablate this choice by comparing it with two standard fits of the mismatch, judged by how close each lands to the published scales. Let
\begin{equation}
r_t(w)=\underset{x_t\sim p_t}{\Var}[D_w(x_t)]-\underset{x_0\sim p_{\mathrm{data}}}{\Var}[x_0]
\label{eq:obj-r}
\end{equation}
be the mismatch at level $t$. We compare three ways to choose the scale from it:
\begin{align}
\text{Ours:}\quad & \E_t\bigl[r_t(w)\bigr]=0, \label{eq:obj-ours}\\
L_1\text{:}\quad & \min_{w>0}\ \E_t\bigl|r_t(w)\bigr|, \label{eq:obj-l1}\\
L_2\text{:}\quad & \min_{w>0}\ \E_t\bigl[r_t(w)^2\bigr], \label{eq:obj-l2}
\end{align}
with $t$ uniform over the sampler's steps. The first is our matching condition \eqref{eq:varmatch} written in terms of $r_t$, with the closed-form solution \eqref{eq:wconstant}; the other two are the least-absolute-error and least-squares fits. Least squares is less robust to outliers, which $r_t$ has (the first, noisiest sampling steps, where $D_1$ and $D_0$ differ most).

\begin{table}[H]
\centering
\caption{The constant scale under the three objectives \eqref{eq:obj-ours}--\eqref{eq:obj-l2}, next to the published scale (Pub.; for SVD the mean of its per-frame ramp). Bold marks the candidate closest to Pub. The last row is the median relative distance to Pub.}
\label{tab:objective}
\footnotesize
\begin{tabular}{@{}lcccc@{}}
\toprule
Model & Ours & $L_1$ & $L_2$ & Pub. \\
\midrule
SD~1.5 & \textbf{7.44} & $6.88$ & $6.74$ & $7.5$ \\
SD~3.5 Medium & \textbf{4.14} & $3.79$ & $3.24$ & $4.5$ \\
SD~3.5 Large & \textbf{2.88} & $2.65$ & $2.66$ & $3.5$ \\
FLUX.2 [klein] 4B & $4.74$ & $4.18$ & \textbf{4.02} & $4.0$ \\
SVD & \textbf{1.50} & $1.46$ & $1.47$ & $2.0$ \\
Wan2.1 & \textbf{4.40} & $3.49$ & $3.46$ & $5.0$ \\
CogVideoX-2B & \textbf{6.22} & $4.48$ & $3.82$ & $6.0$ \\
Cosmos-Predict2.5 & $1.00$ & $1.00$ & $1.00$ & $1.0$ \\
AudioLDM~2 & \textbf{4.11} & $2.84$ & $2.86$ & $3.5$ \\
Stable Audio Open & $7.88$ & $6.26$ & \textbf{6.58} & $7.0$ \\
MDM & \textbf{2.23} & $1.56$ & $1.63$ & $2.5$ \\
\midrule
Median error & \textbf{12\%} & $19\%$ & $24\%$ & \\
\bottomrule
\end{tabular}
\end{table}

The choice matters for our main claim, that the constant scale recovers the scales used in practice (\cref{fig:wone}). \Cref{tab:objective} compares the three scales with the published one on the models of \cref{fig:wone}. Our condition \eqref{eq:obj-ours} is the closest to the published scale on eight of the eleven models and has the smallest median error, while the other two give smaller scales than ours on every model but Cosmos-Predict2.5. Matching the average variance is therefore both the simplest of the three objectives, with a closed form, and the one that best recovers the scales chosen in practice.

\subsection{How many calibration examples the fit needs}
\label{app:ncalib}

In \cref{sec:experiments}, we fit the constant scale (\cref{eq:wconstant}) from every example of a calibration set, $5{,}705$ COCO photos for SD~3.5 Medium. Here, we ask how many examples the fit actually needs, since this sets the cost of our rule and whether it applies when little data is at hand. We refit the constant scale on random subsets of the calibration set, for SD~3.5 Medium ($40$ levels) and AudioLDM~2 ($2{,}223$ AudioCaps captions, $200$ levels), keeping every level and each example's noise draws. Each subset is fitted from its own examples alone, including the means in \cref{eq:vm-coefficients}, and we draw $2{,}000$ subsets per size.

\begin{table}[H]
\centering
\caption{The constant scale fitted from $n$ calibration examples: SD~3.5 Medium on $n$ of its $5{,}705$ COCO photos ($40$ levels) and AudioLDM~2 on $n$ of its $2{,}223$ AudioCaps captions of $445$ clips ($200$ levels). Network evaluations are $n$ times the levels times $2$ predictions. The standard deviation and the central 95\% range are over $2{,}000$ random subsets per size, and both are corrected for the overlap of large subsets drawn from the same calibration set.}
\label{tab:ncalib}
\footnotesize
\begin{tabular}{@{}rrcc@{}}
\toprule
\begin{tabular}[b]{@{}r@{}}Examples\\$n$\end{tabular} & \begin{tabular}[b]{@{}r@{}}Network\\evaluations\end{tabular} & \begin{tabular}[b]{@{}c@{}}Standard\\deviation of $w$\end{tabular} & \begin{tabular}[b]{@{}c@{}}95\% range\\of $w$\end{tabular} \\
\midrule
\multicolumn{4}{@{}l}{\emph{SD~3.5 Medium, COCO photos}} \\
$16$ & $1{,}280$ & $0.23$ & $3.72$--$4.62$ \\
$32$ & $2{,}560$ & $0.17$ & $3.82$--$4.49$ \\
$64$ & $5{,}120$ & $0.12$ & $3.91$--$4.36$ \\
$128$ & $10{,}240$ & $0.08$ & $3.98$--$4.29$ \\
$256$ & $20{,}480$ & $0.06$ & $4.02$--$4.25$ \\
$512$ & $40{,}960$ & $0.04$ & $4.05$--$4.21$ \\
$1{,}024$ & $81{,}920$ & $0.03$ & $4.07$--$4.19$ \\
$2{,}048$ & $163{,}840$ & $0.02$ & $4.09$--$4.17$ \\
\midrule
\multicolumn{4}{@{}l}{\emph{AudioLDM~2, AudioCaps captions}} \\
$16$ & $6{,}400$ & $0.45$ & $3.32$--$5.04$ \\
$32$ & $12{,}800$ & $0.31$ & $3.58$--$4.77$ \\
$64$ & $25{,}600$ & $0.22$ & $3.73$--$4.58$ \\
$128$ & $51{,}200$ & $0.16$ & $3.83$--$4.43$ \\
$256$ & $102{,}400$ & $0.11$ & $3.89$--$4.33$ \\
$512$ & $204{,}800$ & $0.08$ & $3.97$--$4.27$ \\
$1{,}024$ & $409{,}600$ & $0.06$ & $4.01$--$4.23$ \\
$2{,}048$ & $819{,}200$ & $0.04$ & $4.04$--$4.20$ \\
\bottomrule
\end{tabular}
\end{table}

The spread of the fitted scale falls as $1/\sqrt{n}$ over the whole range (\cref{tab:ncalib}), and the audio model needs about $3.5\times$ as many examples as the image model for the same relative precision. About $250$ examples give the image scale to within $\pm3\%$ and AudioLDM~2's to within $\pm5\%$. A few hundred calibration examples, a small fraction of the sets we use, are therefore enough to fit the scale.

\subsection{Noise with the right variance but not the posterior's covariance}
\label{app:noise-control}

In \cref{sec:more-moments}, we argue that the variance added to the posterior mean must follow the posterior covariance $\Sigma(x_t)=\Cov_{x_0\sim p(x_0|x_t)}(x_0)$, as in the sampler of \cref{eq:M(2)}: the right amount of variance with the wrong shape does not help. Guidance adds variance along $D_1-D_0$, which follows $\Sigma(x_t)$ at each input (\cref{eq:cov-delta}), while our rule only sets how much (\cref{sec:vm-guidance}). Here, we ablate the direction by adding the same amount of variance with noise that ignores $x_t$. We expect this to fail, because $\Sigma(x_t)$ depends on the input: it is large along the directions in which $x_t$ leaves $x_0$ uncertain and small along those it already determines. Noise that ignores $x_t$ puts variance in directions the posterior has already settled, pushing samples off the data.

To test this on SD~3.5 Medium with the COCO protocol of \cref{app:impl}, we sample without guidance and at every Euler step replace the prediction by $D_1(x_t)+\eta$, with $\eta$ zero-mean Gaussian, independent of $x_t$ and drawn afresh at each step. Its variance restores that of the data at each step, as in \cref{eq:varmatch_t}, so it adds as much variance as our schedule does. The noise is either isotropic or shaped like the data covariance $\Cov_{x_0\sim p_{\mathrm{data}}}(x_0)$.

\begin{table}[H]
\centering
\caption{Matching the variance without the guidance direction, on SD~3.5 Medium (COCO, 10,000 images per row, 40 Euler steps). The noise rows sample without guidance and add Gaussian noise to the prediction at every step, as much variance as our schedule adds. All rows share prompts and initial noise.}
\label{tab:noise-control}
\footnotesize
\setlength{\tabcolsep}{2pt}
\begin{adjustbox}{max width=\linewidth}
\begin{tabular}{@{}lrrrrrrr@{}}
\toprule
 & & & \multicolumn{2}{c}{DINOv2} & \multicolumn{2}{c}{Inception} & \\
\cmidrule(lr){4-5}\cmidrule(lr){6-7}
Sampling & FID $\downarrow$ & FD$_\mathrm{DINOv2}$ $\downarrow$ & P $\uparrow$ & R $\uparrow$ & P $\uparrow$ & R $\uparrow$ & CLIP $\uparrow$ \\
\midrule
No guidance ($w=1$) & 21.20 & 388.5 & 0.542 & 0.524 & 0.464 & 0.612 & 0.2409 \\
\quad + isotropic noise & 31.68 & 422.0 & 0.507 & 0.502 & 0.401 & 0.541 & 0.2370 \\
\quad + data-shaped noise & 36.64 & 464.0 & 0.525 & 0.438 & 0.362 & 0.528 & 0.2369 \\
Ours, $w^\star_{\mathrm{VM}}(t)$ & 20.13 & 224.0 & 0.765 & 0.578 & 0.667 & 0.440 & 0.2749 \\
\bottomrule
\end{tabular}
\end{adjustbox}
\end{table}

Both kinds of noise make every metric worse than sampling without guidance, and the data-shaped noise is mostly the worse of the two (\cref{tab:noise-control}). Our schedule, which adds the same variance along $D_1-D_0$, improves on no guidance in nearly every metric. Added variance therefore helps only when it follows the posterior covariance at each input, as \cref{sec:more-moments} claims.

\section{Full results and user study}
\label{app:results}

\subsection{Published guidance versus our schedule: every metric}
\label{app:extra-models}

\Cref{tab:full-image,tab:full-video,tab:full-motion,tab:full-audio} report every metric behind \cref{tab:pub-all}, with the same rows and seeds; protocols are in \cref{app:impl}. Pub.\ is the published scale, rerun under our protocol. Ours is colored as in \cref{tab:pub-all}.

\begin{table}[H]
\centering
\caption{Text-to-image, every metric ($10{,}000$ held-out prompts per model). $k$-NN metrics use $k=3$.}
\label{tab:full-image}
\footnotesize
\setlength{\tabcolsep}{2pt}
\begin{adjustbox}{max width=\linewidth}
\begin{tabular}{@{}lrrrrrrrr@{}}
\toprule
 & \multicolumn{2}{c}{SD 1.5} & \multicolumn{2}{c}{SD 3.5 M} & \multicolumn{2}{c}{SD 3.5 L} & \multicolumn{2}{c}{FLUX.2 [klein] 4B} \\
 & \multicolumn{2}{c}{\data{COCO}} & \multicolumn{2}{c}{\data{COCO}} & \multicolumn{2}{c}{\data{COCO}} & \multicolumn{2}{c}{\data{MJHQ}} \\
\cmidrule(lr){2-3}\cmidrule(lr){4-5}\cmidrule(lr){6-7}\cmidrule(lr){8-9}
 & \hd{Pub.\ (7.5)} & \hd{Ours} & \hd{Pub.\ (4.5)} & \hd{Ours} & \hd{Pub.\ (3.5)} & \hd{Ours} & \hd{Pub.\ (4.0)} & \hd{Ours} \\
\midrule
\multicolumn{9}{@{}l}{\textit{Inception}} \\
\quad FID $\downarrow$ & 17.51 & {\color{vmgood}17.27} & 20.58 & {\color{vmgood}20.13} & 21.00 & {\color{vmgood}20.33} & 19.48 & {\color{vmbad}19.69} \\
\quad P $\uparrow$ & 0.629 & {\bfseries\color{vmbad}0.585} & 0.685 & {\color{vmbad}0.667} & 0.595 & {\color{vmbad}0.570} & 0.627 & {\color{vmbad}0.610} \\
\quad R $\uparrow$ & 0.482 & {\bfseries\color{vmgood}0.513} & 0.461 & {\color{vmbad}0.440} & 0.448 & {\color{vmgood}0.470} & 0.536 & {\color{vmgood}0.545} \\
\quad Dens. $\uparrow$ & 0.841 & {\bfseries\color{vmbad}0.742} & 0.944 & {\color{vmbad}0.915} & 0.746 & {\bfseries\color{vmbad}0.686} & 0.785 & {\color{vmbad}0.746} \\
\quad Cov. $\uparrow$ & 0.678 & {\color{vmbad}0.652} & 0.713 & {\color{vmbad}0.695} & 0.630 & {\color{vmbad}0.612} & 0.682 & {\color{vmbad}0.662} \\
\addlinespace
\multicolumn{9}{@{}l}{\textit{DINOv2}} \\
\quad FD $\downarrow$ & 233.3 & {\color{vmbad}239.4} & 242.4 & {\bfseries\color{vmgood}224.0} & 323.0 & {\bfseries\color{vmgood}301.0} & 197.9 & {\color{vmgood}194.0} \\
\quad P $\uparrow$ & 0.691 & {\color{vmbad}0.663} & 0.744 & {\color{vmgood}0.765} & 0.710 & {\color{vmgood}0.721} & 0.754 & 0.747 \\
\quad R $\uparrow$ & 0.597 & {\bfseries\color{vmbad}0.567} & 0.555 & {\color{vmgood}0.578} & 0.438 & {\color{vmgood}0.453} & 0.779 & 0.776 \\
\quad Dens. $\uparrow$ & 0.605 & {\color{vmbad}0.576} & 0.662 & {\bfseries\color{vmgood}0.708} & 0.610 & {\color{vmgood}0.635} & 0.637 & {\color{vmbad}0.626} \\
\quad Cov. $\uparrow$ & 0.526 & {\color{vmbad}0.508} & 0.573 & {\color{vmgood}0.594} & 0.503 & {\color{vmgood}0.511} & 0.654 & {\color{vmbad}0.644} \\
\addlinespace
CLIP $\uparrow$ & 0.2690 & 0.2699 & 0.2715 & {\color{vmgood}0.2749} & 0.2690 & 0.2705 & 0.2746 & 0.2745 \\
\bottomrule
\end{tabular}
\end{adjustbox}
\end{table}

\begin{table}[H]
\centering
\caption{Video and world model, every metric. $k$-NN metrics use I3D features. Cosmos-Predict2.5 is rolled out over 100 Bridge test episodes with its released configuration ($w=1$); PSNR and SSIM average predicted frames $1$ to $N{-}1$, close to its reported 24.95 and 0.85. Real OpenVid-1M clips score CLIP-T 28.2.}
\label{tab:full-video}
\footnotesize
\setlength{\tabcolsep}{2pt}
\begin{adjustbox}{max width=\linewidth}
\begin{tabular}{@{}lrrrrrrrr@{}}
\toprule
 & \multicolumn{2}{c}{SVD} & \multicolumn{2}{c}{Wan2.1} & \multicolumn{2}{c}{CogVideoX-2B} & \multicolumn{2}{c}{Cosmos-Predict2.5} \\
 & \multicolumn{2}{c}{\data{UCF-101}} & \multicolumn{2}{c}{\data{OpenVid}} & \multicolumn{2}{c}{\data{OpenVid}} & \multicolumn{2}{c}{\data{Bridge}} \\
\cmidrule(lr){2-3}\cmidrule(lr){4-5}\cmidrule(lr){6-7}\cmidrule(lr){8-9}
 & \hd{Pub.\ (1$\to$3)} & \hd{Ours} & \hd{Pub.\ (5)} & \hd{Ours} & \hd{Pub.\ (6)} & \hd{Ours} & \hd{Pub.\ (1)} & \hd{Ours} \\
\midrule
\multicolumn{9}{@{}l}{\textit{I3D}} \\
\quad FVD $\downarrow$ & 141.2 & {\bfseries\color{vmgood}113.7} & 261.0 & {\bfseries\color{vmgood}220.0} & 236.3 & {\bfseries\color{vmbad}262.1} & 62.6 & 63.0 \\
\quad P $\uparrow$ & 0.804 & {\color{vmgood}0.816} & 0.589 & {\color{vmgood}0.606} & 0.589 & {\bfseries\color{vmbad}0.555} & \na & \na \\
\quad R $\uparrow$ & 0.817 & {\color{vmgood}0.838} & 0.635 & {\bfseries\color{vmgood}0.677} & 0.701 & {\bfseries\color{vmgood}0.752} & \na & \na \\
\quad Dens. $\uparrow$ & 0.883 & {\bfseries\color{vmgood}0.972} & 0.770 & {\color{vmbad}0.738} & 0.829 & {\bfseries\color{vmbad}0.592} & \na & \na \\
\quad Cov. $\uparrow$ & 0.871 & {\color{vmgood}0.894} & 0.754 & {\color{vmgood}0.765} & 0.737 & {\bfseries\color{vmbad}0.685} & \na & \na \\
\addlinespace
CLIP-T $\uparrow$ & \na & \na & 30.35 & 30.25 & 30.31 & {\color{vmgood}30.66} & \na & \na \\
\addlinespace
\multicolumn{9}{@{}l}{\textit{Frame accuracy}} \\
\quad PSNR $\uparrow$ & \na & \na & \na & \na & \na & \na & 25.50 & 25.49 \\
\quad SSIM $\uparrow$ & \na & \na & \na & \na & \na & \na & 0.8536 & 0.8539 \\
\bottomrule
\end{tabular}
\end{adjustbox}
\end{table}

\begin{table}[H]
\centering
\caption{Text-to-motion, every metric: MDM on KIT-ML with its corrected evaluator, mean over 20 replications of 1,000 test motions. Real scores the test set against itself; diversity is best closest to Real and is not colored.}
\label{tab:full-motion}
\footnotesize
\setlength{\tabcolsep}{2pt}
\begin{adjustbox}{max width=\linewidth}
\begin{tabular}{@{}lrrr@{}}
\toprule
 & \hd{Real} & \hd{Pub.\ (2.5)} & \hd{Ours} \\
\midrule
\multicolumn{4}{@{}l}{\textit{R-precision}} \\
\quad Top-1 $\uparrow$ & 0.429 & 0.400 & 0.399 \\
\quad Top-2 $\uparrow$ & 0.655 & 0.604 & 0.598 \\
\quad Top-3 $\uparrow$ & 0.783 & 0.721 & 0.720 \\
\addlinespace
FID $\downarrow$ & 0.028 & 0.536 & {\color{vmgood}0.523} \\
MM-Dist $\downarrow$ & 2.779 & 3.115 & {\color{vmbad}3.158} \\
Diversity $\rightarrow$ & 11.09 & 10.80 & 10.74 \\
MMod. $\uparrow$ & \na & 1.54 & {\bfseries\color{vmgood}1.72} \\
\bottomrule
\end{tabular}
\end{adjustbox}
\end{table}

\begin{table}[H]
\centering
\caption{Text-to-audio, every metric. AudioLDM 2 on the 4,411 AudioCaps test captions, scored with its own suite and with independent metrics (real clips: CLAP 0.373, LAION 0.355, IS 9.32). Stable Audio Open on 586 Song Describer captions with its own suite; Pub.\ reproduces its reported KL and CLAP (0.55, 0.41) but not its FD$_{\mathrm{OpenL3}}$ (96.5).}
\label{tab:full-audio}
\footnotesize
\setlength{\tabcolsep}{2pt}
\begin{adjustbox}{max width=\linewidth}
\begin{tabular}{@{}lrr@{}}
\toprule
 & \hd{Pub.} & \hd{Ours} \\
\midrule
\multicolumn{3}{@{}l}{\textbf{AudioLDM 2} \data{$w=3.5$}} \\
\multicolumn{3}{@{}l}{\textit{Own suite}} \\
\quad FAD (VGGish) $\downarrow$ & 2.00 & {\bfseries\color{vmbad}2.46} \\
\quad KL (softmax) $\downarrow$ & 1.47 & {\bfseries\color{vmgood}1.29} \\
\quad IS $\uparrow$ & 7.93 & {\bfseries\color{vmgood}9.43} \\
\quad CLAP $\uparrow$ & 0.241 & {\bfseries\color{vmgood}0.283} \\
\addlinespace
\multicolumn{3}{@{}l}{\textit{Independent}} \\
\quad KL (sigmoid) $\downarrow$ & 3.28 & {\bfseries\color{vmgood}2.74} \\
\quad CLAP (LAION) $\uparrow$ & 0.241 & {\bfseries\color{vmgood}0.279} \\
\quad FD (PaSST) $\downarrow$ & 185.8 & {\bfseries\color{vmbad}200.6} \\
\quad P $\uparrow$ & 0.825 & {\bfseries\color{vmbad}0.721} \\
\quad R $\uparrow$ & 0.201 & {\bfseries\color{vmbad}0.167} \\
\quad Dens. $\uparrow$ & 0.936 & {\bfseries\color{vmbad}0.807} \\
\quad Cov. $\uparrow$ & 0.868 & {\bfseries\color{vmbad}0.814} \\
\midrule
\multicolumn{3}{@{}l}{\textbf{Stable Audio Open} \data{$w=7$}} \\
\quad FD (OpenL3) $\downarrow$ & 165.18 & 164.95 \\
\quad KL $\downarrow$ & 0.549 & {\bfseries\color{vmgood}0.513} \\
\quad CLAP $\uparrow$ & 0.421 & 0.418 \\
\bottomrule
\end{tabular}
\end{adjustbox}
\end{table}

\subsection{User study}
\label{app:user-study}\label{app:user-study-protocol}

The study of \cref{sec:exp-user} compares seven settings on SD~3.5 Medium: the FID- and FD$_\mathrm{DINOv2}$-tuned constants ($w=1.6$, $2.4$), our constant ($w=4.14$), our schedule, and the interval, linear, and TV-CFG schedules at the released $w=4.5$. Each setting renders held-out COCO captions 5,705--6,204 (40 Euler steps, $480^2$), with one initial noise per caption shared by all seven. We omit the published constant ($w=4.5$), since our predicted scale is already close to it.

A trial shows a caption and two of its images, from two settings, on random sides; the rater picks the better image or calls a tie. Each rater's trials go through the $21$ pairs of settings evenly and favor the captions they have seen least. Raters are identified by the name they enter, so one person on several devices counts once. The study was not reviewed by an institutional review board: it is an informal preference poll among volunteers, run through a link that the authors shared with their research group, friends, and an online community of AI developers, with no compensation, no personal or sensitive data beyond a self-chosen name used to count each person once, and no risk beyond viewing generated images, which were screened for unsuitable content. No formal consent was collected; following the link and voting was voluntary and could be stopped at any time. The tables count the $1{,}458$ votes of $41$ people; the authors' votes and those of AI raters are left out. We fit a Bradley--Terry model in which a tie counts as half a win for each side, with one prior tie per pair, and report it on the Elo scale (mean $1000$; a $100$-point gap means the higher setting is preferred about $64\%$ of the time). The $\pm$ in \cref{tab:user-study} is $1.96$ standard deviations over $1{,}000$ bootstrap resamples of raters, then of votes within each rater. \Cref{tab:user-study-h2h} gives every pairwise result.

\begin{table}[H]
\centering
\caption{Pairwise results of the informal user study (\cref{tab:user-study}): how often the row setting beat the column setting, ties counting half, with the number of games in parentheses. Shading: \swatch{vmbetter}{green} where the row wins more often, \swatch{vmworse}{red} where it loses more often, deeper for a more one-sided result. Constants come first, then schedules, as in \cref{tab:user-study}.}
\label{tab:user-study-h2h}
\footnotesize
\setlength{\tabcolsep}{2pt}
\begin{adjustbox}{max width=\linewidth}
\begin{tabular}{lccc@{\hspace{8pt}}cccc}
\toprule
Row vs.\ column & Ours 4.14 & 2.4 & 1.6 & Ours $w(\sigma)$ & TV & Int. & Lin. \\
\midrule
Ours 4.14 & -- & \cellcolor{vmbetter!45}68\,{\scriptsize(67)} & \cellcolor{vmbetter!62}75\,{\scriptsize(69)} & 50\,{\scriptsize(75)} & \cellcolor{vmworse!11}46\,{\scriptsize(71)} & \cellcolor{vmworse!2}49\,{\scriptsize(71)} & \cellcolor{vmbetter!24}59\,{\scriptsize(69)} \\
2.4 & \cellcolor{vmworse!45}32\,{\scriptsize(67)} & -- & \cellcolor{vmbetter!32}63\,{\scriptsize(67)} & \cellcolor{vmworse!50}30\,{\scriptsize(65)} & \cellcolor{vmworse!30}38\,{\scriptsize(70)} & \cellcolor{vmbetter!30}62\,{\scriptsize(66)} & \cellcolor{vmbetter!25}60\,{\scriptsize(69)} \\
1.6 & \cellcolor{vmworse!62}25\,{\scriptsize(69)} & \cellcolor{vmworse!32}37\,{\scriptsize(67)} & -- & \cellcolor{vmworse!67}23\,{\scriptsize(71)} & \cellcolor{vmworse!62}25\,{\scriptsize(68)} & \cellcolor{vmworse!29}38\,{\scriptsize(69)} & \cellcolor{vmworse!47}31\,{\scriptsize(75)} \\
\midrule
Ours $w(\sigma)$ & 50\,{\scriptsize(75)} & \cellcolor{vmbetter!50}70\,{\scriptsize(65)} & \cellcolor{vmbetter!67}77\,{\scriptsize(71)} & -- & \cellcolor{vmbetter!21}58\,{\scriptsize(67)} & \cellcolor{vmbetter!29}62\,{\scriptsize(69)} & \cellcolor{vmbetter!47}69\,{\scriptsize(72)} \\
TV & \cellcolor{vmbetter!11}54\,{\scriptsize(71)} & \cellcolor{vmbetter!30}62\,{\scriptsize(70)} & \cellcolor{vmbetter!62}75\,{\scriptsize(68)} & \cellcolor{vmworse!21}42\,{\scriptsize(67)} & -- & \cellcolor{vmbetter!2}51\,{\scriptsize(69)} & \cellcolor{vmbetter!46}68\,{\scriptsize(73)} \\
Int. & \cellcolor{vmbetter!2}51\,{\scriptsize(71)} & \cellcolor{vmworse!30}38\,{\scriptsize(66)} & \cellcolor{vmbetter!29}62\,{\scriptsize(69)} & \cellcolor{vmworse!29}38\,{\scriptsize(69)} & \cellcolor{vmworse!2}49\,{\scriptsize(69)} & -- & \cellcolor{vmbetter!8}53\,{\scriptsize(66)} \\
Lin. & \cellcolor{vmworse!24}41\,{\scriptsize(69)} & \cellcolor{vmworse!25}40\,{\scriptsize(69)} & \cellcolor{vmbetter!47}69\,{\scriptsize(75)} & \cellcolor{vmworse!47}31\,{\scriptsize(72)} & \cellcolor{vmworse!46}32\,{\scriptsize(73)} & \cellcolor{vmworse!8}47\,{\scriptsize(66)} & -- \\
\bottomrule
\end{tabular}

\end{adjustbox}
\end{table}

\paragraph{LLM as a judge.}
\label{app:ai-rater}
We also ran Claude Fable 5.1 (Anthropic) as a rater, through an API that gives it the same trials as people: a caption and two of its images on random sides, with the same instruction to pick the image it prefers overall, for how good it looks, how natural it is, and how well it fits the caption, or to call a tie. It cast $2{,}500$ votes, kept apart from the people's. \Cref{tab:user-study-fable} gives its ratings, with the $\pm$ over $1{,}000$ bootstrap resamples of its votes. Its ranking agrees with the people's: the same three settings lead (our constant, our schedule, and TV-CFG, within each other's intervals), and the other four follow in the same order. It separates the settings more sharply ($445$ Elo points from best to worst, against $214$ for people) and almost never calls a tie ($18$ of its $2{,}500$ votes, against $15\%$ of the people's).

\begin{table}[H]
\centering
\caption{Claude Fable 5.1 as a rater in the study of \cref{tab:user-study}: Bradley--Terry ratings on the Elo scale from its $2{,}500$ votes alone, $\pm$ a $95\%$ bootstrap interval over votes. Same settings and layout as \cref{tab:user-study}.}
\label{tab:user-study-fable}
\footnotesize
\setlength{\tabcolsep}{3pt}
\begin{adjustbox}{max width=\linewidth}
\begin{tabular}[t]{@{}llr@{\,}c@{\,}l@{}}
\toprule
\multicolumn{5}{@{}c@{}}{\emph{Constant weight}} \\
\cmidrule{1-5}
Setting & $w$ & \multicolumn{3}{c@{}}{Elo} \\
\midrule
\textbf{Ours} & {\boldmath4.14} & \textbf{1148} & {\boldmath$\pm$} & \textbf{25} \\
\multicolumn{5}{@{}l}{\emph{Grid search over:}} \\
\quad FD$_\mathrm{DINOv2}$ & 2.4 & 982 & $\pm$ & 25 \\
\quad FID & 1.6 & 703 & $\pm$ & 35 \\
\bottomrule
\end{tabular}\hspace{14pt}%
\begin{tabular}[t]{@{}llr@{\,}c@{\,}l@{}}
\toprule
\multicolumn{5}{@{}c@{}}{\emph{Schedule}} \\
\cmidrule{1-5}
Setting & $w$ & \multicolumn{3}{c@{}}{Elo} \\
\midrule
\textbf{Ours} & {\boldmath$w(\sigma)$} & \textbf{1135} & {\boldmath$\pm$} & \textbf{25} \\
TV-CFG & 4.5 & 1130 & $\pm$ & 30 \\
Interval & 4.5 & 990 & $\pm$ & 24 \\
Linear & 4.5 & 912 & $\pm$ & 24 \\
\bottomrule
\end{tabular}

\end{adjustbox}
\end{table}

\onecolumn
\section{Additional samples}
\label{app:samples}

This section provides qualitative comparisons for our image models: $23$ SD~3.5 Medium generations comparing all seven user-study settings and the released scale (\cref{fig:samples-sd35m}), side-by-side comparisons of published scales against our schedule for SD~1.5, SD~3.5 Large, and FLUX.2~[klein]~4B (\cref{fig:samples-pub}), and $100$ ImageNet-512 pairs each for EDM2-XXL under CFG and autoguidance (\cref{fig:samples-edm2-cfg,fig:samples-edm2-ag}). Except for \cref{fig:samples-pub}, whose pairs were picked by hand, all grids are presented in fixed order without cherry-picking; every row or pair shares its prompt or class and initial noise.

\vspace{4pt}
\noindent\begin{adjustbox}{max totalheight=\dimexpr\textheight-95pt\relax,center}%
\begin{minipage}{\textwidth}\centering
{\scriptsize\centering%
\hbox to \textwidth{\makebox[0.123870\textwidth]{\shortstack{Ours\\$w=4.14$}}\hfill\makebox[0.123870\textwidth]{\shortstack{Ours\\$w(\sigma)$}}\hfill\makebox[0.123870\textwidth]{\shortstack{FID\\$w=1.6$}}\hfill\makebox[0.123870\textwidth]{\shortstack{FD$_{\mathrm{DINOv2}}$\\$w=2.4$}}\hfill\makebox[0.123870\textwidth]{\shortstack{Released\\$w=4.5$}}\hfill\makebox[0.123870\textwidth]{\shortstack{Interval\\$w=4.5$}}\hfill\makebox[0.123870\textwidth]{\shortstack{Linear\\$w=4.5$}}\hfill\makebox[0.123870\textwidth]{\shortstack{TV-CFG\\$w=4.5$}}}\par\vspace{3pt}
\includegraphics[width=\textwidth,page=1]{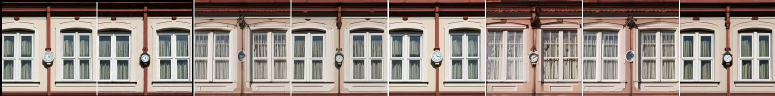}\par\vspace{1pt}
{\itshape A clock that is in between two windows on a building.}\par\vspace{5pt}
\includegraphics[width=\textwidth,page=2]{Figures/samples-sd35m-rows.pdf}\par\vspace{1pt}
{\itshape a person standing at a counter getting some bread ready for grilled cheese sandwiches}\par\vspace{5pt}
\includegraphics[width=\textwidth,page=3]{Figures/samples-sd35m-rows.pdf}\par\vspace{1pt}
{\itshape A boy sits on a rock in a hat as he holds a trumpet near an old barn and some farm animals.}\par\vspace{5pt}
\includegraphics[width=\textwidth,page=4]{Figures/samples-sd35m-rows.pdf}\par\vspace{1pt}
{\itshape A large brown bear walking across a lush green field.}\par\vspace{5pt}
\includegraphics[width=\textwidth,page=5]{Figures/samples-sd35m-rows.pdf}\par\vspace{1pt}
{\itshape A delicious looking donut or cinnamon roll covered in icing}\par\vspace{5pt}
\includegraphics[width=\textwidth,page=6]{Figures/samples-sd35m-rows.pdf}\par\vspace{1pt}
{\itshape Two zebras that are standing near each other in the dirt.}\par\vspace{5pt}
\includegraphics[width=\textwidth,page=7]{Figures/samples-sd35m-rows.pdf}\par\vspace{1pt}
{\itshape a mug of beef sitting next to a plate with a slice of deep dish pizza on it}\par\vspace{5pt}
}

\captionsetup{hypcap=false}%
\captionof{figure}{More samples from SD~3.5 Medium on held-out COCO captions (40 Euler steps, $480^2$), with the seven settings of the user study (\cref{tab:user-study}): our constant ($w=4.14$) and our schedule $w(\sigma)$ (framed in black), the FID- and FD$_{\mathrm{DINOv2}}$-tuned constants ($w=1.6$, $2.4$), and the interval, linear and TV-CFG schedules at the released $w=4.5$; we also show the released constant $w=4.5$ itself. These are the images shown to raters. Each row shares its caption and initial noise. The rows are the first 23 captions shown to raters, in COCO order and not selected; this page and the next two.}
\label{fig:samples-sd35m}
\end{minipage}\end{adjustbox}\par

{\renewcommand{\twocolumn}[1][]{\clearpage\noindent\begin{adjustbox}{max totalheight=\textheight,center}\begin{minipage}{\textwidth}#1\end{minipage}\end{adjustbox}\par}%
\twocolumn[{%
\centering
{\scriptsize\centering%
\hbox to \textwidth{\makebox[0.123870\textwidth]{\shortstack{Ours\\$w=4.14$}}\hfill\makebox[0.123870\textwidth]{\shortstack{Ours\\$w(\sigma)$}}\hfill\makebox[0.123870\textwidth]{\shortstack{FID\\$w=1.6$}}\hfill\makebox[0.123870\textwidth]{\shortstack{FD$_{\mathrm{DINOv2}}$\\$w=2.4$}}\hfill\makebox[0.123870\textwidth]{\shortstack{Released\\$w=4.5$}}\hfill\makebox[0.123870\textwidth]{\shortstack{Interval\\$w=4.5$}}\hfill\makebox[0.123870\textwidth]{\shortstack{Linear\\$w=4.5$}}\hfill\makebox[0.123870\textwidth]{\shortstack{TV-CFG\\$w=4.5$}}}\par\vspace{3pt}
\includegraphics[width=\textwidth,page=8]{Figures/samples-sd35m-rows.pdf}\par\vspace{1pt}
{\itshape a sign on a wall with a clock in the background}\par\vspace{5pt}
\includegraphics[width=\textwidth,page=9]{Figures/samples-sd35m-rows.pdf}\par\vspace{1pt}
{\itshape A man in a blue shirt catches a Frisbee thrown to him.}\par\vspace{5pt}
\includegraphics[width=\textwidth,page=10]{Figures/samples-sd35m-rows.pdf}\par\vspace{1pt}
{\itshape A woman holds a basket with a hot dog while she is surrounded by two other people.}\par\vspace{5pt}
\includegraphics[width=\textwidth,page=11]{Figures/samples-sd35m-rows.pdf}\par\vspace{1pt}
{\itshape A cat laying next to a laptop and a computer mouse.}\par\vspace{5pt}
\includegraphics[width=\textwidth,page=12]{Figures/samples-sd35m-rows.pdf}\par\vspace{1pt}
{\itshape there is a black bear that has caught a fish in his mouth}\par\vspace{5pt}
\includegraphics[width=\textwidth,page=13]{Figures/samples-sd35m-rows.pdf}\par\vspace{1pt}
{\itshape A chocolate doughnut with sprinkles in a basket next to a cup of coffee on a saucer.}\par\vspace{5pt}
\includegraphics[width=\textwidth,page=14]{Figures/samples-sd35m-rows.pdf}\par\vspace{1pt}
{\itshape Many packages of bread have been placed on the small kitchen counter.}\par\vspace{5pt}
\includegraphics[width=\textwidth,page=15]{Figures/samples-sd35m-rows.pdf}\par\vspace{1pt}
{\itshape Two loafs of bread sitting on a counter behind a window.}\par\vspace{5pt}
}

\captionsetup{hypcap=false}%
\captionof{figure}{More samples from SD~3.5 Medium, continued from \cref{fig:samples-sd35m} (captions 8--15 of 23); same settings and layout.}
}]

\twocolumn[{%
\centering
{\scriptsize\centering%
\hbox to \textwidth{\makebox[0.123870\textwidth]{\shortstack{Ours\\$w=4.14$}}\hfill\makebox[0.123870\textwidth]{\shortstack{Ours\\$w(\sigma)$}}\hfill\makebox[0.123870\textwidth]{\shortstack{FID\\$w=1.6$}}\hfill\makebox[0.123870\textwidth]{\shortstack{FD$_{\mathrm{DINOv2}}$\\$w=2.4$}}\hfill\makebox[0.123870\textwidth]{\shortstack{Released\\$w=4.5$}}\hfill\makebox[0.123870\textwidth]{\shortstack{Interval\\$w=4.5$}}\hfill\makebox[0.123870\textwidth]{\shortstack{Linear\\$w=4.5$}}\hfill\makebox[0.123870\textwidth]{\shortstack{TV-CFG\\$w=4.5$}}}\par\vspace{3pt}
\includegraphics[width=\textwidth,page=16]{Figures/samples-sd35m-rows.pdf}\par\vspace{1pt}
{\itshape An old fashioned stove top oven with pots sitting on top of it.}\par\vspace{5pt}
\includegraphics[width=\textwidth,page=17]{Figures/samples-sd35m-rows.pdf}\par\vspace{1pt}
{\itshape A dog stands on the bow of a white sailboat in the water.}\par\vspace{5pt}
\includegraphics[width=\textwidth,page=18]{Figures/samples-sd35m-rows.pdf}\par\vspace{1pt}
{\itshape A man juggling while standing on top of a sign riding a skateboard.}\par\vspace{5pt}
\includegraphics[width=\textwidth,page=19]{Figures/samples-sd35m-rows.pdf}\par\vspace{1pt}
{\itshape A plate topped with rolls next to a bowl of food.}\par\vspace{5pt}
\includegraphics[width=\textwidth,page=20]{Figures/samples-sd35m-rows.pdf}\par\vspace{1pt}
{\itshape Two people, a man and a woman, are toasting with wine glasses.}\par\vspace{5pt}
\includegraphics[width=\textwidth,page=21]{Figures/samples-sd35m-rows.pdf}\par\vspace{1pt}
{\itshape A man in his ski gear is smiling for the camera.}\par\vspace{5pt}
\includegraphics[width=\textwidth,page=22]{Figures/samples-sd35m-rows.pdf}\par\vspace{1pt}
{\itshape A kitchen area with counter, sink , stove and windows.}\par\vspace{5pt}
\includegraphics[width=\textwidth,page=23]{Figures/samples-sd35m-rows.pdf}\par\vspace{1pt}
{\itshape Three dogs bite at a frisbee while at a park.}\par\vspace{5pt}
}

\captionsetup{hypcap=false}%
\captionof{figure}{More samples from SD~3.5 Medium, continued from \cref{fig:samples-sd35m} (captions 16--23 of 23); same settings and layout.}
}]
}

\twocolumn[{%
\centering
{\setlength{\tabcolsep}{0.6pt}\renewcommand{\arraystretch}{0.5}\scriptsize%
\ifdefined\pubcell\else\newlength{\pubcell}\newlength{\pairgap}\fi
\setlength{\pairgap}{0.1in}%
\setlength{\pubcell}{\dimexpr(\textwidth-\pairgap*3-1.2pt*8)/8\relax}%
\begin{tabular}{@{}cc@{\hspace{\pairgap}}cc@{\hspace{\pairgap}}cc@{\hspace{\pairgap}}cc@{}}
\multicolumn{8}{@{}l}{\small\bfseries SD~1.5}\\[2pt]
Pub.\ ($w=7.5$) & Ours ($w(\sigma)$) & Pub.\ ($w=7.5$) & Ours ($w(\sigma)$) & Pub.\ ($w=7.5$) & Ours ($w(\sigma)$) & Pub.\ ($w=7.5$) & Ours ($w(\sigma)$)\\[2pt]
\includegraphics[width=\pubcell]{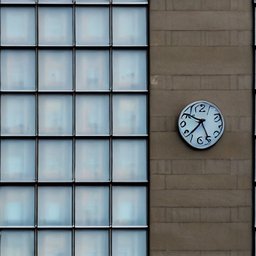} & \includegraphics[width=\pubcell]{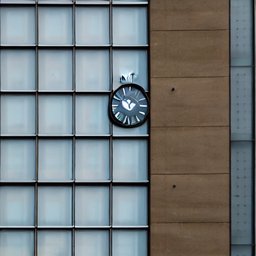} & \includegraphics[width=\pubcell]{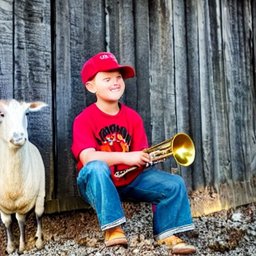} & \includegraphics[width=\pubcell]{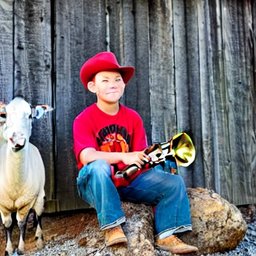} & \includegraphics[width=\pubcell]{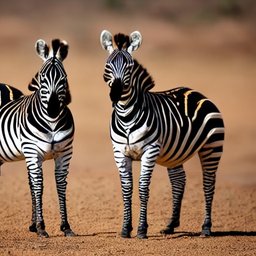} & \includegraphics[width=\pubcell]{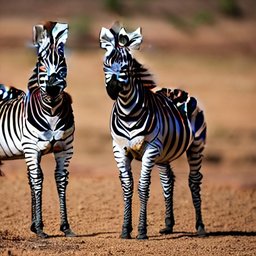} & \includegraphics[width=\pubcell]{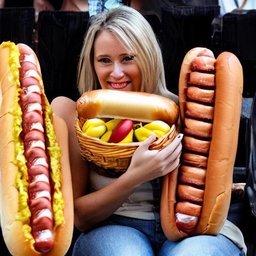} & \includegraphics[width=\pubcell]{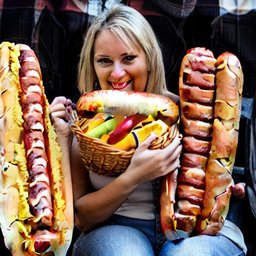}\\[1pt]
\multicolumn{2}{p{\dimexpr2\pubcell+1.2pt\relax}}{\centering\itshape A clock that is in between two windows on a building.\rule[-3pt]{0pt}{0pt}} & \multicolumn{2}{p{\dimexpr2\pubcell+1.2pt\relax}}{\centering\itshape A boy sits on a rock in a hat as he holds a trumpet near an old barn and some farm…\rule[-3pt]{0pt}{0pt}} & \multicolumn{2}{p{\dimexpr2\pubcell+1.2pt\relax}}{\centering\itshape Two zebras that are standing near each other in the dirt.\rule[-3pt]{0pt}{0pt}} & \multicolumn{2}{p{\dimexpr2\pubcell+1.2pt\relax}}{\centering\itshape A woman holds a basket with a hot dog while she is surrounded by two other people.\rule[-3pt]{0pt}{0pt}}\\[4pt]
\includegraphics[width=\pubcell]{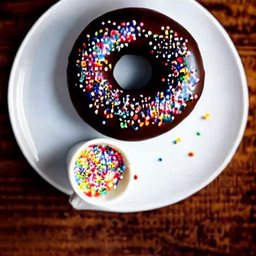} & \includegraphics[width=\pubcell]{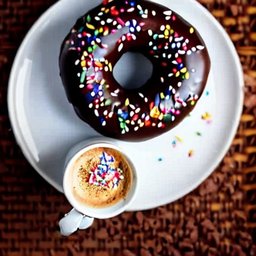} & \includegraphics[width=\pubcell]{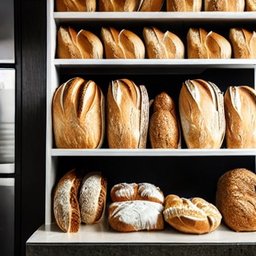} & \includegraphics[width=\pubcell]{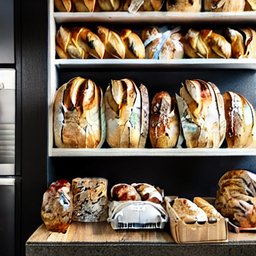} & \includegraphics[width=\pubcell]{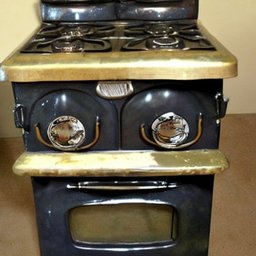} & \includegraphics[width=\pubcell]{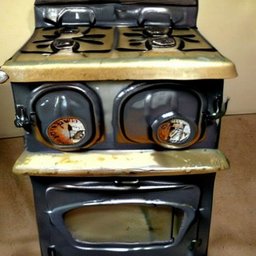} & \includegraphics[width=\pubcell]{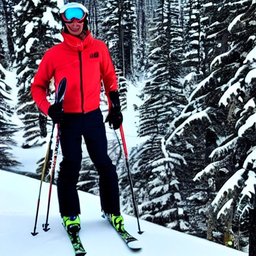} & \includegraphics[width=\pubcell]{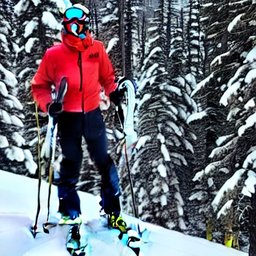}\\[1pt]
\multicolumn{2}{p{\dimexpr2\pubcell+1.2pt\relax}}{\centering\itshape A chocolate doughnut with sprinkles in a basket next to a cup of coffee on a saucer.\rule[-3pt]{0pt}{0pt}} & \multicolumn{2}{p{\dimexpr2\pubcell+1.2pt\relax}}{\centering\itshape Many packages of bread have been placed on the small kitchen counter.\rule[-3pt]{0pt}{0pt}} & \multicolumn{2}{p{\dimexpr2\pubcell+1.2pt\relax}}{\centering\itshape An old fashioned stove top oven with pots sitting on top of it.\rule[-3pt]{0pt}{0pt}} & \multicolumn{2}{p{\dimexpr2\pubcell+1.2pt\relax}}{\centering\itshape A man in his ski gear is smiling for the camera.\rule[-3pt]{0pt}{0pt}}\\[4pt]
\noalign{\vskip 4pt}\multicolumn{8}{@{}l}{\small\bfseries SD~3.5 Large}\\[2pt]
Pub.\ ($w=3.5$) & Ours ($w(\sigma)$) & Pub.\ ($w=3.5$) & Ours ($w(\sigma)$) & Pub.\ ($w=3.5$) & Ours ($w(\sigma)$) & Pub.\ ($w=3.5$) & Ours ($w(\sigma)$)\\[2pt]
\includegraphics[width=\pubcell]{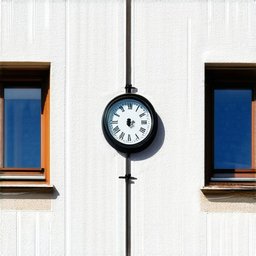} & \includegraphics[width=\pubcell]{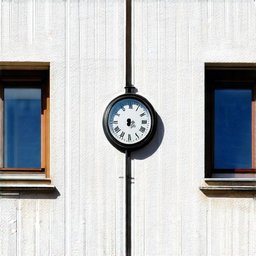} & \includegraphics[width=\pubcell]{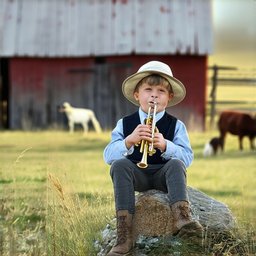} & \includegraphics[width=\pubcell]{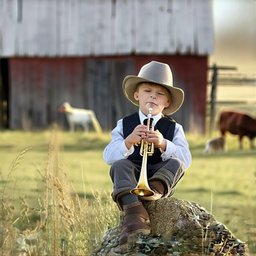} & \includegraphics[width=\pubcell]{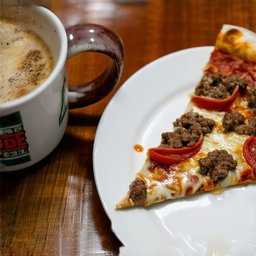} & \includegraphics[width=\pubcell]{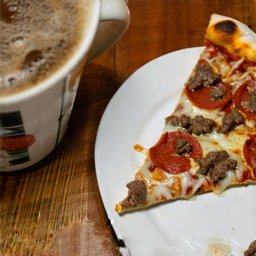} & \includegraphics[width=\pubcell]{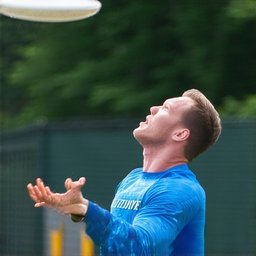} & \includegraphics[width=\pubcell]{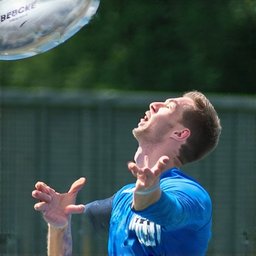}\\[1pt]
\multicolumn{2}{p{\dimexpr2\pubcell+1.2pt\relax}}{\centering\itshape A clock that is in between two windows on a building.\rule[-3pt]{0pt}{0pt}} & \multicolumn{2}{p{\dimexpr2\pubcell+1.2pt\relax}}{\centering\itshape A boy sits on a rock in a hat as he holds a trumpet near an old barn and some farm…\rule[-3pt]{0pt}{0pt}} & \multicolumn{2}{p{\dimexpr2\pubcell+1.2pt\relax}}{\centering\itshape a mug of beef sitting next to a plate with a slice of deep dish pizza on it\rule[-3pt]{0pt}{0pt}} & \multicolumn{2}{p{\dimexpr2\pubcell+1.2pt\relax}}{\centering\itshape A man in a blue shirt catches a Frisbee thrown to him.\rule[-3pt]{0pt}{0pt}}\\[4pt]
\includegraphics[width=\pubcell]{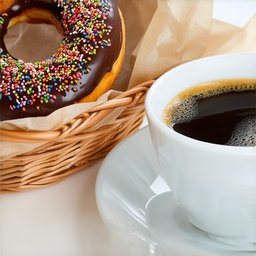} & \includegraphics[width=\pubcell]{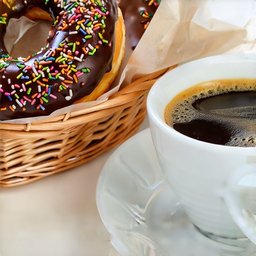} & \includegraphics[width=\pubcell]{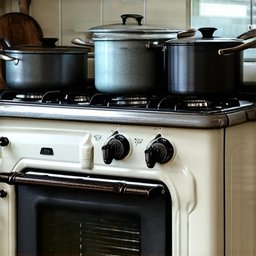} & \includegraphics[width=\pubcell]{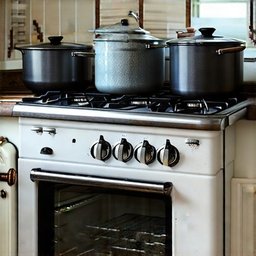} & \includegraphics[width=\pubcell]{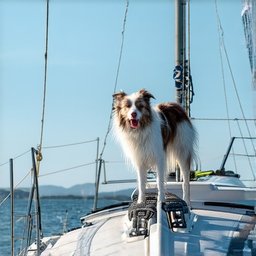} & \includegraphics[width=\pubcell]{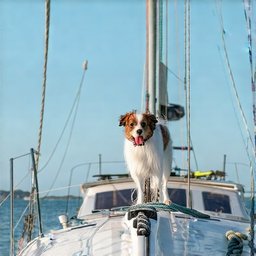} & \includegraphics[width=\pubcell]{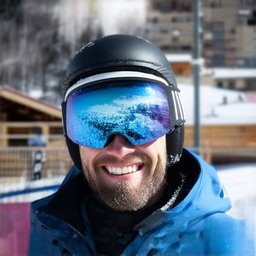} & \includegraphics[width=\pubcell]{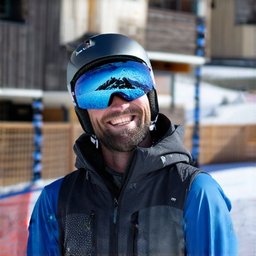}\\[1pt]
\multicolumn{2}{p{\dimexpr2\pubcell+1.2pt\relax}}{\centering\itshape A chocolate doughnut with sprinkles in a basket next to a cup of coffee on a saucer.\rule[-3pt]{0pt}{0pt}} & \multicolumn{2}{p{\dimexpr2\pubcell+1.2pt\relax}}{\centering\itshape An old fashioned stove top oven with pots sitting on top of it.\rule[-3pt]{0pt}{0pt}} & \multicolumn{2}{p{\dimexpr2\pubcell+1.2pt\relax}}{\centering\itshape A dog stands on the bow of a white sailboat in the water.\rule[-3pt]{0pt}{0pt}} & \multicolumn{2}{p{\dimexpr2\pubcell+1.2pt\relax}}{\centering\itshape A man in his ski gear is smiling for the camera.\rule[-3pt]{0pt}{0pt}}\\[4pt]
\noalign{\vskip 4pt}\multicolumn{8}{@{}l}{\small\bfseries FLUX.2 [klein] 4B}\\[2pt]
Pub.\ ($w=4$) & Ours ($w(\sigma)$) & Pub.\ ($w=4$) & Ours ($w(\sigma)$) & Pub.\ ($w=4$) & Ours ($w(\sigma)$) & Pub.\ ($w=4$) & Ours ($w(\sigma)$)\\[2pt]
\includegraphics[width=\pubcell]{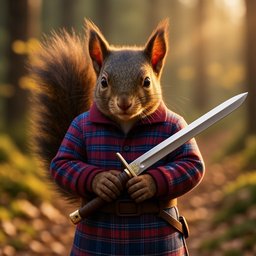} & \includegraphics[width=\pubcell]{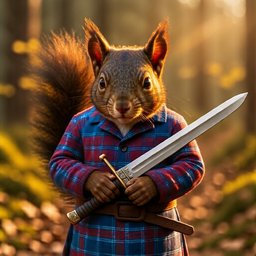} & \includegraphics[width=\pubcell]{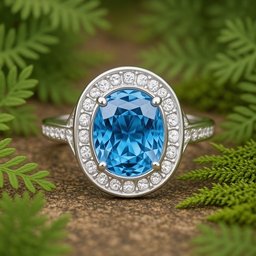} & \includegraphics[width=\pubcell]{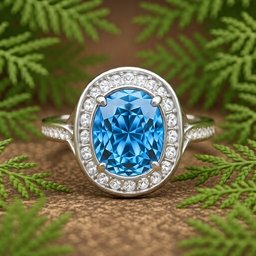} & \includegraphics[width=\pubcell]{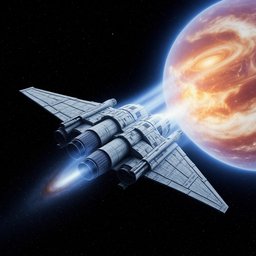} & \includegraphics[width=\pubcell]{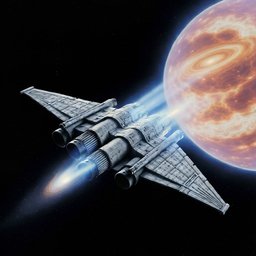} & \includegraphics[width=\pubcell]{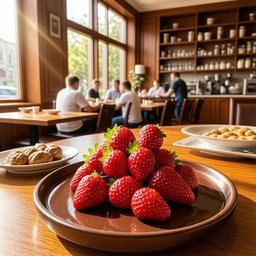} & \includegraphics[width=\pubcell]{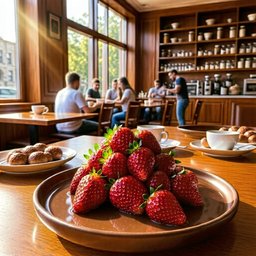}\\[1pt]
\multicolumn{2}{p{\dimexpr2\pubcell+1.2pt\relax}}{\centering\itshape SCOTTISH WARRIOR SQUIRREL WEARING A KILT WITH A sheathed sword High contrast Sharp…\rule[-3pt]{0pt}{0pt}} & \multicolumn{2}{p{\dimexpr2\pubcell+1.2pt\relax}}{\centering\itshape White Gold ring with sapphire diamond nature theme HD realistic\rule[-3pt]{0pt}{0pt}} & \multicolumn{2}{p{\dimexpr2\pubcell+1.2pt\relax}}{\centering\itshape flight back to earth, spaceship, solar system\rule[-3pt]{0pt}{0pt}} & \multicolumn{2}{p{\dimexpr2\pubcell+1.2pt\relax}}{\centering\itshape strawberries in chocolate restaurant feed appetizing sunny background realist…\rule[-3pt]{0pt}{0pt}}\\[4pt]
\includegraphics[width=\pubcell]{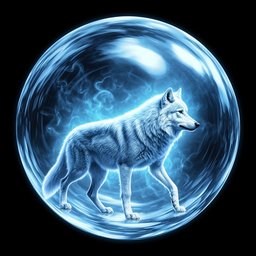} & \includegraphics[width=\pubcell]{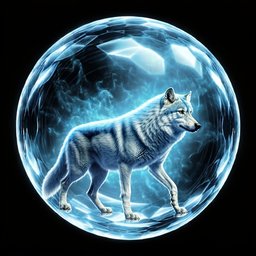} & \includegraphics[width=\pubcell]{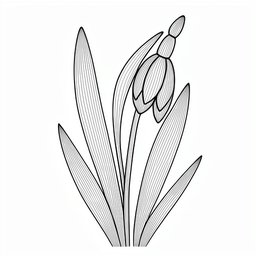} & \includegraphics[width=\pubcell]{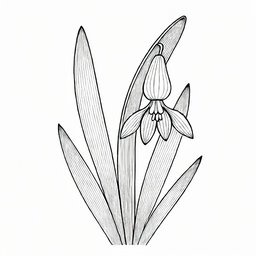} & \includegraphics[width=\pubcell]{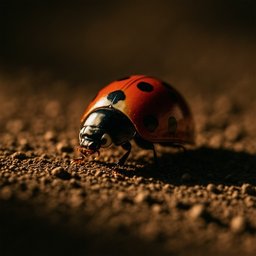} & \includegraphics[width=\pubcell]{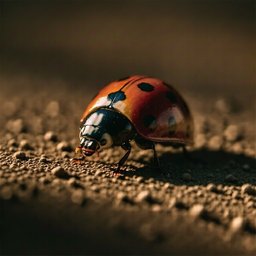} & \includegraphics[width=\pubcell]{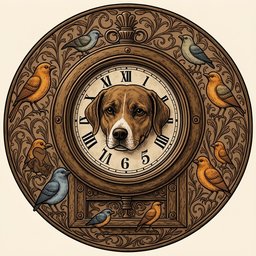} & \includegraphics[width=\pubcell]{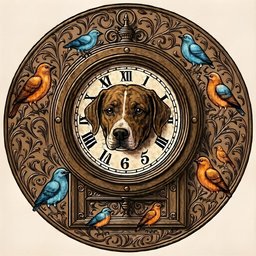}\\[1pt]
\multicolumn{2}{p{\dimexpr2\pubcell+1.2pt\relax}}{\centering\itshape Futuristic wolf inside a huge electric glass ball, the wolf fits into the glas…\rule[-3pt]{0pt}{0pt}} & \multicolumn{2}{p{\dimexpr2\pubcell+1.2pt\relax}}{\centering\itshape colouring page for adults, fine lines, no background, white background, snowdrop…\rule[-3pt]{0pt}{0pt}} & \multicolumn{2}{p{\dimexpr2\pubcell+1.2pt\relax}}{\centering\itshape lady bug, photography, cinematic light\rule[-3pt]{0pt}{0pt}} & \multicolumn{2}{p{\dimexpr2\pubcell+1.2pt\relax}}{\centering\itshape a dog has its head under an old clock, in the style of ildiko neer, dark beige and…\rule[-3pt]{0pt}{0pt}}\\[4pt]
\end{tabular}}

\captionsetup{hypcap=false}%
\captionof{figure}{Samples from the other text-to-image models of \cref{tab:pub-all} at the released scale (Pub.) and with our schedule $w(\sigma)$ (Ours): SD~1.5 and SD~3.5 Large on held-out COCO captions, FLUX.2 [klein] 4B on MJHQ-30K prompts (long prompts are cut short). Each pair shares its caption and initial noise; the pairs were picked by hand from each model's first 24 evaluation captions.}
\label{fig:samples-pub}
}]

\onecolumn
\noindent\begin{adjustbox}{max totalheight=\dimexpr\textheight-26pt\relax,center}%
\begin{minipage}{\textwidth}\centering
{\scriptsize\centering%
\hbox to \textwidth{\makebox[0.121518\textwidth]{\shortstack{Pub.\\$w=1.20$}}\hspace{0.001265\textwidth}\makebox[0.121518\textwidth]{\shortstack{Ours\\$w(\sigma)$}}\hspace{0.007594\textwidth}\makebox[0.121518\textwidth]{\shortstack{Pub.\\$w=1.20$}}\hspace{0.001265\textwidth}\makebox[0.121518\textwidth]{\shortstack{Ours\\$w(\sigma)$}}\hspace{0.007594\textwidth}\makebox[0.121518\textwidth]{\shortstack{Pub.\\$w=1.20$}}\hspace{0.001265\textwidth}\makebox[0.121518\textwidth]{\shortstack{Ours\\$w(\sigma)$}}\hspace{0.007594\textwidth}\makebox[0.121518\textwidth]{\shortstack{Pub.\\$w=1.20$}}\hspace{0.001265\textwidth}\makebox[0.121518\textwidth]{\shortstack{Ours\\$w(\sigma)$}}\hss}\par\vspace{3pt}
\includegraphics[width=\textwidth,page=1]{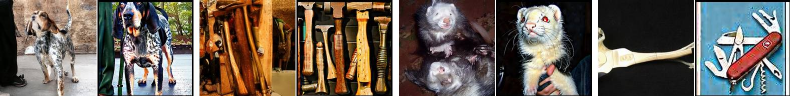}\par\vspace{1pt}
\hbox to \textwidth{\makebox[0.244303\textwidth]{\itshape bluetick}\hspace{0.007594\textwidth}\makebox[0.244303\textwidth]{\itshape hammer}\hspace{0.007594\textwidth}\makebox[0.244303\textwidth]{\itshape black-footed ferret}\hspace{0.007594\textwidth}\makebox[0.244303\textwidth]{\itshape can opener}\hss}\par\vspace{5pt}
\includegraphics[width=\textwidth,page=2]{Figures/samples-edm2-cfg-rows.pdf}\par\vspace{1pt}
\hbox to \textwidth{\makebox[0.244303\textwidth]{\itshape letter opener}\hspace{0.007594\textwidth}\makebox[0.244303\textwidth]{\itshape otter}\hspace{0.007594\textwidth}\makebox[0.244303\textwidth]{\itshape space bar}\hspace{0.007594\textwidth}\makebox[0.244303\textwidth]{\itshape limousine}\hss}\par\vspace{5pt}
\includegraphics[width=\textwidth,page=3]{Figures/samples-edm2-cfg-rows.pdf}\par\vspace{1pt}
\hbox to \textwidth{\makebox[0.244303\textwidth]{\itshape pay-phone}\hspace{0.007594\textwidth}\makebox[0.244303\textwidth]{\itshape crayfish}\hspace{0.007594\textwidth}\makebox[0.244303\textwidth]{\itshape chain}\hspace{0.007594\textwidth}\makebox[0.244303\textwidth]{\itshape porcupine}\hss}\par\vspace{5pt}
\includegraphics[width=\textwidth,page=4]{Figures/samples-edm2-cfg-rows.pdf}\par\vspace{1pt}
\hbox to \textwidth{\makebox[0.244303\textwidth]{\itshape banana}\hspace{0.007594\textwidth}\makebox[0.244303\textwidth]{\itshape teapot}\hspace{0.007594\textwidth}\makebox[0.244303\textwidth]{\itshape goose}\hspace{0.007594\textwidth}\makebox[0.244303\textwidth]{\itshape guinea pig}\hss}\par\vspace{5pt}
\includegraphics[width=\textwidth,page=5]{Figures/samples-edm2-cfg-rows.pdf}\par\vspace{1pt}
\hbox to \textwidth{\makebox[0.244303\textwidth]{\itshape American coot}\hspace{0.007594\textwidth}\makebox[0.244303\textwidth]{\itshape marmoset}\hspace{0.007594\textwidth}\makebox[0.244303\textwidth]{\itshape barn}\hspace{0.007594\textwidth}\makebox[0.244303\textwidth]{\itshape goose}\hss}\par\vspace{5pt}
\includegraphics[width=\textwidth,page=6]{Figures/samples-edm2-cfg-rows.pdf}\par\vspace{1pt}
\hbox to \textwidth{\makebox[0.244303\textwidth]{\itshape kit fox}\hspace{0.007594\textwidth}\makebox[0.244303\textwidth]{\itshape mailbox}\hspace{0.007594\textwidth}\makebox[0.244303\textwidth]{\itshape bell pepper}\hspace{0.007594\textwidth}\makebox[0.244303\textwidth]{\itshape buckeye}\hss}\par\vspace{5pt}
\includegraphics[width=\textwidth,page=7]{Figures/samples-edm2-cfg-rows.pdf}\par\vspace{1pt}
\hbox to \textwidth{\makebox[0.244303\textwidth]{\itshape worm fence}\hspace{0.007594\textwidth}\makebox[0.244303\textwidth]{\itshape waffle iron}\hspace{0.007594\textwidth}\makebox[0.244303\textwidth]{\itshape alligator lizard}\hspace{0.007594\textwidth}\makebox[0.244303\textwidth]{\itshape groenendael}\hss}\par\vspace{5pt}
}

\captionsetup{hypcap=false}%

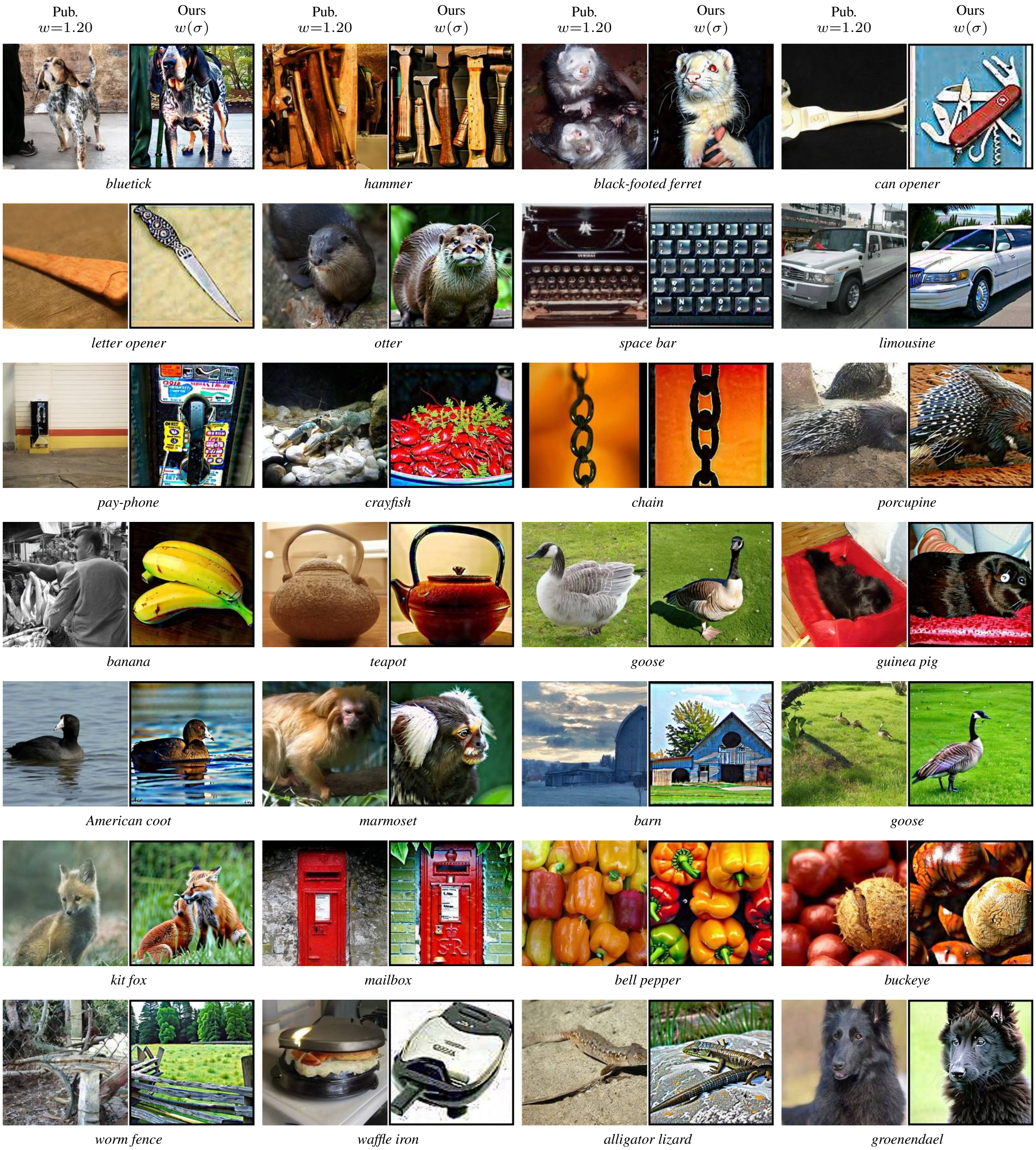
\captionof{figure}{Samples from EDM2-XXL on ImageNet-512 with classifier-free guidance by the unconditional EDM2-XS (\cref{tab:edm2-fail}), at the FID-tuned published scale (Pub., $w=1.2$) and with our schedule $w(\sigma)$ (framed in black); 32-step Heun sampler, $512^2$. Each pair shares its class and initial noise. The pairs are seeds 0--99 in order and not selected, except that four seeds whose drawn class gave images unsuitable for print (horizontal bar, bath towel, shower cap, sea cucumber) keep their initial noise but take another class (crayfish, guinea pig, sorrel, chime); this page and the next two.}
\label{fig:samples-edm2-cfg}
\end{minipage}\end{adjustbox}\par

{\renewcommand{\twocolumn}[1][]{\clearpage\noindent\begin{adjustbox}{max totalheight=\textheight,center}\begin{minipage}{\textwidth}#1\end{minipage}\end{adjustbox}\par}%
\twocolumn[{%
\centering
{\scriptsize\centering%
\hbox to \textwidth{\makebox[0.121518\textwidth]{\shortstack{Pub.\\$w=1.20$}}\hspace{0.001265\textwidth}\makebox[0.121518\textwidth]{\shortstack{Ours\\$w(\sigma)$}}\hspace{0.007594\textwidth}\makebox[0.121518\textwidth]{\shortstack{Pub.\\$w=1.20$}}\hspace{0.001265\textwidth}\makebox[0.121518\textwidth]{\shortstack{Ours\\$w(\sigma)$}}\hspace{0.007594\textwidth}\makebox[0.121518\textwidth]{\shortstack{Pub.\\$w=1.20$}}\hspace{0.001265\textwidth}\makebox[0.121518\textwidth]{\shortstack{Ours\\$w(\sigma)$}}\hspace{0.007594\textwidth}\makebox[0.121518\textwidth]{\shortstack{Pub.\\$w=1.20$}}\hspace{0.001265\textwidth}\makebox[0.121518\textwidth]{\shortstack{Ours\\$w(\sigma)$}}\hss}\par\vspace{3pt}
\includegraphics[width=\textwidth,page=8]{Figures/samples-edm2-cfg-rows.pdf}\par\vspace{1pt}
\hbox to \textwidth{\makebox[0.244303\textwidth]{\itshape redshank}\hspace{0.007594\textwidth}\makebox[0.244303\textwidth]{\itshape turnstile}\hspace{0.007594\textwidth}\makebox[0.244303\textwidth]{\itshape cicada}\hspace{0.007594\textwidth}\makebox[0.244303\textwidth]{\itshape thatch}\hss}\par\vspace{5pt}
\includegraphics[width=\textwidth,page=9]{Figures/samples-edm2-cfg-rows.pdf}\par\vspace{1pt}
\hbox to \textwidth{\makebox[0.244303\textwidth]{\itshape crayfish}\hspace{0.007594\textwidth}\makebox[0.244303\textwidth]{\itshape mortarboard}\hspace{0.007594\textwidth}\makebox[0.244303\textwidth]{\itshape hair slide}\hspace{0.007594\textwidth}\makebox[0.244303\textwidth]{\itshape rock python}\hss}\par\vspace{5pt}
\includegraphics[width=\textwidth,page=10]{Figures/samples-edm2-cfg-rows.pdf}\par\vspace{1pt}
\hbox to \textwidth{\makebox[0.244303\textwidth]{\itshape mortar}\hspace{0.007594\textwidth}\makebox[0.244303\textwidth]{\itshape goose}\hspace{0.007594\textwidth}\makebox[0.244303\textwidth]{\itshape parking meter}\hspace{0.007594\textwidth}\makebox[0.244303\textwidth]{\itshape freight car}\hss}\par\vspace{5pt}
\includegraphics[width=\textwidth,page=11]{Figures/samples-edm2-cfg-rows.pdf}\par\vspace{1pt}
\hbox to \textwidth{\makebox[0.244303\textwidth]{\itshape mongoose}\hspace{0.007594\textwidth}\makebox[0.244303\textwidth]{\itshape rain barrel}\hspace{0.007594\textwidth}\makebox[0.244303\textwidth]{\itshape long-horned beetle}\hspace{0.007594\textwidth}\makebox[0.244303\textwidth]{\itshape bell cote}\hss}\par\vspace{5pt}
\includegraphics[width=\textwidth,page=12]{Figures/samples-edm2-cfg-rows.pdf}\par\vspace{1pt}
\hbox to \textwidth{\makebox[0.244303\textwidth]{\itshape horned viper}\hspace{0.007594\textwidth}\makebox[0.244303\textwidth]{\itshape Tibetan terrier}\hspace{0.007594\textwidth}\makebox[0.244303\textwidth]{\itshape rain barrel}\hspace{0.007594\textwidth}\makebox[0.244303\textwidth]{\itshape shopping cart}\hss}\par\vspace{5pt}
\includegraphics[width=\textwidth,page=13]{Figures/samples-edm2-cfg-rows.pdf}\par\vspace{1pt}
\hbox to \textwidth{\makebox[0.244303\textwidth]{\itshape Bouvier des Flandres}\hspace{0.007594\textwidth}\makebox[0.244303\textwidth]{\itshape switch}\hspace{0.007594\textwidth}\makebox[0.244303\textwidth]{\itshape Maltese dog}\hspace{0.007594\textwidth}\makebox[0.244303\textwidth]{\itshape Maltese dog}\hss}\par\vspace{5pt}
\includegraphics[width=\textwidth,page=14]{Figures/samples-edm2-cfg-rows.pdf}\par\vspace{1pt}
\hbox to \textwidth{\makebox[0.244303\textwidth]{\itshape minibus}\hspace{0.007594\textwidth}\makebox[0.244303\textwidth]{\itshape leopard}\hspace{0.007594\textwidth}\makebox[0.244303\textwidth]{\itshape wok}\hspace{0.007594\textwidth}\makebox[0.244303\textwidth]{\itshape jack-o'-lantern}\hss}\par\vspace{5pt}
\includegraphics[width=\textwidth,page=15]{Figures/samples-edm2-cfg-rows.pdf}\par\vspace{1pt}
\hbox to \textwidth{\makebox[0.244303\textwidth]{\itshape running shoe}\hspace{0.007594\textwidth}\makebox[0.244303\textwidth]{\itshape shopping basket}\hspace{0.007594\textwidth}\makebox[0.244303\textwidth]{\itshape greenhouse}\hspace{0.007594\textwidth}\makebox[0.244303\textwidth]{\itshape goldfish}\hss}\par\vspace{5pt}
\includegraphics[width=\textwidth,page=16]{Figures/samples-edm2-cfg-rows.pdf}\par\vspace{1pt}
\hbox to \textwidth{\makebox[0.244303\textwidth]{\itshape common newt}\hspace{0.007594\textwidth}\makebox[0.244303\textwidth]{\itshape mitten}\hspace{0.007594\textwidth}\makebox[0.244303\textwidth]{\itshape wooden spoon}\hspace{0.007594\textwidth}\makebox[0.244303\textwidth]{\itshape hair spray}\hss}\par\vspace{5pt}
}
\captionsetup{hypcap=false}%
\captionof{figure}{More samples from EDM2-XXL with classifier-free guidance, continued from \cref{fig:samples-edm2-cfg} (seeds 28--63 of 100); same settings and layout.}
}]

\twocolumn[{%
\centering
{\scriptsize\centering%
\hbox to \textwidth{\makebox[0.121518\textwidth]{\shortstack{Pub.\\$w=1.20$}}\hspace{0.001265\textwidth}\makebox[0.121518\textwidth]{\shortstack{Ours\\$w(\sigma)$}}\hspace{0.007594\textwidth}\makebox[0.121518\textwidth]{\shortstack{Pub.\\$w=1.20$}}\hspace{0.001265\textwidth}\makebox[0.121518\textwidth]{\shortstack{Ours\\$w(\sigma)$}}\hspace{0.007594\textwidth}\makebox[0.121518\textwidth]{\shortstack{Pub.\\$w=1.20$}}\hspace{0.001265\textwidth}\makebox[0.121518\textwidth]{\shortstack{Ours\\$w(\sigma)$}}\hspace{0.007594\textwidth}\makebox[0.121518\textwidth]{\shortstack{Pub.\\$w=1.20$}}\hspace{0.001265\textwidth}\makebox[0.121518\textwidth]{\shortstack{Ours\\$w(\sigma)$}}\hss}\par\vspace{3pt}
\includegraphics[width=\textwidth,page=17]{Figures/samples-edm2-cfg-rows.pdf}\par\vspace{1pt}
\hbox to \textwidth{\makebox[0.244303\textwidth]{\itshape can opener}\hspace{0.007594\textwidth}\makebox[0.244303\textwidth]{\itshape Tibetan terrier}\hspace{0.007594\textwidth}\makebox[0.244303\textwidth]{\itshape barbershop}\hspace{0.007594\textwidth}\makebox[0.244303\textwidth]{\itshape trolleybus}\hss}\par\vspace{5pt}
\includegraphics[width=\textwidth,page=18]{Figures/samples-edm2-cfg-rows.pdf}\par\vspace{1pt}
\hbox to \textwidth{\makebox[0.244303\textwidth]{\itshape sorrel}\hspace{0.007594\textwidth}\makebox[0.244303\textwidth]{\itshape altar}\hspace{0.007594\textwidth}\makebox[0.244303\textwidth]{\itshape jaguar}\hspace{0.007594\textwidth}\makebox[0.244303\textwidth]{\itshape Bedlington terrier}\hss}\par\vspace{5pt}
\includegraphics[width=\textwidth,page=19]{Figures/samples-edm2-cfg-rows.pdf}\par\vspace{1pt}
\hbox to \textwidth{\makebox[0.244303\textwidth]{\itshape capuchin}\hspace{0.007594\textwidth}\makebox[0.244303\textwidth]{\itshape pencil box}\hspace{0.007594\textwidth}\makebox[0.244303\textwidth]{\itshape monastery}\hspace{0.007594\textwidth}\makebox[0.244303\textwidth]{\itshape conch}\hss}\par\vspace{5pt}
\includegraphics[width=\textwidth,page=20]{Figures/samples-edm2-cfg-rows.pdf}\par\vspace{1pt}
\hbox to \textwidth{\makebox[0.244303\textwidth]{\itshape malamute}\hspace{0.007594\textwidth}\makebox[0.244303\textwidth]{\itshape Boston bull}\hspace{0.007594\textwidth}\makebox[0.244303\textwidth]{\itshape spider monkey}\hspace{0.007594\textwidth}\makebox[0.244303\textwidth]{\itshape submarine}\hss}\par\vspace{5pt}
\includegraphics[width=\textwidth,page=21]{Figures/samples-edm2-cfg-rows.pdf}\par\vspace{1pt}
\hbox to \textwidth{\makebox[0.244303\textwidth]{\itshape racket}\hspace{0.007594\textwidth}\makebox[0.244303\textwidth]{\itshape studio couch}\hspace{0.007594\textwidth}\makebox[0.244303\textwidth]{\itshape crossword puzzle}\hspace{0.007594\textwidth}\makebox[0.244303\textwidth]{\itshape beaver}\hss}\par\vspace{5pt}
\includegraphics[width=\textwidth,page=22]{Figures/samples-edm2-cfg-rows.pdf}\par\vspace{1pt}
\hbox to \textwidth{\makebox[0.244303\textwidth]{\itshape barrel}\hspace{0.007594\textwidth}\makebox[0.244303\textwidth]{\itshape parking meter}\hspace{0.007594\textwidth}\makebox[0.244303\textwidth]{\itshape bobsled}\hspace{0.007594\textwidth}\makebox[0.244303\textwidth]{\itshape swing}\hss}\par\vspace{5pt}
\includegraphics[width=\textwidth,page=23]{Figures/samples-edm2-cfg-rows.pdf}\par\vspace{1pt}
\hbox to \textwidth{\makebox[0.244303\textwidth]{\itshape Band Aid}\hspace{0.007594\textwidth}\makebox[0.244303\textwidth]{\itshape pirate}\hspace{0.007594\textwidth}\makebox[0.244303\textwidth]{\itshape chime}\hspace{0.007594\textwidth}\makebox[0.244303\textwidth]{\itshape tow truck}\hss}\par\vspace{5pt}
\includegraphics[width=\textwidth,page=24]{Figures/samples-edm2-cfg-rows.pdf}\par\vspace{1pt}
\hbox to \textwidth{\makebox[0.244303\textwidth]{\itshape pick}\hspace{0.007594\textwidth}\makebox[0.244303\textwidth]{\itshape vulture}\hspace{0.007594\textwidth}\makebox[0.244303\textwidth]{\itshape desk}\hspace{0.007594\textwidth}\makebox[0.244303\textwidth]{\itshape Brittany spaniel}\hss}\par\vspace{5pt}
\includegraphics[width=\textwidth,page=25]{Figures/samples-edm2-cfg-rows.pdf}\par\vspace{1pt}
\hbox to \textwidth{\makebox[0.244303\textwidth]{\itshape conch}\hspace{0.007594\textwidth}\makebox[0.244303\textwidth]{\itshape French horn}\hspace{0.007594\textwidth}\makebox[0.244303\textwidth]{\itshape cornet}\hspace{0.007594\textwidth}\makebox[0.244303\textwidth]{\itshape fire screen}\hss}\par\vspace{5pt}
}
\captionsetup{hypcap=false}%
\captionof{figure}{More samples from EDM2-XXL with classifier-free guidance, continued from \cref{fig:samples-edm2-cfg} (seeds 64--99 of 100); same settings and layout.}
}]
}

\clearpage
\noindent\begin{adjustbox}{max totalheight=\dimexpr\textheight-26pt\relax,center}%
\begin{minipage}{\textwidth}\centering
{\scriptsize\centering%
\hbox to \textwidth{\makebox[0.121518\textwidth]{\shortstack{Pub.\\$w=2.05$}}\hspace{0.001265\textwidth}\makebox[0.121518\textwidth]{\shortstack{Ours\\$w(\sigma)$}}\hspace{0.007594\textwidth}\makebox[0.121518\textwidth]{\shortstack{Pub.\\$w=2.05$}}\hspace{0.001265\textwidth}\makebox[0.121518\textwidth]{\shortstack{Ours\\$w(\sigma)$}}\hspace{0.007594\textwidth}\makebox[0.121518\textwidth]{\shortstack{Pub.\\$w=2.05$}}\hspace{0.001265\textwidth}\makebox[0.121518\textwidth]{\shortstack{Ours\\$w(\sigma)$}}\hspace{0.007594\textwidth}\makebox[0.121518\textwidth]{\shortstack{Pub.\\$w=2.05$}}\hspace{0.001265\textwidth}\makebox[0.121518\textwidth]{\shortstack{Ours\\$w(\sigma)$}}\hss}\par\vspace{3pt}
\includegraphics[width=\textwidth,page=1]{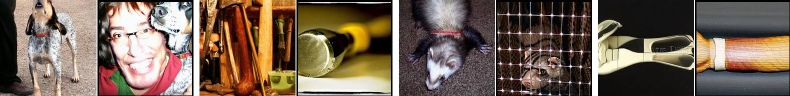}\par\vspace{1pt}
\hbox to \textwidth{\makebox[0.244303\textwidth]{\itshape bluetick}\hspace{0.007594\textwidth}\makebox[0.244303\textwidth]{\itshape hammer}\hspace{0.007594\textwidth}\makebox[0.244303\textwidth]{\itshape black-footed ferret}\hspace{0.007594\textwidth}\makebox[0.244303\textwidth]{\itshape can opener}\hss}\par\vspace{5pt}
\includegraphics[width=\textwidth,page=2]{Figures/samples-edm2-ag-rows.pdf}\par\vspace{1pt}
\hbox to \textwidth{\makebox[0.244303\textwidth]{\itshape letter opener}\hspace{0.007594\textwidth}\makebox[0.244303\textwidth]{\itshape otter}\hspace{0.007594\textwidth}\makebox[0.244303\textwidth]{\itshape space bar}\hspace{0.007594\textwidth}\makebox[0.244303\textwidth]{\itshape limousine}\hss}\par\vspace{5pt}
\includegraphics[width=\textwidth,page=3]{Figures/samples-edm2-ag-rows.pdf}\par\vspace{1pt}
\hbox to \textwidth{\makebox[0.244303\textwidth]{\itshape pay-phone}\hspace{0.007594\textwidth}\makebox[0.244303\textwidth]{\itshape crayfish}\hspace{0.007594\textwidth}\makebox[0.244303\textwidth]{\itshape chain}\hspace{0.007594\textwidth}\makebox[0.244303\textwidth]{\itshape porcupine}\hss}\par\vspace{5pt}
\includegraphics[width=\textwidth,page=4]{Figures/samples-edm2-ag-rows.pdf}\par\vspace{1pt}
\hbox to \textwidth{\makebox[0.244303\textwidth]{\itshape banana}\hspace{0.007594\textwidth}\makebox[0.244303\textwidth]{\itshape teapot}\hspace{0.007594\textwidth}\makebox[0.244303\textwidth]{\itshape goose}\hspace{0.007594\textwidth}\makebox[0.244303\textwidth]{\itshape guinea pig}\hss}\par\vspace{5pt}
\includegraphics[width=\textwidth,page=5]{Figures/samples-edm2-ag-rows.pdf}\par\vspace{1pt}
\hbox to \textwidth{\makebox[0.244303\textwidth]{\itshape American coot}\hspace{0.007594\textwidth}\makebox[0.244303\textwidth]{\itshape marmoset}\hspace{0.007594\textwidth}\makebox[0.244303\textwidth]{\itshape barn}\hspace{0.007594\textwidth}\makebox[0.244303\textwidth]{\itshape goose}\hss}\par\vspace{5pt}
\includegraphics[width=\textwidth,page=6]{Figures/samples-edm2-ag-rows.pdf}\par\vspace{1pt}
\hbox to \textwidth{\makebox[0.244303\textwidth]{\itshape kit fox}\hspace{0.007594\textwidth}\makebox[0.244303\textwidth]{\itshape mailbox}\hspace{0.007594\textwidth}\makebox[0.244303\textwidth]{\itshape bell pepper}\hspace{0.007594\textwidth}\makebox[0.244303\textwidth]{\itshape buckeye}\hss}\par\vspace{5pt}
\includegraphics[width=\textwidth,page=7]{Figures/samples-edm2-ag-rows.pdf}\par\vspace{1pt}
\hbox to \textwidth{\makebox[0.244303\textwidth]{\itshape worm fence}\hspace{0.007594\textwidth}\makebox[0.244303\textwidth]{\itshape waffle iron}\hspace{0.007594\textwidth}\makebox[0.244303\textwidth]{\itshape alligator lizard}\hspace{0.007594\textwidth}\makebox[0.244303\textwidth]{\itshape groenendael}\hss}\par\vspace{5pt}
}

\captionsetup{hypcap=false}%

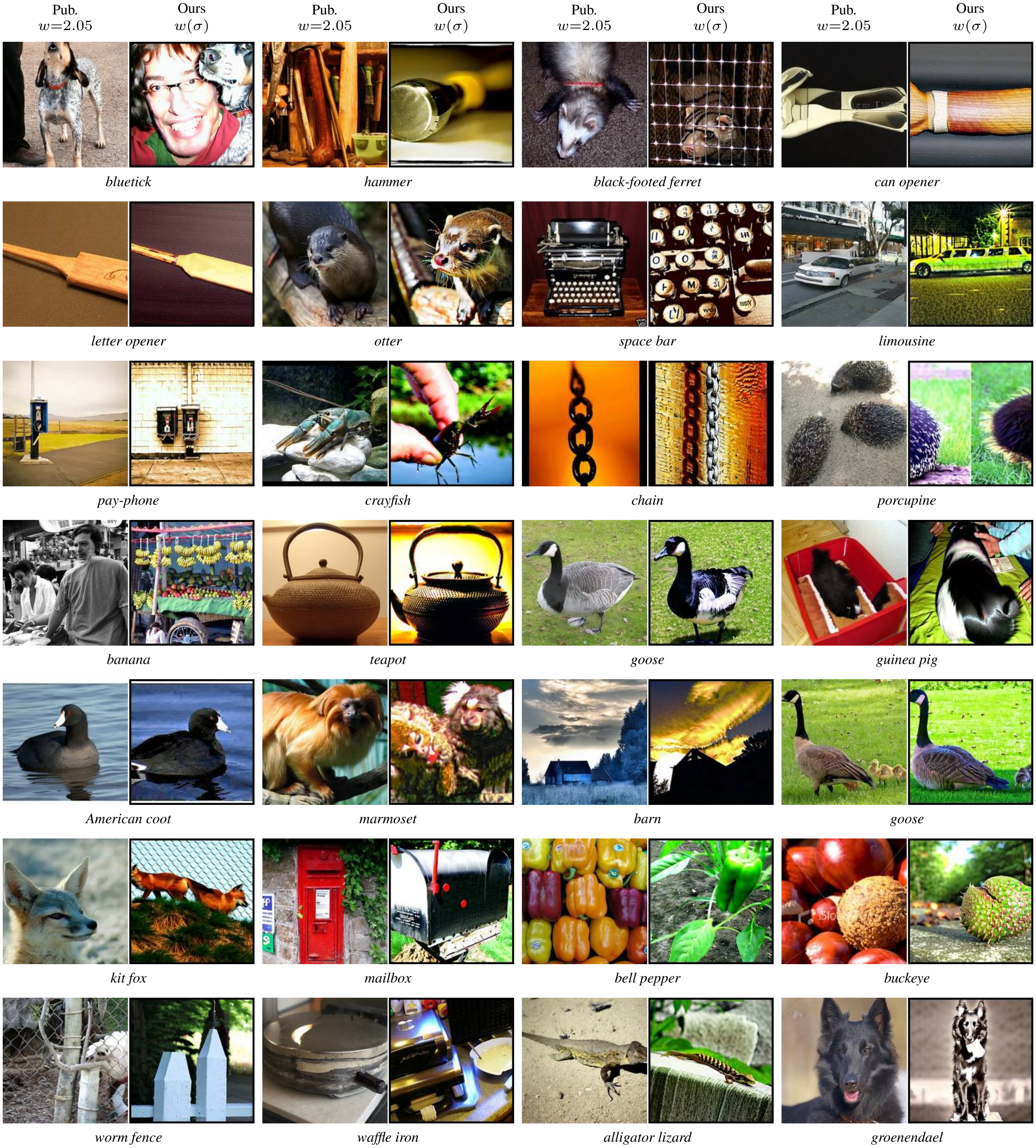
\captionof{figure}{Samples from EDM2-XXL on ImageNet-512 with autoguidance by EDM2-M (\cref{tab:edm2-fail}), at the FID-tuned published scale (Pub., $w=2.05$) and with our schedule $w(\sigma)$ (framed in black); 32-step Heun sampler, $512^2$. Each pair shares its class and initial noise, and the seeds and classes are those of \cref{fig:samples-edm2-cfg}; this page and the next two.}
\label{fig:samples-edm2-ag}
\end{minipage}\end{adjustbox}\par

{\renewcommand{\twocolumn}[1][]{\clearpage\noindent\begin{adjustbox}{max totalheight=\textheight,center}\begin{minipage}{\textwidth}#1\end{minipage}\end{adjustbox}\par}%
\twocolumn[{%
\centering
{\scriptsize\centering%
\hbox to \textwidth{\makebox[0.121518\textwidth]{\shortstack{Pub.\\$w=2.05$}}\hspace{0.001265\textwidth}\makebox[0.121518\textwidth]{\shortstack{Ours\\$w(\sigma)$}}\hspace{0.007594\textwidth}\makebox[0.121518\textwidth]{\shortstack{Pub.\\$w=2.05$}}\hspace{0.001265\textwidth}\makebox[0.121518\textwidth]{\shortstack{Ours\\$w(\sigma)$}}\hspace{0.007594\textwidth}\makebox[0.121518\textwidth]{\shortstack{Pub.\\$w=2.05$}}\hspace{0.001265\textwidth}\makebox[0.121518\textwidth]{\shortstack{Ours\\$w(\sigma)$}}\hspace{0.007594\textwidth}\makebox[0.121518\textwidth]{\shortstack{Pub.\\$w=2.05$}}\hspace{0.001265\textwidth}\makebox[0.121518\textwidth]{\shortstack{Ours\\$w(\sigma)$}}\hss}\par\vspace{3pt}
\includegraphics[width=\textwidth,page=8]{Figures/samples-edm2-ag-rows.pdf}\par\vspace{1pt}
\hbox to \textwidth{\makebox[0.244303\textwidth]{\itshape redshank}\hspace{0.007594\textwidth}\makebox[0.244303\textwidth]{\itshape turnstile}\hspace{0.007594\textwidth}\makebox[0.244303\textwidth]{\itshape cicada}\hspace{0.007594\textwidth}\makebox[0.244303\textwidth]{\itshape thatch}\hss}\par\vspace{5pt}
\includegraphics[width=\textwidth,page=9]{Figures/samples-edm2-ag-rows.pdf}\par\vspace{1pt}
\hbox to \textwidth{\makebox[0.244303\textwidth]{\itshape crayfish}\hspace{0.007594\textwidth}\makebox[0.244303\textwidth]{\itshape mortarboard}\hspace{0.007594\textwidth}\makebox[0.244303\textwidth]{\itshape hair slide}\hspace{0.007594\textwidth}\makebox[0.244303\textwidth]{\itshape rock python}\hss}\par\vspace{5pt}
\includegraphics[width=\textwidth,page=10]{Figures/samples-edm2-ag-rows.pdf}\par\vspace{1pt}
\hbox to \textwidth{\makebox[0.244303\textwidth]{\itshape mortar}\hspace{0.007594\textwidth}\makebox[0.244303\textwidth]{\itshape goose}\hspace{0.007594\textwidth}\makebox[0.244303\textwidth]{\itshape parking meter}\hspace{0.007594\textwidth}\makebox[0.244303\textwidth]{\itshape freight car}\hss}\par\vspace{5pt}
\includegraphics[width=\textwidth,page=11]{Figures/samples-edm2-ag-rows.pdf}\par\vspace{1pt}
\hbox to \textwidth{\makebox[0.244303\textwidth]{\itshape mongoose}\hspace{0.007594\textwidth}\makebox[0.244303\textwidth]{\itshape rain barrel}\hspace{0.007594\textwidth}\makebox[0.244303\textwidth]{\itshape long-horned beetle}\hspace{0.007594\textwidth}\makebox[0.244303\textwidth]{\itshape bell cote}\hss}\par\vspace{5pt}
\includegraphics[width=\textwidth,page=12]{Figures/samples-edm2-ag-rows.pdf}\par\vspace{1pt}
\hbox to \textwidth{\makebox[0.244303\textwidth]{\itshape horned viper}\hspace{0.007594\textwidth}\makebox[0.244303\textwidth]{\itshape Tibetan terrier}\hspace{0.007594\textwidth}\makebox[0.244303\textwidth]{\itshape rain barrel}\hspace{0.007594\textwidth}\makebox[0.244303\textwidth]{\itshape shopping cart}\hss}\par\vspace{5pt}
\includegraphics[width=\textwidth,page=13]{Figures/samples-edm2-ag-rows.pdf}\par\vspace{1pt}
\hbox to \textwidth{\makebox[0.244303\textwidth]{\itshape Bouvier des Flandres}\hspace{0.007594\textwidth}\makebox[0.244303\textwidth]{\itshape switch}\hspace{0.007594\textwidth}\makebox[0.244303\textwidth]{\itshape Maltese dog}\hspace{0.007594\textwidth}\makebox[0.244303\textwidth]{\itshape Maltese dog}\hss}\par\vspace{5pt}
\includegraphics[width=\textwidth,page=14]{Figures/samples-edm2-ag-rows.pdf}\par\vspace{1pt}
\hbox to \textwidth{\makebox[0.244303\textwidth]{\itshape minibus}\hspace{0.007594\textwidth}\makebox[0.244303\textwidth]{\itshape leopard}\hspace{0.007594\textwidth}\makebox[0.244303\textwidth]{\itshape wok}\hspace{0.007594\textwidth}\makebox[0.244303\textwidth]{\itshape jack-o'-lantern}\hss}\par\vspace{5pt}
\includegraphics[width=\textwidth,page=15]{Figures/samples-edm2-ag-rows.pdf}\par\vspace{1pt}
\hbox to \textwidth{\makebox[0.244303\textwidth]{\itshape running shoe}\hspace{0.007594\textwidth}\makebox[0.244303\textwidth]{\itshape shopping basket}\hspace{0.007594\textwidth}\makebox[0.244303\textwidth]{\itshape greenhouse}\hspace{0.007594\textwidth}\makebox[0.244303\textwidth]{\itshape goldfish}\hss}\par\vspace{5pt}
\includegraphics[width=\textwidth,page=16]{Figures/samples-edm2-ag-rows.pdf}\par\vspace{1pt}
\hbox to \textwidth{\makebox[0.244303\textwidth]{\itshape common newt}\hspace{0.007594\textwidth}\makebox[0.244303\textwidth]{\itshape mitten}\hspace{0.007594\textwidth}\makebox[0.244303\textwidth]{\itshape wooden spoon}\hspace{0.007594\textwidth}\makebox[0.244303\textwidth]{\itshape hair spray}\hss}\par\vspace{5pt}
}
\captionsetup{hypcap=false}%
\captionof{figure}{More samples from EDM2-XXL with autoguidance, continued from \cref{fig:samples-edm2-ag} (seeds 28--63 of 100); same settings and layout.}
}]

\twocolumn[{%
\centering
{\scriptsize\centering%
\hbox to \textwidth{\makebox[0.121518\textwidth]{\shortstack{Pub.\\$w=2.05$}}\hspace{0.001265\textwidth}\makebox[0.121518\textwidth]{\shortstack{Ours\\$w(\sigma)$}}\hspace{0.007594\textwidth}\makebox[0.121518\textwidth]{\shortstack{Pub.\\$w=2.05$}}\hspace{0.001265\textwidth}\makebox[0.121518\textwidth]{\shortstack{Ours\\$w(\sigma)$}}\hspace{0.007594\textwidth}\makebox[0.121518\textwidth]{\shortstack{Pub.\\$w=2.05$}}\hspace{0.001265\textwidth}\makebox[0.121518\textwidth]{\shortstack{Ours\\$w(\sigma)$}}\hspace{0.007594\textwidth}\makebox[0.121518\textwidth]{\shortstack{Pub.\\$w=2.05$}}\hspace{0.001265\textwidth}\makebox[0.121518\textwidth]{\shortstack{Ours\\$w(\sigma)$}}\hss}\par\vspace{3pt}
\includegraphics[width=\textwidth,page=17]{Figures/samples-edm2-ag-rows.pdf}\par\vspace{1pt}
\hbox to \textwidth{\makebox[0.244303\textwidth]{\itshape can opener}\hspace{0.007594\textwidth}\makebox[0.244303\textwidth]{\itshape Tibetan terrier}\hspace{0.007594\textwidth}\makebox[0.244303\textwidth]{\itshape barbershop}\hspace{0.007594\textwidth}\makebox[0.244303\textwidth]{\itshape trolleybus}\hss}\par\vspace{5pt}
\includegraphics[width=\textwidth,page=18]{Figures/samples-edm2-ag-rows.pdf}\par\vspace{1pt}
\hbox to \textwidth{\makebox[0.244303\textwidth]{\itshape sorrel}\hspace{0.007594\textwidth}\makebox[0.244303\textwidth]{\itshape altar}\hspace{0.007594\textwidth}\makebox[0.244303\textwidth]{\itshape jaguar}\hspace{0.007594\textwidth}\makebox[0.244303\textwidth]{\itshape Bedlington terrier}\hss}\par\vspace{5pt}
\includegraphics[width=\textwidth,page=19]{Figures/samples-edm2-ag-rows.pdf}\par\vspace{1pt}
\hbox to \textwidth{\makebox[0.244303\textwidth]{\itshape capuchin}\hspace{0.007594\textwidth}\makebox[0.244303\textwidth]{\itshape pencil box}\hspace{0.007594\textwidth}\makebox[0.244303\textwidth]{\itshape monastery}\hspace{0.007594\textwidth}\makebox[0.244303\textwidth]{\itshape conch}\hss}\par\vspace{5pt}
\includegraphics[width=\textwidth,page=20]{Figures/samples-edm2-ag-rows.pdf}\par\vspace{1pt}
\hbox to \textwidth{\makebox[0.244303\textwidth]{\itshape malamute}\hspace{0.007594\textwidth}\makebox[0.244303\textwidth]{\itshape Boston bull}\hspace{0.007594\textwidth}\makebox[0.244303\textwidth]{\itshape spider monkey}\hspace{0.007594\textwidth}\makebox[0.244303\textwidth]{\itshape submarine}\hss}\par\vspace{5pt}
\includegraphics[width=\textwidth,page=21]{Figures/samples-edm2-ag-rows.pdf}\par\vspace{1pt}
\hbox to \textwidth{\makebox[0.244303\textwidth]{\itshape racket}\hspace{0.007594\textwidth}\makebox[0.244303\textwidth]{\itshape studio couch}\hspace{0.007594\textwidth}\makebox[0.244303\textwidth]{\itshape crossword puzzle}\hspace{0.007594\textwidth}\makebox[0.244303\textwidth]{\itshape beaver}\hss}\par\vspace{5pt}
\includegraphics[width=\textwidth,page=22]{Figures/samples-edm2-ag-rows.pdf}\par\vspace{1pt}
\hbox to \textwidth{\makebox[0.244303\textwidth]{\itshape barrel}\hspace{0.007594\textwidth}\makebox[0.244303\textwidth]{\itshape parking meter}\hspace{0.007594\textwidth}\makebox[0.244303\textwidth]{\itshape bobsled}\hspace{0.007594\textwidth}\makebox[0.244303\textwidth]{\itshape swing}\hss}\par\vspace{5pt}
\includegraphics[width=\textwidth,page=23]{Figures/samples-edm2-ag-rows.pdf}\par\vspace{1pt}
\hbox to \textwidth{\makebox[0.244303\textwidth]{\itshape Band Aid}\hspace{0.007594\textwidth}\makebox[0.244303\textwidth]{\itshape pirate}\hspace{0.007594\textwidth}\makebox[0.244303\textwidth]{\itshape chime}\hspace{0.007594\textwidth}\makebox[0.244303\textwidth]{\itshape tow truck}\hss}\par\vspace{5pt}
\includegraphics[width=\textwidth,page=24]{Figures/samples-edm2-ag-rows.pdf}\par\vspace{1pt}
\hbox to \textwidth{\makebox[0.244303\textwidth]{\itshape pick}\hspace{0.007594\textwidth}\makebox[0.244303\textwidth]{\itshape vulture}\hspace{0.007594\textwidth}\makebox[0.244303\textwidth]{\itshape desk}\hspace{0.007594\textwidth}\makebox[0.244303\textwidth]{\itshape Brittany spaniel}\hss}\par\vspace{5pt}
\includegraphics[width=\textwidth,page=25]{Figures/samples-edm2-ag-rows.pdf}\par\vspace{1pt}
\hbox to \textwidth{\makebox[0.244303\textwidth]{\itshape conch}\hspace{0.007594\textwidth}\makebox[0.244303\textwidth]{\itshape French horn}\hspace{0.007594\textwidth}\makebox[0.244303\textwidth]{\itshape cornet}\hspace{0.007594\textwidth}\makebox[0.244303\textwidth]{\itshape fire screen}\hss}\par\vspace{5pt}
}
\captionsetup{hypcap=false}%
\captionof{figure}{More samples from EDM2-XXL with autoguidance, continued from \cref{fig:samples-edm2-ag} (seeds 64--99 of 100); same settings and layout.}
}]
}
\twocolumn

{
    \small
    \bibliographystyle{ieeenat_fullname}
    \bibliography{bib/references}
}

\end{document}